%% file: main_10.16.tex
\documentclass{article}
\usepackage[main, final]{neurips_2026}

\usepackage[utf8]{inputenc}
\usepackage[T1]{fontenc}
\usepackage{nicefrac}
\usepackage{microtype}



\usepackage[hidelinks]{hyperref}
\usepackage{url}
\usepackage{amsmath,amssymb,amsthm,bm}
\usepackage{graphicx}
\usepackage{wrapfig}
\usepackage{caption}
\usepackage{multirow}
\usepackage{titletoc}
\usepackage{longtable}

\usepackage[table]{xcolor}   
\usepackage{booktabs}
\usepackage{array}
\usepackage{float}
\usepackage{placeins}
\usepackage{enumitem}

\usepackage{algorithm}
\usepackage{algpseudocode}

\usepackage{tikz}
\usetikzlibrary{arrows.meta,positioning,decorations.pathreplacing,calc}
\usepackage{pifont}

\newtheorem{theorem}{Theorem}
\newtheorem{proposition}[theorem]{Proposition}
\newtheorem{lemma}[theorem]{Lemma}
\newtheorem{corollary}[theorem]{Corollary}
\theoremstyle{definition}

\theoremstyle{remark}
\newtheorem{remark}[theorem]{Remark}

\definecolor{keybg}{gray}{0.92}         
\definecolor{tabband}{HTML}{E8F1FA}     

\providecolor{eqblue}{HTML}{DCE9F7}      
\providecolor{eqgreen}{HTML}{DFF0E3}     
\providecolor{lblblue}{HTML}{2F6FB0}     
\providecolor{lblgreen}{HTML}{2E7D4F}    

\newcommand{\bluebox}[1]{\colorbox{eqblue}{$\displaystyle #1$}}
\newcommand{\greenbox}[1]{\colorbox{eqgreen}{$\displaystyle #1$}}
\newcommand{\blbl}[1]{\text{\scriptsize\textcolor{lblblue}{#1}}}
\newcommand{\glbl}[1]{\text{\scriptsize\textcolor{lblgreen}{#1}}}

\definecolor{tierAlg}{HTML}{DCE9D5}     
\definecolor{tierStat}{HTML}{FFF3D6}    
\definecolor{tierMis}{HTML}{FBEDF1}     

\definecolor{dsband}{HTML}{DCE6F4}\definecolor{dsbtx}{HTML}{2F5F9E}
\newcommand{\tbandn}[4]{\rowcolor{#2}[\tabcolsep][\tabcolsep]\multicolumn{#1}{l}{\textcolor{#3}{\rule{0pt}{1.8ex}\scriptsize\textit{#4}}}\\}

\definecolor{ghead}{HTML}{2F6FB0}\definecolor{gband}{HTML}{EAF2FB}\definecolor{gbtx}{HTML}{1F4E79}
\newcommand{\ghead}{\rowcolor{ghead}}
\newcommand{\gband}[2]{\hline\rowcolor{gband}\multicolumn{#1}{|l|}{\rule{0pt}{2.3ex}\textcolor{gbtx}{\textbf{#2}}}\\}

\definecolor{mkgreen}{HTML}{2E9E6A}\definecolor{mkred}{HTML}{D64545}\definecolor{mkorange}{HTML}{E59A2F}
\definecolor{oursrow}{HTML}{E2F0D2}
\definecolor{trband}{HTML}{D6EEEB}\definecolor{trbtx}{HTML}{1E7A70}
\definecolor{grband}{HTML}{FBEFC8}\definecolor{grbtx}{HTML}{A87A0F}
\definecolor{opband}{HTML}{F3E3EC}\definecolor{opbtx}{HTML}{8E3A6B}
\newcommand{\mkyes}{\textcolor{mkgreen}{\ding{51}}}
\newcommand{\mkno}{\textcolor{mkred}{\ding{55}}}
\newcommand{\mkpart}{\textcolor{mkorange}{\ding{51}}\textsuperscript{\textcolor{mkorange}{\tiny p}}}

\definecolor{dr0}{HTML}{FFFFFF}\definecolor{dr1}{HTML}{F1F6FC}\definecolor{dr2}{HTML}{E0EBF7}
\definecolor{dr3}{HTML}{CBDCF1}\definecolor{dr4}{HTML}{B2CBE9}\definecolor{dr5}{HTML}{97B8DF}
\definecolor{hdr}{HTML}{EDEDED}               
\newcommand{\tgroup}[2]{\multicolumn{#1}{l}{\rule{0pt}{2.2ex}\textit{#2}}\\}
\newcommand{\best}[1]{\mathbf{#1}}
\newcommand{\second}[1]{\underline{#1}}

\newcommand{\Lop}{\mathbf{L}}                 
\newcommand{\Ham}{\mathbf{H}}                 

\newcommand{\Cay}{\operatorname{Cay}}         
\newcommand{\Iden}[1]{I_{#1}}                 
\newcommand{\zmodal}{\tilde{z}}               
\newcommand{\dk}{d_k}                         
\newcommand{\lk}{\lambda_k}                   
\newcommand{\kzero}{k^{*}}                    
\newcommand{\onevec}{\mathbf{1}}              
\newcommand{\uniform}{\onevec/\sqrt{N}}       
\newcommand{\cost}{O(N^{3}+N^{2}m)}
\newcommand{\costfull}{O(N^{3}m^{3})}

\newsavebox{\keybox}

\title{Beyond Unit-Circle Eigenvalues: Invariant Bases for Stable State Space Dynamics}

\author{%
  \textbf{Jiaqian Zhu}$^{1,\dagger}$ \quad
  \textbf{Yang Zhang}$^{2,*}$ \quad
  \textbf{Junhua Ding}$^{3}$ \quad
  \textbf{Xiaowei Yu}$^{1,*}$ \\
  {\small $^{1}$Missouri University of Science and Technology \quad
  $^{2}$University of North Texas \quad
  $^{3}$University of Wyoming} \\
  {\small\texttt{jzbmn@mst.edu, xych8@mst.edu, Yang.Zhang@unt.edu, Junhua.Ding@uwyo.edu}}
}

\begin{document}
\maketitle

\begingroup
\makeatletter\renewcommand{\@makefntext}[1]{\noindent#1}\makeatother
\renewcommand{\thefootnote}{}
\footnotetext{$^{*}$Corresponding authors.\enspace $^{\dagger}$Project leader.}
\endgroup

\begin{abstract}
Diagonal state space models process long sequences at linear cost, yet the same transition is
applied at every step of a rollout, so an unconstrained recurrence holds no quantity fixed and the
state decays or grows without bound.
Current designs constrain one quantity, and 
preserving one does not guarantee preservation of the others,
so a model can hold its norm to machine precision while momentum drifts
by hundreds of percent.
We claim that the conserved quantities of a diagonal recurrence are
layered, that each layer requires a strictly stronger condition on the operator, and that a
unit-modulus spectrum reaches only the outermost.
Acting on the claim, we enforce the conditions required for every layer by construction.
The generator
of the Kronecker--Cayley transition (KC) is a graph Laplacian over object subspaces,
$\mathbf{H}=\mathbf{L}\otimes I_m$, advanced by a Cayley transform, and the Kronecker structure
passes through the transform, $\mathrm{Cay}(\mathbf{L}\otimes I_m)=\mathrm{Cay}(\mathbf{L})\otimes
I_m$, so the step stays diagonal at $O(N^3{+}N^2m)$ cost.
The conserved set is exactly a $2m^2$-dimensional family of quadratic forms, held at every
parameter value, so the quantities and the dynamics the construction reaches are fixed before
training.
We evaluate KC on four conservation benchmarks and show that it holds total mass at or near
the reference floor while every diagonal baseline loses it by orders of magnitude.
We further compare against mass correction and find that two correctors fail once the
underlying prediction diverges, which shows that assembly rather than correction is necessary for
a guarantee that holds across the rollout.
A free modal basis trained under a conservation penalty conserves only by freezing its spectrum,
whereas the assembled operator conserves with every mode but one free to rotate.
Beyond these benchmarks, KC holds the linear momentum of three $N$-body systems. At the
operator level, the same layering holds the conserved quantity in molecular
dynamics, chemical kinetics, pharmacokinetics and power-grid swing dynamics.
Our code is available at \url{https://github.com/Jiaqian-Janelle/NeurIPS26-Kronecker-Cayley-Stable-State-Space-Dynamics}.
\end{abstract}


\input{S1_introduction}
\input{S2_related_work}
\input{S3_method}
\input{S4_experiments_10.14}
\input{S5_conclusions}


\begin{ack}
\paragraph{Author contributions.}
Jiaqian Zhu conceived the project, developed the theory, designed the method, implemented the
code, conducted all experiments, and wrote the manuscript. Yang Zhang provided guidance on the
revision of the manuscript and contributed to discussions of the work. Xiaowei Yu provided the
primary computing resources. Junhua Ding provided additional computing resources.

\paragraph{Funding.}
Experiments were run on an NVIDIA RTX PRO 6000 (96\,GB) workstation provided by Xiaowei Yu at
Missouri University of Science and Technology. Additional computing resources were provided by
Junhua Ding at the University of Wyoming. Jiaqian Zhu was supported by a teaching assistantship at
Missouri University of Science and Technology. This work received no other third-party funding.

\paragraph{Competing interests.}
The authors declare no competing interests.
\end{ack}

\bibliography{refs}
\bibliographystyle{plainnat}


\input{S6_supplementary}

\clearpage
\input{checklist}

\end{document}

%% file: S1_introduction.tex
\section{Introduction}
\label{sec:intro}

Diagonal state space models have become a general parameterisation for long sequences, advancing
a latent state by one diagonal transition per step at linear cost, which has carried them from
language to audio, vision and physical dynamics \citep{gu2022parameterization,
smith2023simplified, gu2024mamba}. Yet the same transition is re-applied to every step of an
autoregressive rollout, so that a
deviation in one mode compounds rather than corrects and the latent norm decays towards zero or
grows without bound. Two lines of work address this drift, one through a spectral or manifold
constraint that places the eigenvalues on the unit circle \citep{orvieto2023resurrecting,
arjovsky2016unitary, lezcano2019cheap}, and another through a structured generator that builds a
prescribed quantity into a Hamiltonian \citep{greydanus2019hamiltonian, chen2020symplectic} or
into a constraint set solved at every step \citep{lee2021deep}. Both constrain the quantity they
target, which leaves open the question we address: \emph{which quantities can a
diagonal recurrence preserve, which condition on the operator does each one require, and why
does a conservation objective added to training not supply them?}

\begin{wrapfigure}{r}{0.30\textwidth}
\vspace{-10pt}
\centering
\includegraphics{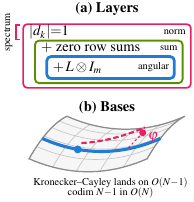}
\caption{Conserved quantities are layered, and constraining the spectrum reaches only the
outermost. \textbf{(a)} Each layer requires a strictly stronger condition.
\textbf{(b)} The conserving pairs of basis and spectrum form a lower-dimensional set, and a
basis trained under a penalty remains near its initialisation instead of moving to it.
}
\label{fig:layers}
\vspace{-14pt}
\end{wrapfigure}

Three gaps separate that question from current practice. A transition with every eigenvalue on
the unit circle holds the norm to machine precision, yet a summed quantity such as momentum
drifts by $141$ percent, because the norm is a property of the spectrum whereas a sum is
preserved only when the uniform vector is an eigenvector, a property of the coordinates.
\emph{Stability and conservation are therefore distinct properties, and a constraint on the
spectrum reaches only the first.} A conservation penalty is the usual response to the second,
and it changes the objective without changing the structure, so it may not recover the physics
even on modestly complex problems \citep{krishnapriyan2021characterizing}. An exact operator
resolves that gap but leaves a third, since its range of applicability is unknown and a
norm-preserving map degrades accuracy on dynamics that contract or expand
\citep{vorontsov2017orthogonality, kerg2019nonnormal}. The conditions that decide conservation
must therefore be placed on the coordinates, encoded in the structure of the generator, and
derived from a single algebraic condition that also yields the criterion for where they apply.

\begingroup\looseness=-1
Following this principle, we design the transition from the condition that decides what it
conserves. For a transition obtained by the Cayley transform of a Hermitian generator, a
quadratic form is conserved exactly when its generator commutes with $\Ham$, so the conserved
set is fixed by the algebraic structure of $\Ham$ alone. Three requirements on that structure
follow, each strictly stronger than the last: Hermiticity for the norm, vanishing row sums for
the summed quantity, and the Kronecker form $\Lop\otimes\Iden{m}$ for the angular quantity, so
the three are nested and a unit-modulus spectrum reaches only the outermost
(Figure~\ref{fig:layers}a). We therefore propose the Kronecker--Cayley transition (KC), which
assembles $\Ham=\Lop\otimes\Iden{m}$ from one learned coupling per object pair and advances the
latent by $\Cay(\Ham)$, so that all three requirements are met at once. They are properties of
the pattern rather than of the values placed in it, so they hold at every value of the coupling
and no correction is required. The same equivalence supplies the missing criterion, since a
quadratic physical quantity is held exactly when its generator lies in that family, and a
unitary step cannot represent contraction. Holding the summed quantity requires every rotating
mode to carry none of the uniform vector, which the assembly satisfies by placing that vector in
the null mode and which a trained basis reaches only by freezing every mode
(Figure~\ref{fig:layers}b).\par\endgroup


On four conservation benchmarks \citep{takamoto2022pdebench, gupta2022pdearena}, KC holds total
mass to between $4\times10^{-6}$ and $9\times10^{-3}$ in absolute units, at or near the reference
floor, while four diagonal baselines under the same encoder lose between $4$ and
$4\times10^{9}$. Two mass correctors, one learned end-to-end \citep{liu2025cpfno} and one fixed
and uniform, both fail once the underlying prediction has drifted too far, whereas KC reaches
the final horizon on every seed. Removing one condition of the assembly costs exactly the
quantity it holds, and a modal basis trained under a conservation penalty conserves
only by freezing every mode. KC also holds the linear momentum of three
$N$-body systems \citep{satorras2021en, brandstetter2022geometric} to $2\times10^{-6}$ against
$0.5$ to $8$ for every baseline, and at the operator level the conserved quantity of molecular
dynamics, chemical kinetics, pharmacokinetics and power-grid swing dynamics to within $10^{-2}$
percent, while on a dissipative system it fits less accurately than baselines that can contract,
as the theory predicts.

\begingroup\looseness=-1
\noindent\textbf{Our contributions are as follows.} \textbf{(i)} We prove that the conserved
quantities of a diagonal recurrence are nested, each requiring a strictly stronger condition, and
that a unit-modulus spectrum reaches only the outermost (\S\ref{sec:family}). \textbf{(ii)} We
propose the Kronecker--Cayley transition, which satisfies every condition at once, and prove
that its conserved set is exactly a $2m^2$-dimensional family held at every parameter value
(\S\ref{sec:assembly}). \textbf{(iii)} We characterise the transitions that hold the summed
quantity, show that a trained basis reaches them only by freezing its spectrum, and give a
criterion, quadratic in the quantity and conservative in the dynamics, for where the
construction applies (\S\ref{sec:scope}). \textbf{(iv)} We validate KC on four
benchmarks, where it holds the conserved quantity at or near the floor while every baseline
loses it and mass correction fails, and on dynamical systems across physics, chemistry and
engineering (\S\ref{sec:experiments}).\par\endgroup

%% file: S2_related_work.tex
\section{Related Work}
\label{sec:related}

\begin{table}[!h]
\centering\footnotesize
\setlength{\tabcolsep}{5.5pt}
\renewcommand{\arraystretch}{0.92}
\setlength{\aboverulesep}{0.15ex}\setlength{\belowrulesep}{0.3ex}
\caption{\textbf{What each design constrains, which layer it reaches, and what it leaves free.}
The first three mark columns are the layers of Corollary~\ref{cor:layered}, and a partial mark
\mkpart\ denotes a bounded norm, a quantity held only if named as a constraint, or one restored
by a corrector while the underlying prediction stays within its reach.}
\label{tab:layer_constraint}
\begin{tabular}{l l ccc cc l}
\toprule
\textbf{Design} & \textbf{Designed object} & \textbf{Norm} & \textbf{Sum} & \textbf{Angular} & \textbf{Diag.} & \textbf{All $\theta$} & \textbf{Leaves free} \\
\midrule
\tbandn{8}{dsband}{dsbtx}{Spectrum: constrain the eigenvalues}
Diagonal SSM & eigenvalues & \mkno & \mkno & \mkno & \mkyes & \mkno & eigenvectors \\
LRU & spectrum & \mkpart & \mkno & \mkno & \mkyes & \mkyes & eigenvectors \\
Unitary RNN & operator & \mkyes & \mkno & \mkno & \mkno & \mkyes & designated quantity \\
\tbandn{8}{dsband}{dsbtx}{Generator: name the quantity}
Hamiltonian, SymODEN & scalar generator & \mkno & \mkpart & \mkpart & \mkno & \mkyes & diagonal form \\
Deep conservation & constraint set & \mkpart & \mkpart & \mkpart & \mkno & \mkyes & diagonal form \\
\tbandn{8}{dsband}{dsbtx}{Operator learning: correct the output}
FNO & none & \mkno & \mkno & \mkno & \mkno & \mkno & conserved quantity \\
CPFNO & learned corrector & \mkno & \mkpart & \mkno & \mkno & \mkpart & conserved quantity \\
\midrule
\rowcolor{oursrow}[\tabcolsep][\tabcolsep]
\textbf{KC (ours)} & generator, $\Lop\otimes\Iden{m}$ & \mkyes & \mkyes & \mkyes & \mkyes & \mkyes & coupling values \\
\bottomrule
\end{tabular}
\vspace{-10pt}
\end{table}

\noindent\textbf{Stability and conservation in learned dynamics.}\;
Diagonal transitions have made long-sequence modelling practical \citep{gu2022efficiently,
gupta2022diagonal, gu2022parameterization, smith2023simplified, gu2024mamba, dao2024mamba2},
and stability has been secured by restricting the operator to the unitary manifold
\citep{arjovsky2016unitary, wisdom2016full, helfrich2018orthogonal, lezcano2019cheap} or the
eigenvalues to the unit disk \citep{orvieto2023resurrecting}. A second line places the conserved
quantity in the generator, as a constraint set \citep{lee2021deep} or a learned Hamiltonian
\citep{greydanus2019hamiltonian, cranmer2020lagrangian, chen2020symplectic, zhong2020symoden}.
What neither line constrains is the modal basis, which is inherited from the encoder in every
design of Table~\ref{tab:layer_constraint}, so a spectral constraint holds the norm alone, a
named constraint holds only what is named, and the quantities that depend on the coordinates
are left free. We fix the basis in which the spectrum acts, so that conservation follows from
the assembly rather than from the values placed in it.

\noindent\textbf{Graph Laplacians and conserved quantities.}\;
Zero row sums make the constant vector an eigenvector of a graph Laplacian at eigenvalue zero,
so any sum over nodes is preserved by the associated flow. That identity is the object that
carries the invariant in reaction network theory \citep{veerman2023chemical} and in average
consensus \citep{olfatisaber2007consensus}. Within machine learning, the same graph instead
carries a learned message function \citep{battaglia2016interaction, sanchez2020learning,
satorras2021en, kipf2018neural}, which fixes the sparsity pattern of the dynamics but leaves
the conserved quantity free, and the identity itself has not been carried into a recurrence. We
assemble the Laplacian as the generator of a diagonal transition, so that the zero mode fixes
which quantity the recurrence preserves at every parameter value.

\noindent\textbf{Structure preservation under long integration.}\;
The loss of a conserved quantity under long integration is a recurring difficulty across
scientific computing \citep{hairer2006geometric}, and each field has developed its own repair.
In molecular dynamics, velocity rescaling drains kinetic energy into centre-of-mass motion
until the internal degrees of freedom freeze \citep{harvey1998flyingicecube}. In stiff kinetics,
a mass-conserving scheme leaves the positive orthant and is projected back onto the reaction
simplex \citep{sandu2001positive}. In climate emulation, a water-budget corrector has been shown
to hide a degrading model \citep{chapman2026correctors}, a failure mode we observe in
\S\ref{sec:experiments}. In each case, the violation is corrected after it occurs, because no
condition on the operator excludes it in advance. We give that condition in \S\ref{sec:method},
and extended related work is provided in Appendix~\ref{app:related}.

%% file: S3_method.tex
\section{Method}
\label{sec:method}
\setlength{\abovedisplayskip}{3pt}\setlength{\belowdisplayskip}{3pt}
\setlength{\abovedisplayshortskip}{1pt}\setlength{\belowdisplayshortskip}{3pt}

\subsection{Problem formulation}
\label{sec:formulation}

We study long-horizon rollout in a diagonal state space model, where the same transition is
applied at every step and any quantity the transition does not preserve drifts with the horizon
\citep{gu2022parameterization, smith2023simplified}. By diagonal we mean that each step acts as
a diagonal matrix in a modal basis, which for the construction below depends on the state and
is recomputed at every step. Let an encoder map an observation $x_t$ to a latent
$z_t\in\mathbb{C}^{d}$, partitioned into $N$ object subspaces $z=(z_1,\dots,z_N)$ with
$z_i\in\mathbb{C}^{m}$ and $d=Nm$. Given the latent and an action $a_t$, we seek a transition $U$
that advances the latent and holds a set $\mathcal{Q}$ of functions of $z$ fixed,
\begin{equation}
\label{eq:req}
z_{t+1}=U(z_t,a_t)\,z_t,
\qquad
Q\bigl(U(z,a)\,z\bigr)=Q(z)
\quad\text{for all } Q\in\mathcal{Q},\ z,\ a,\ \text{and all }\theta .
\end{equation}
The transition should satisfy (i) exactness, so that Equation~\ref{eq:req} holds at every
parameter value $\theta$ rather than at a minimiser of a loss, and (ii) completeness, so that
$\mathcal{Q}$ follows from the algebraic form of $U$ rather than from a list named in advance. We
follow three members of $\mathcal{Q}$ throughout,
\begin{equation}
\label{eq:qthree}
\|z\|^{2},\qquad \textstyle\sum_{i=1}^{N} z_i,\qquad
\operatorname{Im}\bigl(z^\dagger\mathbf{J}z\bigr),
\end{equation}
the latent norm, the sum over subspaces, and an angular pairing generated by
$\mathbf{J}=\operatorname{diag}(\mathbf{J}_2,\dots,\mathbf{J}_2)$, which arise from the unitary
group, from translation $z_i\mapsto z_i+\epsilon$, and from rotation respectively. To meet (i) and
(ii), we assemble $U$ from a generator whose algebraic form fixes $\mathcal{Q}$
(\S\ref{sec:assembly}), characterise $\mathcal{Q}$ exactly (\S\ref{sec:family}), and state where
the assembly applies (\S\ref{sec:scope}).

\subsection{The Kronecker--Cayley assembly}
\label{sec:assembly}

\begingroup
\makeatletter
\setlength{\thm@preskip}{4pt}\setlength{\thm@postskip}{4pt}
\makeatother
\begin{theorem}[Characterisation of the conserved set]
\label{thm:commutant}
Let $\Ham$ be Hermitian and $U=\Cay(\Ham)$. For a Hermitian $G$, the quadratic form
$Q_G(z)=z^\dagger Gz$ satisfies
\begingroup
\setlength{\abovedisplayskip}{3pt}\setlength{\belowdisplayskip}{3pt}\setlength{\fboxsep}{1.5pt}
\begin{equation}
\label{eq:thmcomm}
\bluebox{\ Q_G(Uz)=Q_G(z)\ \text{ for every } z
\quad\Longleftrightarrow\quad
[\Ham,G]=0 .\ }
\end{equation}
\endgroup
\end{theorem}
\endgroup

\vspace{-10pt}

\noindent
By Theorem~\ref{thm:commutant}, a quantity is conserved exactly when its generator commutes
with $\Ham$, so we make the transition conserve by assembling $\Ham$ rather than by fitting it,
and the three quantities of Equation~\ref{eq:qthree} translate into three requirements on
$\Ham$, each stronger than the last. For the norm, it suffices that $\Ham$ be Hermitian, since
$\Iden{}$ then commutes with $\Ham$. For the sum, $\Ham$ must in addition annihilate
$\onevec\otimes e$ for every $e$, which puts the uniform direction at eigenvalue one under the
transform. For the angular pairing, it suffices that every block of $\Ham$ be a scalar multiple
of $\Iden{m}$, which makes every $\Iden{N}\otimes C$ commute with $\Ham$ and in particular
$\Iden{N}\otimes\mathbf{J}$. Three stages meet these in turn, and only the first is learned.

\noindent
\begin{minipage}[t]{0.47\textwidth}
\vspace{0pt}
Algorithm~\ref{alg:kc} summarises the rollout. Line~1 encodes and partitions the observation.
Inside the loop, line~3 evaluates the couplings of Equation~\ref{eq:coupling} and is the only
line that depends on a parameter. Line~4 assembles them into $\Lop$ by
Equation~\ref{eq:generator}, line~5 applies the Cayley transform by Equation~\ref{eq:cayley},
and line~6 advances the latent, which Theorem~\ref{thm:kcd} shows is diagonal in the modal
coordinates of $\Lop$ and which \S\ref{sec:family} shows leaves every member of $\mathcal{Q}$
unchanged. The batched form is provided in Appendix~\ref{app:batched}.
\end{minipage}\hfill
\begin{minipage}[t]{0.50\textwidth}
\vspace{0pt}
\small
\refstepcounter{algorithm}\label{alg:kc}
\hrule\vspace{2pt}
\textbf{Algorithm~\thealgorithm} Kronecker--Cayley rollout
\vspace{2pt}\hrule\vspace{2pt}
\begin{algorithmic}[1]
\Require $x_0$, $a_{0:T-1}$, encoder $\Phi$, network $\mathrm{MLP}_\theta$
\State $z_0\gets\Phi(x_0)$, split as $(z_{0,1},\dots,z_{0,N})$
\For{$t=0,\dots,T-1$}
  \State \textcolor{lblgreen}{$w_{ij}\gets\mathrm{MLP}_\theta(\|z_{t,i}-z_{t,j}\|^2,a_t)$}
    \Comment{Eq.~\ref{eq:coupling}}
  \State \textcolor{lblblue}{$\Lop_{ij}\gets-w_{ij}$, $\Lop_{ii}\gets\textstyle\sum_{k\neq i}w_{ik}$}
    \Comment{Eq.~\ref{eq:generator}}
  \State \textcolor{lblblue}{$\mathbf{V}\gets\Cay(\Lop)$}
    \Comment{Eq.~\ref{eq:cayley}}
  \State \textcolor{lblblue}{$z_{t+1}\gets(\mathbf{V}\otimes\Iden{m})\,z_t$}
    \Comment{Thm.~\ref{thm:kcd}}
\EndFor
\State \Return $z_{1:T}$
\end{algorithmic}
\vspace{1pt}\hrule
\end{minipage}
\vspace{2pt}

\noindent\textbf{Learned coupling for input dependence.}\;
A transition fixed in advance cannot represent dynamics that depend on the state and the action.
We parameterise that dependence by one real scalar per pair of objects,
\begingroup
\setlength{\abovedisplayskip}{3pt}\setlength{\belowdisplayskip}{3pt}\setlength{\fboxsep}{1.5pt}
\begin{equation}
\label{eq:coupling}
\underbrace{\greenbox{\ w_{ij}=\mathrm{MLP}_\theta\bigl(\|z_i-z_j\|^2,\ a_t\bigr)\ }}
  _{\glbl{$\binom{N}{2}$ scalars, the only learned part}}
\qquad
\underbrace{\greenbox{\ w_{ij}=w_{ji}\ }}_{\glbl{one value shared per pair}}
\end{equation}
\endgroup
so the coupling varies with the input while its placement is fixed by the next two stages.

\noindent\textbf{Laplacian assembly for vanishing row sums.}\;
A free Hermitian generator built from the same scalars would yield a unitary step, but its
conserved set would move with $\theta$. We place the couplings in a graph Laplacian over the $N$
subspaces and tensor it with the identity,
\begingroup
\setlength{\abovedisplayskip}{3pt}\setlength{\belowdisplayskip}{3pt}\setlength{\fboxsep}{1.5pt}
\begin{equation}
\label{eq:generator}
\Lop_{ij}=
\begin{cases}
\underbrace{\greenbox{\ -w_{ij}\ }}_{\glbl{learned}}, & i\neq j,\\[4pt]
\underbrace{\bluebox{\ \textstyle\sum_{k\neq i} w_{ik}\ }}_{\blbl{set by the row}}, & i=j,
\end{cases}
\qquad\text{giving}\qquad
\underbrace{\bluebox{\ \Ham=\Lop\otimes\Iden{m}\ }}_{\blbl{block-scalar, zero row sums}}
\end{equation}
\endgroup
for a symmetric Laplacian $\Lop\in\mathbb{R}^{N\times N}$. Real symmetric entries make $\Ham$
Hermitian, each diagonal entry cancels its own row so that $\Lop\onevec=0$, and every block is a
multiple of $\Iden{m}$. The three requirements are met by the pattern in
Equation~\ref{eq:generator} rather than by its values, so they hold at every $\theta$.

\begingroup
\setlength{\abovedisplayskip}{3pt}\setlength{\belowdisplayskip}{3pt}\setlength{\fboxsep}{1.5pt}
\begin{equation}
\label{eq:cayley}
\underbrace{\bluebox{\ U=\Cay(\Ham)=\bigl(\Iden{}+\tfrac{\mathrm{i}}{2}\Ham\bigr)^{-1}
 \bigl(\Iden{}-\tfrac{\mathrm{i}}{2}\Ham\bigr)\ }}
 _{\blbl{unitary for every $\Ham=\Ham^\dagger$}}
\qquad
\underbrace{\bluebox{\ U=\mathbf{V}\otimes\Iden{m}\ }}
 _{\blbl{$\mathbf{V}=\Cay(\Lop)\in\mathcal{U}(N)$}}
\end{equation}
\endgroup

\begingroup
\makeatletter
\setlength{\thm@preskip}{4pt}\setlength{\thm@postskip}{4pt}
\makeatother
\begin{theorem}[Kronecker--Cayley decomposition]
\label{thm:kcd}
Let $\Ham=\Lop\otimes\Iden{m}$ with $\Lop=P\Lambda P^{\!\top}$ a symmetric graph Laplacian, and
let $\zmodal=(P^{\!\top}\otimes\Iden{m})z$. Then
\begingroup
\setlength{\abovedisplayskip}{3pt}\setlength{\belowdisplayskip}{3pt}\setlength{\fboxsep}{1.5pt}
\begin{equation}
\label{eq:thmkcd}
\bluebox{\ \zmodal^{(k)}_{t+1}=\dk\,\zmodal^{(k)}_{t},\qquad
\dk=\frac{1-\mathrm{i}\lk/2}{1+\mathrm{i}\lk/2},\qquad
|\dk|=1,\qquad d_{\kzero}=1 .\ }
\end{equation}
\endgroup
\end{theorem}
\endgroup

\noindent
Here $\mathrm{i}=\sqrt{-1}$. Equation~\ref{eq:cayley} rotates the $N$ objects into
one another and leaves every coordinate channel untouched, and by Theorem~\ref{thm:kcd} the
rotation is diagonal in the eigenbasis of $\Lop$. Since $\Lop$ depends on $z_t$ through the
couplings, that eigenbasis changes from one step to the next, and the two modal transforms in
Algorithm~\ref{alg:kc} are the cost of the change. A diagonal state space model with
input-dependent entries, as in Mamba, instead keeps the basis fixed, $P_t=P$. The step costs one
$N\times N$ eigendecomposition and two modal transforms, $\cost$ against $\costfull$, and the
coupling network carries $3{,}963$ parameters against roughly $15$M in the shared encoder and
decoder, so the guarantee is not obtained through capacity on the transition. The three stages
impose three structural properties on $\Ham$ at every parameter value, and \S\ref{sec:family}
derives the quantities these properties conserve and identifies the boundary of the conserved
set.

\subsection{The invariant family}
\label{sec:family}

We now derive what the three properties of \S\ref{sec:assembly} conserve. Two results follow
from Theorem~\ref{thm:commutant}, the first arranging the three quantities of
Equation~\ref{eq:qthree} into nested layers by the condition each one requires, the second
identifying the full conserved set and showing that nothing lies outside it.

\begingroup
\makeatletter
\setlength{\thm@preskip}{4pt}\setlength{\thm@postskip}{4pt}
\makeatother
\begin{corollary}[Layered conservation]
\label{cor:layered}
Let $U$ be a unitary diagonal transition on $N$ subspaces. Then
\begingroup
\setlength{\abovedisplayskip}{3pt}\setlength{\belowdisplayskip}{3pt}\setlength{\fboxsep}{1.5pt}
\begin{equation}
\label{eq:layers}
\underbrace{\bluebox{\ |\dk|=1\ }}_{\blbl{$\|z\|^2$}}
\ \Leftarrow\
\underbrace{\bluebox{\ \Lop\onevec=0\ }}_{\blbl{$+\ \textstyle\sum_i z_i$}}
\ \Leftarrow\
\underbrace{\bluebox{\ \Ham=\Lop\otimes\Iden{m}\ }}_{\blbl{$+\ \operatorname{Im}(z^\dagger\mathbf{J}z)$}}
\end{equation}
\endgroup
where each condition implies the one to its left and no converse holds. A transition satisfying
only the leftmost condition preserves $\|z\|^2$ and leaves $\sum_i z_i$ unconstrained.
\end{corollary}
\endgroup

\noindent
Corollary~\ref{cor:layered} is falsifiable, since removing one condition should cost exactly the
quantity that condition holds, leaving the others intact, and \S\ref{sec:experiments} confirms
this by removing each condition in turn.

\begingroup
\makeatletter
\setlength{\thm@preskip}{4pt}\setlength{\thm@postskip}{4pt}
\makeatother
\begin{corollary}[The invariant family]
\label{cor:family}
Let $U=\Cay(\Lop\otimes\Iden{m})$ with $\Lop$ ranging over all symmetric $N\times N$ matrices
with zero row sums, $N\ge2$. The quadratic forms held at every such $\Lop$ are exactly
\begingroup
\setlength{\abovedisplayskip}{3pt}\setlength{\belowdisplayskip}{3pt}\setlength{\fboxsep}{1.5pt}
\begin{equation}
\label{eq:family}
\bluebox{\ Q_{C_1,C_2}(z)=z^\dagger\bigl(\Iden{N}\otimes C_1+\onevec\onevec^{\!\top}\otimes C_2\bigr)z,
\qquad C_1,C_2\in\mathbb{C}^{m\times m}\ \text{Hermitian},\ }
\end{equation}
\endgroup
a real vector space of dimension $2m^2$. No quadratic form outside Equation~\ref{eq:family} is
held at every $\Lop$.
\end{corollary}
\endgroup

\noindent
Equation~\ref{eq:family} is the set $\mathcal{Q}$ of Equation~\ref{eq:req} and meets
requirement (ii) of \S\ref{sec:formulation}. It contains $\|z\|^2$ as $(C_1,C_2)=(\Iden{m},0)$
and $\operatorname{Im}(z^\dagger\mathbf{J}z)$ as $(\mathbf{J},0)$, and the summed quantity
is linear rather than quadratic and is held through the kernel of $\Ham$. The assembled operator
holds this family and nothing else at every parameter value. Which physical quantities lie
inside the family, and on which dynamics the assembly applies, we address in \S\ref{sec:scope}.

\subsection{Where the construction applies}
\label{sec:scope}

\begingroup
\makeatletter
\setlength{\thm@preskip}{4pt}\setlength{\thm@postskip}{4pt}
\makeatother
\begin{corollary}[Applicability]
\label{cor:applic}
Let the decoder be linear on the latent, $y=Dz$. A physical quantity $q(y)$ that is a quadratic
form in $y$ is held by the construction for every $\theta$ if and only if there exist Hermitian
$C_1,C_2$ with
\begingroup
\setlength{\abovedisplayskip}{3pt}\setlength{\belowdisplayskip}{3pt}\setlength{\fboxsep}{1.5pt}
\begin{equation}
\label{eq:applic}
\bluebox{\ q(Dz)=z^\dagger\bigl(\Iden{N}\otimes C_1+\onevec\onevec^{\!\top}\otimes C_2\bigr)z
\quad\text{for every } z .\ }
\end{equation}
\endgroup
The characterisation does not determine whether a nonquadratic physical quantity is conserved.
\end{corollary}
\endgroup

\begingroup
\makeatletter
\setlength{\thm@preskip}{4pt}\setlength{\thm@postskip}{4pt}
\makeatother
\begin{proposition}[Lower bound for any unitary step]
\label{prop:bound}
For every unitary $U$ and every pair of consecutive latents $(z_t,z_{t+1})$,
\begingroup
\setlength{\abovedisplayskip}{3pt}\setlength{\belowdisplayskip}{3pt}\setlength{\fboxsep}{1.5pt}
\begin{equation}
\label{eq:bound}
\bluebox{\ \|Uz_t-z_{t+1}\|\;\ge\;\bigl|\,\|z_{t+1}\|-\|z_t\|\,\bigr| .\ }
\end{equation}
\endgroup
\end{proposition}
\endgroup

\noindent
Corollary~\ref{cor:applic} settles which physical quantities are held before any model is
trained. A linear momentum $\sum_i m_i v_i$ is linear in the state and is held through the
kernel. An angular momentum $\sum_i r_i\times m_i v_i$ is bilinear in the state and therefore a
quadratic form. A potential energy of the form $\sum_{i<j}\|r_i-r_j\|^{-1}$ is not a quadratic
form under any linear encoding, so Corollary~\ref{cor:applic} does not apply to it, and
\S\ref{sec:experiments} reports energy as held only where the potential is quadratic.
Proposition~\ref{prop:bound} is the reverse triangle inequality with $\|Uz_t\|=\|z_t\|$. The
bound depends only on the encoded trajectory, so no choice of $\Lop$, $\theta$ or decoder
reduces it. A unitary step thus applies to dynamics whose encoded norm is constant along
the true trajectory and incurs Equation~\ref{eq:bound} as a per-step error on dissipative
dynamics. \S\ref{sec:experiments} tests the quantity boundary on three $N$-body systems and
the dynamics boundary on a dissipative system.

\begingroup
\makeatletter
\setlength{\thm@preskip}{4pt}\setlength{\thm@postskip}{4pt}
\makeatother
\begin{proposition}[Conservation of the summed quantity]
\label{prop:measure}
A unitary diagonal transition on $N$ subspaces is determined by a modal basis and a spectrum,
$U=\bigl(Q\operatorname{diag}(d)Q^{\!\top}\bigr)\otimes\Iden{m}$ with $(Q,d)\in
O(N)\times(S^1)^N$. Let $c=Q^{\!\top}\uniform$. The transition preserves $\sum_i z_i$ for every
$z$ if and only if
\begingroup
\setlength{\abovedisplayskip}{3pt}\setlength{\belowdisplayskip}{3pt}\setlength{\fboxsep}{1.5pt}
\begin{equation}
\label{eq:conservingset}
\bluebox{\ (\dk-1)\,c_k=0\quad\text{for every }k .\ }
\end{equation}
\endgroup
Every mode that rotates carries none of the uniform vector, and every mode that carries it is
stationary.
\end{proposition}
\endgroup

\noindent
We note that Equation~\ref{eq:conservingset} has two kinds of solutions. The assembly satisfies it
with $c=e_{\kzero}$ and $d_{\kzero}=1$, since $\Lop\onevec=0$ puts $\uniform$ in the null mode
before any parameter is set, and leaves the remaining $N-1$ modes free to rotate. A transition
may instead satisfy it with every $\dk=1$, at which point it no longer evolves. A basis
initialised away from Equation~\ref{eq:conservingset} and trained under a conservation penalty
must move $c$ or $d$ to one of these solutions, and we confirm in \S\ref{sec:experiments} that
the seeds which conserve reach the second solution, freezing every mode, while the basis remains
near its initialisation on every seed under both initialisations tested.

%% file: S4_experiments_10.14.tex
\section{Experiments}
\label{sec:experiments}

We conduct experiments to answer five core research questions: \textbf{Q1.} How well does the
Kronecker--Cayley transition hold a conserved quantity over a long rollout, against diagonal
transitions that constrain only the spectrum? \textbf{Q2.} Can a corrector applied to the output
restore what the operator itself does not hold? \textbf{Q3.} Is each condition of the assembly
necessary, and does removing one cost exactly the quantity it holds? \textbf{Q4.} Can a
conservation penalty bring a trained basis to the condition the assembly satisfies by
construction? \textbf{Q5.} Does the same layering appear in dynamical
systems from physics, chemistry and engineering, and where does the construction cease to apply?

\begin{table}[!t]
\centering\footnotesize
\setlength{\tabcolsep}{5pt}
\renewcommand{\arraystretch}{0.95}
\setlength{\aboverulesep}{0.15ex}\setlength{\belowrulesep}{0.25ex}
\caption{\textbf{Conservation and prediction on four grid benchmarks, seeds $42$ to $44$.} Mass
drift $|\Delta M|\!\downarrow$ from the first decoded frame to the final horizon in absolute
units, as a geometric mean over seeds, and nRMSE$\downarrow$ at that horizon as an arithmetic
mean. Best learned value in bold, second underlined, dagger at seed $42$.}
\label{tab:q1}
\begin{tabular}{@{}l@{} >{\hspace{8pt}}r@{}>{\hspace{5pt}}r @{} >{\hspace{11pt}}r@{}>{\hspace{5pt}}r @{} >{\hspace{11pt}}r@{}>{\hspace{5pt}}r @{} >{\hspace{11pt}}r@{}>{\hspace{5pt}}r@{}}
\toprule
& \multicolumn{2}{c}{\textbf{Dam-break}} & \multicolumn{2}{c}{\textbf{Advection}} & \multicolumn{2}{c}{\textbf{Allen--Cahn}} & \multicolumn{2}{c}{\textbf{Sphere}} \\
\textbf{Model} & $|\Delta M|$ & nRMSE & $|\Delta M|$ & nRMSE & $|\Delta M|$ & nRMSE & $|\Delta M|$ & nRMSE \\
\midrule
\textit{reference solver} & $0$ & --- & $10^{-6}$ & --- & $10^{-16}$ & --- & $30.8$ & --- \\
\textit{constant field} & $0$ & $0.103$ & $0$ & $0.702$ & $0$ & $0.999$ & $0$ & $1.00$ \\
\tbandn{9}{dsband}{dsbtx}{Diagonal recurrences, shared encoder and decoder}
level B & $\second{27.7}$ & $0.226$ & $477$ & $1.10$ & $4.70$ & $0.813$ & $2.7\mathrm{e}4$ & $1.68$ \\
level C & $41.9$ & $0.227$ & $409$ & $1.02$ & $3.28$ & $\second{0.811}$ & $2.4\mathrm{e}4$ & $1.62$ \\
free & $286$ & $\second{0.156}$ & $421$ & $\second{1.00}$ & $53.9$ & $\best{0.707}$ & $1.9\mathrm{e}{11}$ & div. \\
LRU & $2.3\mathrm{e}3$ & $0.992$ & $1.8\mathrm{e}3$ & $1.74$ & $18.6$ & $1.07$ & $6.7\mathrm{e}4$ & $3.01$ \\
\tbandn{9}{dsband}{dsbtx}{Neural operators, with and without a mass corrector}
FNO & $3.6\mathrm{e}6^{\dagger}$ & $1.8\mathrm{e}3^{\dagger}$ & $9.3\mathrm{e}3$ & $63.2$ & $61.0$ & $2.20$ & $2.4\mathrm{e}7^{\dagger}$ & div.$^{\dagger}$ \\
CPFNO, uncorrected & $4.4\mathrm{e}{18}^{\dagger}$ & $\infty^{\dagger}$ & $396$ & $1.85$ & $171$ & $1.83$ & $8.3\mathrm{e}4^{\dagger}$ & $1.29^{\dagger}$ \\
CPFNO, corrected & $1.3\mathrm{e}6^{\dagger}$ & $\infty^{\dagger}$ & $\best{2.8\mathrm{e}{-4}}$ & $1.85$ & $\second{2.7\mathrm{e}{-4}}$ & $1.83$ & $1.8\mathrm{e}4^{\dagger}$ & $1.29^{\dagger}$ \\
\midrule
\rowcolor{oursrow}[0pt][0pt]
KC, unconstrained decoder & $115$ & $0.221$ & $7.25$ & $\best{0.727}$ & $33.4$ & $2.07$ & $\second{785}$ & $\second{1.12}$ \\
\rowcolor{oursrow}[0pt][0pt]
\textbf{KC} & $\best{8.3\mathrm{e}{-3}}$ & $\best{0.146}$ & $\second{2.6\mathrm{e}{-3}}$ & $1.79$ & $\best{2.0\mathrm{e}{-5}}$ & $0.961$ & $\best{483}$ & $\best{1.05}$ \\
\bottomrule
\end{tabular}

\vspace{5pt}
\includegraphics[width=\linewidth]{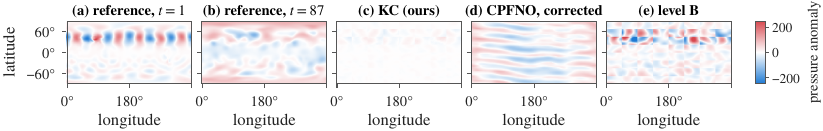}
\captionof{figure}{\textbf{Pressure anomaly on the sphere, seed $42$.} The reference at the
input frame and at the final horizon, beside the decoded fields of KC, of CPFNO after its
corrector and of level~B, on one colour scale. The field of KC stays near the mass-matched
constant, whereas the corrected output keeps a zonal structure and drifts.}
\label{fig:sphere-fields}
\vspace{-22pt}
\end{table}

\subsection{Q1: Mass Is Held Where Every Diagonal Baseline Loses It}
\label{sec:exp-q1}

\noindent\textbf{Setup.}\;
We test Corollary~\ref{cor:layered}, which predicts that reaching only the outermost layer
holds the norm and loses the summed quantity, on the dam-break and advection equations of
PDEBench \citep{takamoto2022pdebench}, the spherical shallow-water system of PDEArena
\citep{gupta2022pdearena} and a conservative Allen--Cahn flow. To isolate the transition, we
fix the encoder and decoder of KC across every baseline, so that the levels reproduce the
layers of the corollary. Level~A leaves the spectrum free, level~B imposes unit modulus, and
level~C fixes one mode per object at unity in the coordinate basis. To separate the operator
from the read-out, we add an LRU-style recurrence \citep{orvieto2023resurrecting} and KC with
an unconstrained decoder, which breaks the tie required by Corollary~\ref{cor:applic}, and the
external references FNO \citep{li2021fourier} and CPFNO \citep{liu2025cpfno}. We therefore
measure conservation as the drift $|Q(T)-Q(1)|$ from the first decoded frame rather than from
the input, which removes the encoder--decoder round trip, and prediction error as
$\|\hat{y}_T-y_T\|/\|y_T\|$. Every run is repeated at seeds $42$, $43$ and $44$, and datasets
are given in Appendix~\ref{app:glossary}.

\begingroup\looseness=-1
\noindent\textbf{Results.}\;
Total mass is held by KC at the round-off level of the float32 rollout on the three flat
benchmarks and within a discretisation residual of the reference on the sphere, whereas every
diagonal baseline loses it by two to nine orders of magnitude: the drift of KC is
$2.0\times10^{-5}$ on Allen--Cahn, $2.6\times10^{-3}$ on advection and $8.3\times10^{-3}$ on the
dam break, against $3$ to $2\times10^{11}$ for every other diagonal transition
(Table~\ref{tab:q1}). On the sphere the residual of $483$ against $30.8$ is the variation of
$\cos(\text{lat})$ within a patch (Appendix~\ref{app:exp-sphere}). What remains is numerical
rather than learned, since it accumulates smoothly with the rollout step and is largest on
advection, whose horizon is twice that of the other benchmarks (Appendix~\ref{app:exp-q1}). The
guarantee reaches the output only through the decoder: an unconstrained decoder holds the latent
invariant just as tightly, between $10^{-6}$ and $10^{-2}$ under both read-outs, while the
physical mass it emits drifts by $7$ to $115$ on the three flat benchmarks, three to six orders
above the tied decoder. Mass correction reaches a lower drift than KC on advection,
$2.8\times10^{-4}$ against $2.6\times10^{-3}$, on a field whose nRMSE of $1.85$ is worse than
the constant field at $0.702$, which is the failure the next question examines. Conservation and
accuracy are separate, as Figure~\ref{fig:sphere-fields} shows: the field of KC lies near the
mass-matched constant, and no model predicts the evolving field better than that constant under
the shared per-patch encoder (Appendix~\ref{app:exp-diag}).\par\endgroup


\subsection{Q2: Correction at the Output Does Not Replace the Condition}
\label{sec:exp-q2}

\begingroup\looseness=-1
\noindent\textbf{Setup.}\;
We compare two correctors against the assembly on the dam break over five seeds. The first is
the corrector of CPFNO \citep{liu2025cpfno}, $\hat u+(m_0-m(\hat u))A$, applied inside the
forward pass with a learned weight $A$ and trained end-to-end on the corrected output, and we
log the prediction before and after it at every step so that the two can be read separately.
The second adds the missing mass uniformly over the grid at inference on top of FNO, without
training, which separates the correction from the fit that precedes it. We count a seed as
diverged at the first step where nRMSE exceeds $100$, after which it enters the median as
infinite, so that divergence is not hidden by the seeds that survive.\par\endgroup

\begin{figure}[!t]
\centering
\captionsetup{skip=0pt}
\includegraphics[width=\linewidth]{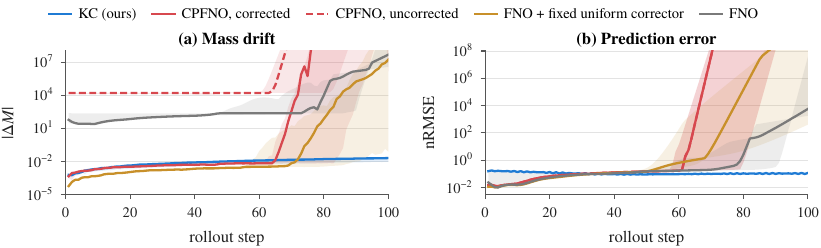}\\[2pt]
\includegraphics[width=\linewidth]{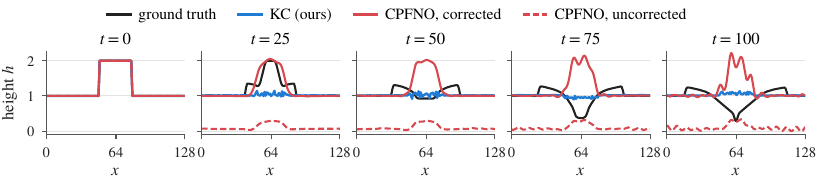}
\caption{\textbf{Mass correction on the dam-break benchmark.} Top: \textbf{(a)} mass drift and
\textbf{(b)} prediction error over a $100$-step rollout, median over five seeds with the
interquartile range shaded. Bottom: water height along the central row at five steps, seed $42$.
Per-seed values are in Tables~\ref{tab:app-q2-seeds} and \ref{tab:app-additive}.}
\label{fig:chapman}
\vspace{-16pt}
\end{figure}

\begingroup\looseness=-1
\noindent\textbf{Results.}\;
Correction restores the scalar and not the field, and it does neither once the prediction it
corrects diverges. What it does achieve is large: the corrected CPFNO output holds mass near
$10^{-3}$ for fifty steps while the prediction before correction drifts by $10^{4}$ from the
first step (Figure~\ref{fig:chapman}a), so seven orders of magnitude are closed at every step.
What the drift curve does not show is visible in the cross-sections, where the uncorrected
prediction lies at $h\approx0.06$ across the domain and the corrected output, with its mass
added back, retains a bump at the centre while the true solution has spread into a ring. The
same split appears wherever the prediction stays bounded, and on advection and Allen--Cahn
correction drives the drift to $3\times10^{-4}$ on fields whose error of $1.8$ exceeds that of a
constant field (Table~\ref{tab:q1}). Once the prediction diverges the corrector goes with it,
with four seeds out of five diverging between steps $48$ and $83$ and the fixed corrector diverging
on every seed between steps $58$ and $97$ (Figure~\ref{fig:chapman}b), and KC is the only model
that reaches the final horizon with bounded error under every seed. On the one seed where FNO
reaches that horizon it is twice as accurate as KC and drifts its mass by $230$, which is
accuracy without the guarantee. The same failure mode has been reported for a
climate emulator \citep{chapman2026correctors}. The condition must therefore be encoded in the
operator, which leaves open which condition carries which quantity.\par\endgroup


\begin{table}[!t]
\centering\footnotesize
\setlength{\tabcolsep}{15pt}
\renewcommand{\arraystretch}{0.95}
\setlength{\aboverulesep}{0.15ex}\setlength{\belowrulesep}{0.25ex}
\caption{\textbf{Which condition holds which quantity.} Two controls, each removing one
condition of Corollary~\ref{cor:layered} while keeping the others. Drift in percent, where a
value near $10^{-6}$ is the floor of the float32 measurement, with the per-seed values of both
controls in Appendix~\ref{app:exp-q3}.}
\label{tab:q3}
\begin{tabular}{l ccc cc}
\toprule
& \multicolumn{3}{c}{\textbf{Condition present}} & \multicolumn{2}{c}{\textbf{Drift (\%)}} \\
\cmidrule(lr){2-4}\cmidrule(lr){5-6}
\textbf{Transition} & $|\dk|{=}1$ & $\Lop\onevec{=}0$ & $\Lop{\otimes}\Iden{m}$ & energy / norm & momentum \\
\midrule
\tbandn{6}{dsband}{dsbtx}{Matched perturbation of the trained Laplacian, $\varepsilon=0.5$, step 25, mean over three seeds}
row sums kept & \mkyes & \mkyes & \mkyes & $\best{1.8\mathrm{e}{-6}}$ & $\best{3.5\mathrm{e}{-5}}$ \\
row sums broken & \mkyes & \mkno & \mkyes & $\best{2.0\mathrm{e}{-6}}$ & $110$ \\
\tbandn{6}{dsband}{dsbtx}{Norm-preserving operators, operator level, 2000 steps, mean over three draws}
Haar unitary & \mkyes & \mkno & \mkno & $3.5\mathrm{e}{-3}$ & $\second{169.7}$ \\
Cayley, dense & \mkyes & \mkno & \mkno & $1.2\mathrm{e}{-3}$ & $170.7$ \\
diagonal unitary & \mkyes & \mkno & \mkno & $\second{2.5\mathrm{e}{-4}}$ & $172.5$ \\
\rowcolor{oursrow}[\tabcolsep][\tabcolsep]
KC (ours) & \mkyes & \mkyes & \mkyes & $\best{3.7\mathrm{e}{-5}}$ & $\best{4.0\mathrm{e}{-3}}$ \\
\bottomrule
\end{tabular}

\vspace{3pt}
\includegraphics[width=\linewidth]{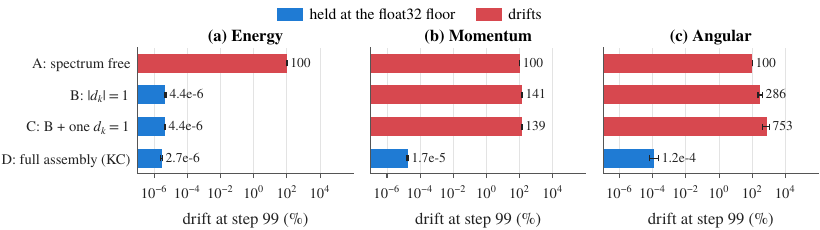}
\captionof{figure}{\textbf{Drift of three quantities under four levels of the assembly.} Visual
$N$-body simulator, step $99$, mean over three seeds with min--max bars, one panel per quantity.
Blue marks a drift at the float32 floor and red one above it. Unit modulus closes the energy
panel at level~B, whereas the summed and the angular quantity stay open until the full assembly
closes them at level~D. Per-seed values are in Table~\ref{tab:app-q3-levels}.}
\label{fig:layering}
\vspace{-10pt}
\end{table}

\subsection{Q3: Each Condition Holds Exactly One Quantity}
\label{sec:exp-q3}

\noindent\textbf{Setup.}\;
We test Corollary~\ref{cor:layered} by removing one condition at a time, since it holds only if
each removal costs exactly the quantity that condition carries and leaves the others at the
floor. We first train the four levels of \S\ref{sec:assembly} on a visual $N$-body simulator,
which carries all three quantities at once. We then perturb the trained Laplacian by a symmetric
draw of scale $\varepsilon$, once with the diagonal corrected so that $\Lop\onevec=0$ still
holds after the change and once with the same draw left uncorrected, which separates the
row-sum condition from the coupling it travels with. We finally compare four norm-preserving
operators at the operator level, where no encoder or decoder is involved. Each control is
repeated three times, over seeds where a model is trained and over draws at the operator level.

\begingroup\looseness=-1
\noindent\textbf{Results.}\;
The three controls separate the three conditions, and each removal costs exactly the quantity
that condition holds (Table~\ref{tab:q3} and Figure~\ref{fig:layering}). Energy is held at the
float32 floor from level~B onward, whereas momentum and the angular pairing are held only at
level~D. Unit modulus alone holds the norm to $4\times10^{-6}$ and leaves momentum at $141$
percent and the angular pairing at $286$, and fixing one mode per object at unity in the
coordinate basis holds neither, since Equation~\ref{eq:conservingset} requires the uniform
vector to lie in a stationary mode, which the coordinate modes do not provide. The perturbation
isolates the row sums: a change of up to a quarter of the operator's Frobenius norm costs
nothing when they are restored and $110$ percent of the summed quantity when they are not, with
energy at $2\times10^{-6}$ in both cases at every $\varepsilon$ (Table~\ref{tab:perturb}). At
the operator level four norm-preserving transitions are equally stable below $10^{-2}$, yet
three lose the summed quantity at $170$ percent, including the diagonal unitary that a spectral
parameterisation produces. Stability and conservation are therefore distinct, and the condition
separating them lies in the basis rather than the spectrum, which raises the question of whether
training can supply it.\par\endgroup


\subsection{Q4: Training Reaches the Condition Only by Freezing the Spectrum}
\label{sec:exp-q4}

\noindent\textbf{Setup.}\;
We test which of the two solutions admitted by Proposition~\ref{prop:measure} training reaches,
since one places the uniform vector in a stationary mode while the remaining modes rotate and
the other freezes every mode, so that the latent no longer evolves. We replace the assembled
basis with a free orthogonal matrix, initialised near the identity or Haar-randomly, and add a
penalty $\lambda\,\mathrm{drift}_{\text{momentum}}$ to the reconstruction loss. We train sixteen
seeds at $\lambda=10$ and sweep the weight over $\lambda\in\{0,0.1,1,10\}$, so that the outcome
does not rest on one choice of penalty, and we train a Hamiltonian and a graph network to
convergence at six training-set sizes, which fixes the reference for a conservation that is
learned rather than assembled. We read every run by the angle of its basis to the uniform
vector, by the mean $|\lambda_k|$ over the modes that do not carry it and by the momentum drift
at step $25$.

\vspace{-10pt}

\begin{figure}[!h]
\centering
\includegraphics[width=\linewidth]{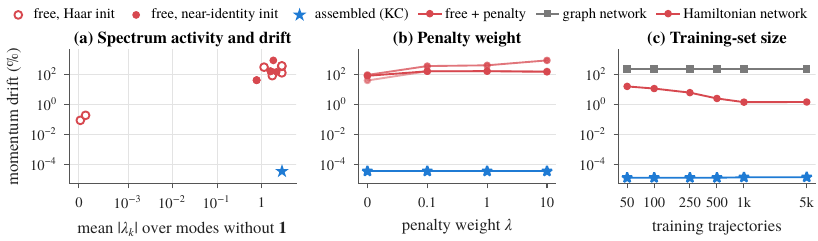}
\vspace{-18pt}
\caption{\textbf{Conservation and spectral activity of a learned modal basis, visual $N$-body
simulator.} \textbf{(a)} Momentum drift at step $25$ against the mean $|\lambda_k|$ over the
modes that do not carry the uniform vector, for sixteen free-basis seeds and for the assembled
operator. \textbf{(b)} Drift as the penalty weight varies, where every nonzero weight lands
above the unpenalised run. \textbf{(c)} Drift as the training-set size varies, where the learned
references flatten and the assembled operator does not move. Values in
Tables~\ref{tab:app-q4-mechanism}--\ref{tab:app-q4-converge}.}
\label{fig:alignment}
\vspace{-10pt}
\end{figure}

\begingroup\looseness=-1
\noindent\textbf{Results.}\;
The condition is reached by training only through the second solution, in which the dynamics are
given up. Fourteen of sixteen seeds keep a rotating spectrum, with mean $|\lambda_k|$ between
$0.77$ and $2.85$ on the modes that do not carry the uniform vector, and drift by $44$ to $932$
percent (Figure~\ref{fig:alignment}a). The remaining two reach a mean $|\lambda_k|$ below
$10^{-3}$ and drift by $0.1$ to $0.2$ percent, having driven every mode to $d_k=1$ so that the
latent no longer evolves (Table~\ref{tab:app-q4-mechanism}). The assembled operator alone
conserves with an active spectrum, at mean $|\lambda_k|=2.87$ and drift $3.5\times10^{-5}$
percent, because $c=e_{\kzero}$ holds by construction. The basis is not moved by the penalty
either, rotating by under ten degrees from its initialisation on every seed under both
initialisations (Tables~\ref{tab:app-q4-align} and \ref{tab:app-q4-haar}). Any nonzero weight
raises drift above the unpenalised run on every seed (Figure~\ref{fig:alignment}b), and a
Hamiltonian network improves from $16.9$ percent at $n=50$ to $1.49$ at $n=1000$ and no further
(Figure~\ref{fig:alignment}c). Assembly is thus the only solution that holds conservation and
the dynamics together, which leaves the question of how far it reaches.\par\endgroup

\subsection{Q5: The Layering Transfers as Far as the Algebra Allows}
\label{sec:exp-q5}

\noindent\textbf{Setup.}\;
The two boundaries of \S\ref{sec:scope} are tested separately. The quantity boundary of
Corollary~\ref{cor:applic} is tested on the charged, gravitational and spring $N$-body systems
of the EGNN benchmark \citep{satorras2021en, brandstetter2022geometric} at $T=1000$, where
linear momentum is linear in the state, angular momentum is bilinear and energy is quadratic
only on the spring system, together with five Hermitian forms outside the family of
Corollary~\ref{cor:family}. The dynamics boundary of Proposition~\ref{prop:bound} is tested on a
stiff reaction network relaxing to equilibrium, and transfer at the operator level on four
systems from other fields, with the row sums intact and broken by a matched perturbation.

\begin{figure}[!t]
\centering
\includegraphics[width=\linewidth]{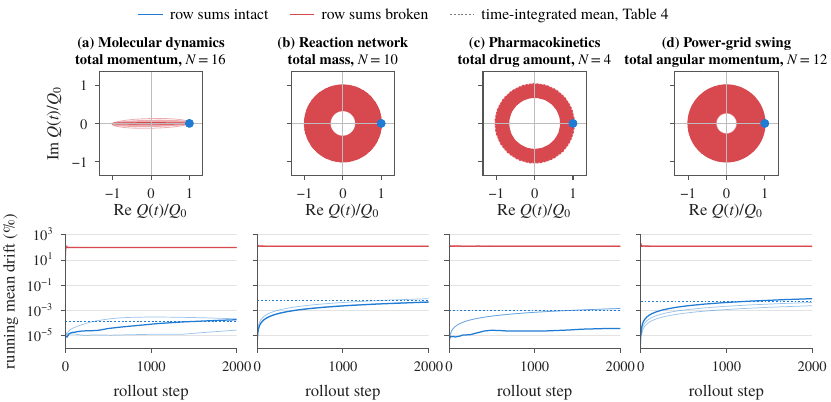}
\caption{\textbf{Four dynamical systems from other fields, operator level, $2000$ steps.} Top:
the conserved quantity $Q(t)$ relative to its initial value as a point in the complex plane, for
molecular dynamics the momentum vector in the plane of $Q_0$, seeds $42$/$43$/$44$. With the row
sums intact it stays at $1$, and with them broken its magnitude is nearly kept while its
direction rotates, which is the summed quantity lost while the norm is held. Bottom: running
time average of $\|Q(t)-Q_0\|/\|Q_0\|$, whose endpoints are the values of
Table~\ref{tab:q5-cross}.}
\label{fig:crossdomain}
\vspace{-12pt}
\end{figure}

\begingroup\looseness=-1
\noindent\textbf{Results.}\;
Exactly the quantities admitted by Corollary~\ref{cor:applic} are held, and the boundary is
visible in the same run. Linear momentum is held at $10^{-6}$ while every other baseline drifts
by $1$ to $8$, angular momentum improves by $3$ to $12$ times without reaching the floor, since a
bilinear quantity is not placed on a held mode by the linear tie, and the lowest energy drift is
reached on the spring system, whose potential is quadratic (Table~\ref{tab:q5-nbody}). The same
boundary appears inside the latent, where the zero mode of the family drifts by $10^{-5}$ while
five Hermitian forms outside it drift by $10$ to $117$ under the same operator
(Table~\ref{app:exp-outside}), and the decoded total momentum accordingly stays at a point over
$1000$ steps under KC while every diagonal transition carries it away
(Figure~\ref{fig:app-nbody-momentum}). On the dynamics boundary the latent zero mode is still held to
$5.6\times10^{-5}$ percent on the stiff system, while baselines that can contract fit the state
more accurately, as Proposition~\ref{prop:bound} predicts (Table~\ref{tab:app-q5-robertson}). The
layering further carries to molecular dynamics, chemical kinetics, pharmacokinetics and
power-grid swing dynamics, where the conserved quantity drifts by $10^{-4}$ to $10^{-2}$ percent
with the row sums intact and by $100$ to $128$ with them broken (Table~\ref{tab:q5-cross} and
Figure~\ref{fig:crossdomain}). What is held, and where the construction ceases to apply, is
therefore fixed before any parameter is set.\par\endgroup

%% file: S5_conclusions.tex
\section{Conclusion}
\label{sec:conclusion}

Conservation in a diagonal recurrence is layered: the norm, the summed quantity and an angular
one each rest on a separate condition on the operator, and a unit-modulus spectrum conserves only
the first. We therefore propose the Kronecker--Cayley transition (KC), which assembles the
generator as a graph Laplacian and advances it by a Cayley transform, so that every quantity the
algebra admits is held at every parameter value without correction. We prove the layering
itself, the exact $2m^2$-dimensional family the assembled operator holds, and the condition
under which a transition holds the summed quantity, which a trained basis reaches only by
freezing its spectrum. On four grid benchmarks, KC holds mass at the round-off level of the
rollout, whereas every diagonal baseline drifts and two mass correctors fail once the prediction
they correct diverges, and the same layering carries to molecular dynamics, chemical kinetics,
pharmacokinetics and power-grid swing dynamics.

%% file: S6_supplementary.tex
\clearpage
\appendix

\begin{center}
{\LARGE\bfseries Appendix}
\end{center}
\vspace{0.8em}

\noindent{\large\bfseries Table of Contents}
\vspace{0.4em}

\hrule height 0.8pt
\vspace{0.8em}

\startcontents[appendix]
\printcontents[appendix]{}{0}{}

\vspace{0.8em}
\hrule height 0.8pt

\clearpage

\input{Appendix_glossary}
\input{Appendix_related_work}
\input{Appendix_method}
\input{Appendix_algorithm}
\input{Appendix_exp_10.16}
\input{Appendix_limitations_future_work}

%% file: Appendix_glossary.tex
\newpage
\section{Notation and Implementation Details}
\label{app:glossary}

{\scriptsize
\renewcommand{\arraystretch}{1.05}
\setlength{\tabcolsep}{4pt}
\setlength{\LTcapwidth}{\linewidth}
\begin{longtable}{|
  >{\raggedright\arraybackslash}p{\dimexpr0.30\linewidth-2\tabcolsep-0.6pt\relax}|
  >{\raggedright\arraybackslash}p{\dimexpr0.57\linewidth-2\tabcolsep-0.6pt\relax}|
  >{\raggedright\arraybackslash}p{\dimexpr0.13\linewidth-2\tabcolsep-0.6pt\relax}|}
\caption{\textbf{Notation.}}
\label{tab:notation}\label{app:notation}\\
\hline
\ghead\textcolor{white}{\textbf{Symbol}} & \textcolor{white}{\textbf{Meaning}} & \textcolor{white}{\textbf{Introduced}} \\
\endfirsthead
\hline
\ghead\textcolor{white}{\textbf{Symbol}} & \textcolor{white}{\textbf{Meaning}} & \textcolor{white}{\textbf{Introduced}} \\
\endhead
\hline
\endfoot
\hline
\endlastfoot
\gband{3}{Latent state and partition}
$N,\;m,\;d$ & object subspaces, width of each subspace, and latent dimension, $d{=}Nm$ & \S\ref{sec:formulation} \\
$z{=}(z_1,\dots,z_N)\in\mathbb{C}^{d}$ & latent state, with $z_i\in\mathbb{C}^{m}$ the subspace of object $i$ & \S\ref{sec:formulation} \\
$x_t,\;a_t,\;t,\;h$ & observation, action, time index, and rollout step & \S\ref{sec:formulation} \\
$E,\;D$ & encoder $x\mapsto z$ and decoder $z\mapsto\hat{x}$, shared by every latent transition compared here & \S\ref{sec:formulation} \\
\gband{3}{The operator}
$w_{ij}{=}\mathrm{MLP}_\theta(\|z_i{-}z_j\|^2,a)$ & learned pairwise coupling, symmetrised as $(w_{ij}{+}w_{ji})/2$. The only trained component of the transition & Eq.~\ref{eq:coupling} \\
$\Lop\in\mathbb{R}^{N\times N}$ & graph Laplacian, $\Lop_{ij}{=}-w_{ij}$ off the diagonal and $\Lop_{ii}{=}\sum_{j\neq i}w_{ij}$, hence symmetric with zero row sums & Eq.~\ref{eq:generator} \\
$\Ham{=}\Lop\otimes\Iden{m}$ & isotropic pairwise generator: Hermitian, zero row sums, blocks proportional to the identity & Eq.~\ref{eq:generator} \\
$\Cay(\Ham){=}(\Iden{}{+}\tfrac{\mathrm{i}}{2}\Ham)^{-1}(\Iden{}{-}\tfrac{\mathrm{i}}{2}\Ham)$ & Cayley transform, unitary for every Hermitian $\Ham$ & Eq.~\ref{eq:cayley} \\
$\Lop{=}P\Lambda P^{\!\top}$, $\zmodal{=}(P^{\!\top}{\otimes}\Iden{m})z$ & eigendecomposition of the Laplacian and the resulting modal coordinates & Thm.~\ref{thm:kcd} \\
$\dk{=}(1{-}\mathrm{i}\lk/2)/(1{+}\mathrm{i}\lk/2)$ & modal eigenvalue, on the unit circle for every real $\lk$ & Eq.~\ref{eq:thmkcd} \\
$\lk,\;\kzero$ & Laplacian eigenvalue and the index of the zero mode, $\lambda_{\kzero}{=}0$ so $d_{\kzero}{=}1$ & Thm.~\ref{thm:kcd} \\
$\uniform$ & uniform vector, the eigenvector of $\Lop$ at eigenvalue zero whenever the row sums vanish & Prop.~\ref{prop:measure} \\
\gband{3}{Conserved quantities and the condition each one needs}
$\|z\|^2$ & latent norm, preserved by $|\dk|{=}1$ alone and therefore insensitive to the row sums & Eq.~\ref{eq:qthree} \\
$p(z){=}\textstyle\sum_i z_i$ & momentum proxy, preserved by $d_{\kzero}{=}1$, which holds only when the row sums vanish & Eq.~\ref{eq:qthree} \\
$\operatorname{Im}(z^\dagger\mathbf{J}z)$ & angular proxy, preserved by the scalar-times-identity block structure of $\Ham$ & Eq.~\ref{eq:qthree} \\
$\mathrm{drift}_Q@h$ & $100\,\|Q(\hat z_h)-Q(z_0)\|/\|Q(z_0)\|$, in percent, at rollout step $h$, measured from the first decoded frame so that the encoder--decoder round trip is excluded. That round trip loses $2$ percent of the dam-break mass once, at $t{=}1$ & \S\ref{sec:experiments} \\
tied readout & fixed linear map carrying total mass, or linear momentum on $N$-body systems, to the latent zero mode $(\onevec^{\!\top}{\otimes}\Iden{m})z$, so that Corollary~\ref{cor:applic} applies. On the sphere each patch is scaled by the inverse of its mean $\cos(\text{lat})$, so the zero mode is the area-weighted integral at patch resolution & Tab.~\ref{tab:hparams} \\
\gband{3}{Basis, spectrum, and the conservation condition}
$O(N)\times(S^1)^N$ & parameter space of a unitary diagonal transition, $U{=}(Q\operatorname{diag}(d)Q^{\!\top})\otimes\Iden{m}$, carrying a modal basis and a spectrum on the torus & Prop.~\ref{prop:measure} \\
$Q\in O(N)$ & modal basis of the transition. Either the Laplacian eigenbasis, or a free orthogonal matrix in the learned-basis control & Prop.~\ref{prop:measure} \\
$c{=}Q^{\!\top}\uniform$ & coefficients of the uniform vector in the modal basis. The summed quantity is preserved exactly when $(\dk{-}1)c_k{=}0$ for every $k$ & Eq.~\ref{eq:conservingset} \\
$\varphi$ & angle between $\uniform$ and the nearest column of $Q$, computed as $\arccos\max_j|c_j|$. Zero when the uniform vector is a column, at most $\arccos(1/\sqrt N)$ when it is split equally & Tab.~\ref{tab:app-q4-align} \\
$|Q^{\!\top}\uniform|$ & projection of the uniform vector on each mode. One entry equal to $1$ means a single mode carries the summed quantity, and $1/\sqrt{3}\approx0.577$ in every entry means it is divided equally among three & Tab.~\ref{tab:app-q4-mechanism} \\
\gband{3}{Operator perturbation}
$\varepsilon$, $\|\Lop'{-}\Lop\|_F/\|\Lop\|_F$ & perturbation scale and the relative Frobenius change it produces & Tab.~\ref{tab:perturb} \\
compensated & diagonal corrected so that the row sums stay zero, which leaves the condition for momentum intact & Tab.~\ref{tab:perturb} \\
uncompensated & same Frobenius magnitude with no diagonal correction, so the row sums no longer vanish & Tab.~\ref{tab:perturb} \\
\gband{3}{Cross-domain drift}
time-integrated mean & $\tfrac{1}{T}\sum_{t=1}^{T}\mathrm{drift}_Q@t$ over $T{=}2000$ steps. Does not saturate & Tab.~\ref{tab:q5-cross} \\
\gband{3}{Evaluation}
$\mathrm{PSNR}@h$ & decoded-frame peak signal-to-noise ratio at step $h$ & Tab.~\ref{tab:app-q4-converge} \\
mean-frame baseline & PSNR of the training-set mean frame used as a constant prediction, reported per evaluation set in \S\ref{sec:experiments}. Sets the level above which PSNR carries information about the dynamics & \S\ref{sec:experiments} \\
$s$ & seed, $s\in\{42,43,44\}$ unless the caption states otherwise. The alignment study extends to $s\in\{42,\dots,49\}$ & App.~\ref{app:experiments} \\
\end{longtable}
}

\newpage
\newcommand{\MCW}[1]{\multicolumn{2}{>{\raggedright\arraybackslash}p{\dimexpr0.83\linewidth-2\tabcolsep-0.6pt\relax}|}{#1}}
{\scriptsize
\renewcommand{\arraystretch}{1.08}
\setlength{\tabcolsep}{4pt}
\setlength{\LTcapwidth}{\linewidth}
\begin{longtable}{|
  >{\raggedright\arraybackslash}p{\dimexpr0.17\linewidth-2\tabcolsep-0.6pt\relax}|
  >{\raggedright\arraybackslash}p{\dimexpr0.63\linewidth-2\tabcolsep-0.6pt\relax}|
  >{\raggedright\arraybackslash}p{\dimexpr0.20\linewidth-2\tabcolsep-0.6pt\relax}|}
\caption{\textbf{Implementation details.}}
\label{tab:hparams}\label{app:impl}\label{tab:datasets}\label{app:benchmarks}\label{tab:controls}\\
\hline
\ghead\textcolor{white}{\textbf{Setting}} & \textcolor{white}{\textbf{Value}} & \textcolor{white}{\textbf{Scope}} \\
\endfirsthead
\caption[]{\textbf{Implementation details} (continued).}\\
\hline
\ghead\textcolor{white}{\textbf{Setting}} & \textcolor{white}{\textbf{Value}} & \textcolor{white}{\textbf{Scope}} \\
\endhead
\hline
\endfoot
\hline
\endlastfoot
\gband{3}{Public, grid}
Dam break
  & \MCW{Shallow-water dam break of PDEBench \citep{takamoto2022pdebench}, $1000$ trajectories at
    $128{\times}128$, split $900/100$. The conserved field is the total mass} \\
Advection
  & \MCW{Advection equation of PDEBench \citep{takamoto2022pdebench}, $1000$ trajectories at $1024$
    points, split $900/100$} \\
Sphere
  & \MCW{Spherical shallow water of PDEArena \citep{gupta2022pdearena}, $96{\times}192$
    latitude--longitude grid, $500$ trajectories split $400/100$ over disjoint
    initial-condition seeds, pressure anomaly as the conserved field. Re-posed as in
    \S\ref{app:exp-sphere}} \\
Allen--Cahn
  & \MCW{Generated by us in the Rubinstein--Sternberg form on a $128{\times}128$ periodic grid,
    $1000$ trajectories of $100$ steps, with mass conserved to $10^{-16}$ and energy monotone
    decreasing} \\
\gband{3}{Public, particles}
$N$-body, EGNN
  & \MCW{Charged, gravitational and spring systems of the EGNN benchmark \citep{satorras2021en,
    brandstetter2022geometric} at the standard $3000/2000/2000$ split} \\
Spring benchmark
  & \MCW{\citet{thangamuthu2022unravelling} under its published protocol: equal masses on a closed
    ring, two connections per body, unit stiffness and rest length,
    $q_0\sim\mathcal{U}(0,2)$ and $p_0\sim\mathcal{U}(0,0.1)$ after mean centring,
    St\"ormer--Verlet at $\Delta t{=}10^{-3}$, subsampled to 100 frames. Scored on its own
    three metrics} \\
\gband{3}{Visual, in distribution}
$N$-body, 30 frames
  & \MCW{Planar gravitational $N$-body, $N{=}3$, unit constant, softening $0.1$, leapfrog at
    $\Delta t{=}0.01$. Positions $\mathcal{N}(0,0.5)$ and velocities $\mathcal{N}(0,0.3)$ with
    the centre of mass removed, so the physical momentum is zero. Rendered at $64{\times}64$,
    one colour channel per object, actions $\mathcal{N}(0,0.01)$, $2000/256$ split} \\
$N$-body, 101 frames
  & \MCW{Same physics and renderer over a longer horizon, 256 trajectories. Used for every
    rollout beyond 29 steps} \\
\gband{3}{Cross domain, operator level}
Molecular dynamics
  & \MCW{$N{=}16$ particles of width $m{=}8$, state encoded as position plus $i$ times momentum per
    axis. Conserved quantity is the total linear momentum $\sum_i m_i v_i$, read from the
    imaginary part of the first two channels. A Lennard-Jones simulation with velocity Verlet
    at $\Delta t{=}0.005$ in float64 supplies the symplectic floor and the fixed operator of
    the spectrum-invariant control} \\
Chemical reaction network
  & \MCW{$N{=}10$ species of width $m{=}4$, conserving the total mass $\sum_i c_i$, read from the
    real part of the first channel} \\
Pharmacokinetics
  & \MCW{$N{=}4$ compartments of width $m{=}4$, conserving the total amount across compartments} \\
Power-grid swing
  & \MCW{$N{=}12$ nodes of width $m{=}4$, conserving the mass-weighted swing momentum
    $\sum_i M_i \dot{x}_i$ with $M_i\sim\mathcal{U}(1,5)$, the uniformly weighted zero mode at
    the operator level} \\
\gband{3}{Architecture and training, one setting per dataset family}
Grid benchmarks
  & $N{=}64$ patches of width $m{=}8$, per-patch MLP encoder and tied per-patch linear
    decoder. Adam at $10^{-3}$, batch $32$, $100$ epochs and $150$ on the sphere. Rollouts of
    $100$ steps on the dam break and Allen--Cahn, $200$ on advection and $86$ on the sphere
  & Dam break, advection, Allen--Cahn, sphere \\
$N$-body
  & $N{=}5$ or $10$ particles of width $m{=}8$, per-particle MLP encoder and tied per-particle
    linear decoder. Adam at $10^{-3}$, batch $32$, $200$ epochs, rollouts of $1000$ steps
  & Charged, gravity, spring \\
Visual $N$-body
  & $z\in\mathbb{C}^{120}$, $N{=}3$ subspaces of width $m{=}40$. Four convolutional blocks
    with channels $[64,128,256,512]$, a U-Net decoder with skip decay $K{=}10$, and $15$M
    parameters in total. Rollouts of $99$ steps, under the protocols below
  & All latent transitions \\
Spring benchmark
  & $N{=}3$ or $5$ objects of width $m{=}40$ under the same convolutional encoder and decoder.
    Adam at $10^{-3}$, batch $16$, $40$ epochs, rollouts of $25$ steps
  & Spring benchmark \\
Coupling network, Cayley step
  & MLP $17\to64\to32\to1$ on $(\|z_i{-}z_j\|^2,a)$, symmetric in $(i,j)$, $3{,}963$
    parameters with the action embedding and the only trained part of the transition. One
    float64 eigendecomposition of the $N{\times}N$ Laplacian, then diagonal phase rotations
  & Constructed basis \\
\gband{3}{Training protocols, Adam throughout, no early stopping}
P1
  & $2{\times}10^{-4}$, batch 16, 30 epochs, seeds $\{42,43,44\}$
  & Four-level ablation, main comparison \\
P1$'$
  & $2{\times}10^{-4}$, batch 16, 40 epochs, seeds $\{42,\dots,49\}$
  & Alignment test and penalty sweep \\
P2
  & $3{\times}10^{-4}$, batch 32, 80 epochs, seeds $\{42,43,44\}$
  & External recurrent baselines, basis ablation, long-horizon rollouts \\
P3
  & $5{\times}10^{-3}$, batch 128, 200 epochs
  & Synthetic invariant task \\
P5
  & $2{\times}10^{-4}$, batch 16, 30 epochs, stopping after 20 epochs without improvement
    and capped at 150, seeds $\{42,43,44\}$
  & Sample-size sweep trained to convergence \\
\gband{3}{Controls}
Four levels
  & A leaves $|\dk|$ free, B adds $|\dk|{=}1$, C adds one mode per object fixed at unity in the
    coordinate basis, and D is the full construction. Each level adds one ingredient
  & Separates the spectrum from the basis \\
Learned orthogonal basis
  & Same architecture and objective with $Q$ a free orthogonal matrix. Penalty weight
    $\lambda\in\{0,0.1,1,10\}$, seeds $\{42,\dots,49\}$. Initialised either near the identity,
    as $Q{=}\Cay((W{-}W^{\!\top})/2)$ with $W$ i.i.d.\ $\mathcal{N}(0,0.01^2)$ so that
    $\|Q{-}\Iden{}\|_F\approx0.03$, or Haar-randomly from the QR factorisation of a standard
    Gaussian
  & Isolates the source of the basis \\
Norm-preserving operators
  & A Haar-random unitary, a dense unitary from the Cayley transform of a skew-Hermitian
    matrix mixing all $d$ modes, a diagonal unitary with an independent phase per mode, and
    the construction. All four have unit-modulus eigenvalues and are therefore stable
  & Separates stability from conservation \\
Spectrum-invariant read-out
  & One fixed Laplacian with its spectrum held constant, read along the eigenvector of
    smallest $|\lambda|$ and along that of largest $|\lambda|$, on the synthetic graph and on
    the molecular-dynamics operator
  & Isolates the read-out direction \\
External operators
  & FNO using the published implementation and default settings. CPFNO reimplemented after
    \citet{liu2025cpfno}, corrector $\hat u{+}(m_0{-}m(\hat u))A$ for a learned softmax weight
    $A$ inside the forward pass and the loss on the corrected output, logged before and after
    the corrector at every step
  & Correction against condition \\
Compensated perturbation
  & Diagonal corrected so that the row sums stay zero, against an uncompensated
    perturbation of the same Frobenius magnitude
  & Separates the two conditions \\
\gband{3}{Evaluation}
Rollout, precision
  & 25 steps on the 30-frame sets, 99 and 2000 on the long-horizon sets, float32 throughout,
    which sets the floor on the reported drift. The symplectic reference is computed in
    float64, since float32 overflows on the Lennard-Jones force
  & All experiments \\
Determinism
  & Deterministic given the seed on the hardware used here, verified by regenerating all four
    benchmarks and the three $N$-body systems at seed $42$ and recovering the stored values to
    three significant figures. FNO and CPFNO on advection are run under
    \texttt{torch.use\_deterministic\_algorithms}, since the tail of the rollout is sensitive
    to one-dimensional CUDA reductions
  & All experiments \\
Seeds
  & $\{42,43,44\}$ for every trained model, extended to $\{42,\dots,49\}$ for the alignment
    study
  & All experiments \\
\end{longtable}
}

%% file: Appendix_related_work.tex
\section{Extended Related Work}
\label{app:related}

\noindent\textbf{Parameterisations of the modal basis.}\;
The constructions that secure stability act on the spectrum and obtain the modal basis as a
by-product. Unitary recurrence restricts the transition to a manifold, through a product of
structured unitary factors \citep{arjovsky2016unitary}, a Stiefel-manifold gradient step on the
full matrix \citep{wisdom2016full}, a scaled Cayley transform of a skew-symmetric parameter
\citep{helfrich2018orthogonal}, or the exponential map of a Lie algebra \citep{lezcano2019cheap},
each resting on standard manifold optimisation \citep{absil2008optimization, hall2015lie,
boumal2023introduction}, and the eigenvectors are whatever the chosen parameterisation returns
at the optimum. Koopman autoencoders instead learn the coordinates, advancing an encoded latent
by a linear operator so that nonlinear dynamics become linear in the learned basis
\citep{lusch2018deep, azencot2020forecasting}, which makes that basis a function of the data and
of the optimiser. Our construction applies the transform of Equation~\ref{eq:cayley} used by
scoRNN and expRNN to a Laplacian rather than to a free skew-symmetric parameter, so that the
eigenbasis is determined by the interaction structure before any value is fitted, and the
vanishing row sums place the uniform vector at eigenvalue one, which is the mode that carries
the summed quantity at every value of the coupling (Proposition~\ref{prop:measure}).

\noindent\textbf{Conservation through a learned generator.}\;
The generator-based designs name the conserved scalar and learn it. A Hamiltonian network learns
the energy and advances the state along its symplectic gradient \citep{greydanus2019hamiltonian},
a Lagrangian network removes the requirement of canonical coordinates
\citep{cranmer2020lagrangian}, a symplectic recurrent network integrates the learned energy over
several steps at once \citep{chen2020symplectic}, and SymODEN extends the construction to systems
with control inputs and to dissipation \citep{zhong2020symoden, zhong2020dissipative}, while a
related design states the quantity as a constraint set and solves it at each step
\citep{lee2021deep}. Benchmarks across this family report that the benefit obtained depends on
how closely the stated structure matches the system at hand \citep{thangamuthu2022unravelling}.
A learned scalar is a general function of the state, so the induced flow is not linear in any
fixed basis and the linear-time diagonal step is unavailable. Restricting the conserved set to
quadratic forms is the price of retaining that step, and the commutant of
Theorem~\ref{thm:commutant} then returns the entire $2m^2$-dimensional family of
Corollary~\ref{cor:family} rather than one named quantity.

\noindent\textbf{Zero modes of a Laplacian in the applied sciences.}\;
For a symmetric Laplacian with vanishing row sums, $\Lop\onevec = 0$, so the flow
$\dot{x} = -\Lop x$ satisfies $\tfrac{d}{dt}\onevec^{\top}x = -\onevec^{\top}\Lop x = 0$. The
eigenvalue is exactly zero and the eigenvector is exactly the uniform vector for every connected
weighted graph, so neither one depends on the coupling strengths and neither is perturbed when
those strengths are estimated from data. Each applied field reads the resulting sum as its own
invariant, the conserved mass of a reaction network \citep{veerman2023chemical}, the agreement
value of average consensus \citep{olfatisaber2007consensus}, the retained dose of a compartmental
pharmacokinetic model, and the total angular momentum of the swing equations of a power network
\citep{dorfler2013synchronization}. In each case the matrix is supplied by the physical model and
the zero mode is read from it. The construction of Equation~\ref{eq:generator} proceeds in the
opposite direction, assembling the matrix from learned couplings under a pattern that imposes
the zero mode, which is the mechanism by which those four systems transfer at the operator level
in \S\ref{sec:exp-q5} and lose their invariant once a matched perturbation breaks the row sums.

%% file: Appendix_method.tex
\section{Proofs and numerical detail}
\label{app:proofs}

We prove the results of \S\ref{sec:method} in the order they appear, then record the numerical
properties that set the floor against which the measured drift is read
(\S\ref{app:numerics}) and the batched form used in the experiments
(\S\ref{app:batched}). Throughout, $\Lop\in\mathbb{R}^{N\times N}$ is the
graph Laplacian of Equation~\ref{eq:generator}, $\Ham=\Lop\otimes\Iden{m}$, $U=\Cay(\Ham)$,
$\Lop=P\Lambda P^{\!\top}$ with $\Lambda=\operatorname{diag}(\lambda_1,\dots,\lambda_N)$, and
$\kzero$ is the index of the zero eigenvalue.

\subsection{Preliminaries}
\label{app:prelim}

\begin{lemma}[Structure of the assembled generator]
\label{lem:structure}
For every value of $\theta$ the matrix $\Ham$ of Equation~\ref{eq:generator} satisfies
\begin{enumerate}[label=(\roman*),leftmargin=2.2em,itemsep=1pt,topsep=2pt]
\item $\Ham^\dagger=\Ham$,
\item $\Ham(\onevec\otimes e)=0$ for every $e\in\mathbb{C}^m$,
\item $[\Ham,\;\Iden{N}\otimes C]=0$ for every $C\in\mathbb{C}^{m\times m}$.
\end{enumerate}
\end{lemma}

\begin{proof}
Property (i). The coupling network returns real values and the assembly symmetrises them,
$w_{ij}=w_{ji}$, so $\Lop$ is real symmetric and $\Lop=\Lop^{\!\top}=\Lop^\dagger$. The Kronecker
product of Hermitian factors is Hermitian, hence $\Ham^\dagger=\Ham$.

Property (ii). Row $i$ of $\Lop$ sums to
$\Lop_{ii}+\sum_{j\neq i}\Lop_{ij}=\sum_{k\neq i}w_{ik}-\sum_{j\neq i}w_{ij}=0$, so
$\Lop\onevec=0$. Then
$\Ham(\onevec\otimes e)=(\Lop\otimes\Iden{m})(\onevec\otimes e)=(\Lop\onevec)\otimes e=0$.

Property (iii). Using $(A\otimes B)(C\otimes D)=AC\otimes BD$,
\begin{equation*}
(\Lop\otimes\Iden{m})(\Iden{N}\otimes C)=\Lop\otimes C=(\Iden{N}\otimes C)(\Lop\otimes\Iden{m}),
\end{equation*}
which is the stated commutator identity. The three properties follow from the pattern of
Equation~\ref{eq:generator} and hold whatever values the coupling network returns.
\end{proof}

\begin{lemma}[Cayley transform of a Kronecker product]
\label{lem:cayley}
Let $A\in\mathbb{C}^{N\times N}$ be Hermitian. Then
$\Cay(A\otimes\Iden{m})=\Cay(A)\otimes\Iden{m}$, and $\Cay(A)$ is unitary.
\end{lemma}

\begin{proof}
Write $S=\tfrac{\mathrm{i}}{2}A$, so $S^\dagger=-S$. The matrix $\Iden{}+S$ is invertible, since
$(\Iden{}+S)v=0$ would imply $v^\dagger v=-v^\dagger Sv$, whose right-hand side is purely imaginary
while the left is real and non-negative, forcing $v=0$. For the Kronecker structure,
$S\otimes\Iden{m}=\tfrac{\mathrm{i}}{2}(A\otimes\Iden{m})$ and
\begin{equation*}
\bigl(\Iden{Nm}+S\otimes\Iden{m}\bigr)^{-1}
=\bigl((\Iden{N}+S)\otimes\Iden{m}\bigr)^{-1}
=(\Iden{N}+S)^{-1}\otimes\Iden{m},
\end{equation*}
using $(X\otimes\Iden{m})^{-1}=X^{-1}\otimes\Iden{m}$. Multiplying by
$(\Iden{N}-S)\otimes\Iden{m}$ and collecting the factors yields
$\Cay(A\otimes\Iden{m})=\Cay(A)\otimes\Iden{m}$. Unitarity follows from
\begin{equation*}
\Cay(A)^\dagger\Cay(A)
=(\Iden{}-S^\dagger)^{-1}(\Iden{}+S^\dagger)(\Iden{}-S)(\Iden{}+S)^{-1}
=(\Iden{}+S)^{-1}(\Iden{}-S)(\Iden{}-S)^{-1}(\Iden{}+S)=\Iden{},
\end{equation*}
where the middle equality uses $S^\dagger=-S$ and the commutation of $(\Iden{}\pm S)$ with each
other.
\end{proof}

\begin{lemma}[Matrices commuting with every Laplacian]
\label{lem:commall}
Let $N\ge2$ and let $\mathcal{L}_N$ be the set of real symmetric $N\times N$ matrices with zero
row sums. A matrix $A\in\mathbb{C}^{N\times N}$ satisfies $[\Lop,A]=0$ for every
$\Lop\in\mathcal{L}_N$ if and only if $A\in\operatorname{span}\{\Iden{N},\onevec\onevec^{\!\top}\}$.
\end{lemma}

\begin{proof}
Sufficiency. $\Iden{N}$ commutes with everything, and
$\Lop\onevec\onevec^{\!\top}=(\Lop\onevec)\onevec^{\!\top}=0=\onevec(\Lop\onevec)^{\!\top}
=\onevec\onevec^{\!\top}\Lop$ for every $\Lop\in\mathcal{L}_N$, using symmetry of $\Lop$.

Necessity. $\mathcal{L}_N$ is a real vector space spanned by the elementary Laplacians
$E_{ij}=(e_i-e_j)(e_i-e_j)^{\!\top}$ for $i<j$, since every symmetric zero-row-sum matrix is a
combination of them. Commuting with every $\Lop$ is equivalent to commuting with every $E_{ij}$.
Write $u=e_i-e_j$, so $E_{ij}=uu^{\!\top}$. Then $[uu^{\!\top},A]=u(u^{\!\top}A)-(Au)u^{\!\top}$,
and this vanishes exactly when $Au=\mu u$ and $u^{\!\top}A=\mu u^{\!\top}$ for some scalar $\mu$,
that is when $u$ is a right and left eigenvector of $A$ with the same eigenvalue. Since this holds
for every difference $e_i-e_j$, the subspace $\onevec^\perp$ spanned by those differences lies in
a single eigenspace of $A$, and $A$ acts on $\onevec^\perp$ as a scalar $\mu$. Symmetry of $E_{ij}$
also forces $A$ to map $\onevec^\perp$ into itself and, by the same argument on the transpose,
$\onevec$ into $\operatorname{span}\{\onevec\}$. Hence $A=\mu\,\Pi_\perp+\nu\,\Pi_\parallel$ with
$\Pi_\parallel=\tfrac1N\onevec\onevec^{\!\top}$ and $\Pi_\perp=\Iden{N}-\Pi_\parallel$, which is
a combination of $\Iden{N}$ and $\onevec\onevec^{\!\top}$.
\end{proof}

\subsection{Characterisation of the conserved set (Theorem~\ref{thm:commutant})}
\label{app:thmcomm}

\begin{proof}
Let $U=\Cay(\Ham)$ with $\Ham$ Hermitian, so $U$ is unitary by Lemma~\ref{lem:cayley}. For a
Hermitian $G$, the identity $Q_G(Uz)=Q_G(z)$ for every $z$ is $z^\dagger U^\dagger GUz
=z^\dagger Gz$ for every $z$. Since $U^\dagger GU$ and $G$ are both Hermitian, equality of the
two quadratic forms on every $z$ forces $U^\dagger GU=G$, and multiplying on the left by $U$ yields
$GU=UG$, that is $[U,G]=0$.

It remains to show $[U,G]=0\Leftrightarrow[\Ham,G]=0$. Write $f(\lambda)=(1-\mathrm{i}\lambda/2)
/(1+\mathrm{i}\lambda/2)$, so $U=f(\Ham)$ in the functional calculus, and $f$ is injective on
$\mathbb{R}$, since $f(\lambda)=f(\mu)$ implies $(1-\mathrm{i}\lambda/2)(1+\mathrm{i}\mu/2)
=(1-\mathrm{i}\mu/2)(1+\mathrm{i}\lambda/2)$ and hence $\lambda=\mu$. Decompose
$\Ham=\sum_k\lambda_k\Pi_k$ into spectral projectors over distinct eigenvalues, so
$U=\sum_k f(\lambda_k)\Pi_k$ with the $f(\lambda_k)$ distinct. A matrix $G$ commutes with $\Ham$
exactly when it commutes with every $\Pi_k$, and likewise for $U$, because in each case the
projectors are polynomials in the matrix with distinct eigenvalues. The two commutants therefore
coincide.
\end{proof}

\subsection{Kronecker--Cayley decomposition (Theorem~\ref{thm:kcd})}
\label{app:thm}

\begin{proof}
By Lemma~\ref{lem:structure}(i) the matrix $\Lop$ is real symmetric, so Lemma~\ref{lem:cayley}
applies with $A=\Lop$ and yields $U=\Cay(\Lop)\otimes\Iden{m}$ together with unitarity.

Substituting $\Lop=P\Lambda P^{\!\top}$ and using that $\Cay$ acts on the spectrum,
\begin{equation}
\label{eq:app:modal}
\Cay(\Lop)=P\,\operatorname{diag}(d_1,\dots,d_N)\,P^{\!\top},
\qquad
\dk=\frac{1-\mathrm{i}\lk/2}{1+\mathrm{i}\lk/2}.
\end{equation}
Each $\lk$ is real because $\Lop$ is symmetric, so
$|\dk|^2=\dk\overline{\dk}=\frac{1+\lk^2/4}{1+\lk^2/4}=1$, so every modal eigenvalue lies on the
unit circle. Since $\Lop\onevec=0$ by Lemma~\ref{lem:structure}(ii), the spectrum contains
$\lambda_{\kzero}=0$, and Equation~\ref{eq:app:modal} yields $d_{\kzero}=1$. In the modal
coordinates $\zmodal=(P^{\!\top}\otimes\Iden{m})z$ the step is
$\zmodal^{(k)}_{t+1}=\dk\,\zmodal^{(k)}_{t}$, which is diagonal.

For the cost, one step evaluates $\binom{N}{2}$ couplings at $O(N^2)$, eigendecomposes $\Lop$ at
$O(N^3)$, applies $P^{\!\top}\otimes\Iden{m}$ and $P\otimes\Iden{m}$ at $O(N^2m)$ each, and scales
$N$ modes of width $m$ at $O(Nm)$, for $\cost$ in total. Forming $\Cay(\Ham)$ directly would
require inverting an $Nm\times Nm$ matrix at $\costfull$, which exceeds $\cost$ by a factor that
grows as $m^3/N$ once $m$ exceeds $N$.
\end{proof}

\subsection{Layered conservation (Corollary~\ref{cor:layered})}
\label{app:cor}

\begin{proof}
For the norm, Lemma~\ref{lem:cayley} makes $U$ unitary, so $\|Uz\|^2=\|z\|^2$ for every $z$, and
in modal coordinates the same statement is $\sum_k |\dk|^2\|\zmodal^{(k)}\|^2
=\sum_k\|\zmodal^{(k)}\|^2$, which requires $|\dk|=1$ and nothing further. For the sum, the
quantity is $p(z)=(\onevec^{\!\top}\otimes\Iden{m})z$, and by Lemma~\ref{lem:structure}(ii) the
subspace $\{\onevec\otimes e: e\in\mathbb{C}^m\}$ lies in the kernel of $\Ham$, which the Cayley
transform maps to the eigenspace of $U$ at eigenvalue one, since $\Ham v=0$ implies
$Uv=(\Iden{}+0)^{-1}(\Iden{}-0)v=v$, so $p(Uz)=p(z)$ for every $z$. For the angular pairing,
Lemma~\ref{lem:structure}(iii) puts every $\Iden{N}\otimes C$ in the commutant of $\Ham$, in
particular $G=\Iden{N}\otimes\mathbf{J}$, and Theorem~\ref{thm:commutant} then yields preservation
of $\operatorname{Im}(z^\dagger Gz)$. A generator of the form $\Lop\otimes\Iden{m}$ with $\Lop$ a
Laplacian satisfies all three properties, so the rightmost condition of Equation~\ref{eq:layers}
implies the other two. For the converses, two counterexamples suffice, each satisfying the
conditions to its left and failing the one stated.
\begin{enumerate}[label=(\alph*),leftmargin=2.2em,itemsep=2pt,topsep=3pt]
\item \emph{Unit modulus does not yield the sum.} Take $\Ham'=\operatorname{diag}(\mu_1,\dots,
\mu_N)\otimes\Iden{m}$ with all $\mu_k$ real and distinct and no $\mu_k$ equal to zero. Then
$\Cay(\Ham')$ is unitary and every modal eigenvalue lies on the unit circle, while
$\Lop'\onevec\neq0$, so $\onevec\otimes e$ is not in the kernel and $p$ is not preserved.
\item \emph{Vanishing row sums do not yield the angular pairing.} Take
$\Ham''=\Lop\otimes B$ with $\Lop$ a Laplacian and $B$ Hermitian with $[B,\mathbf{J}]\neq0$. Then
$\Ham''(\onevec\otimes e)=0$ for every $e$, so $p$ is preserved, while
$[\Ham'',\Iden{N}\otimes\mathbf{J}]=\Lop\otimes[B,\mathbf{J}]\neq0$ and the angular pairing is not.
\item \emph{The layers are strict.} Counterexample (a) satisfies the leftmost condition and fails
the middle one, and counterexample (b) satisfies the middle condition and fails the rightmost,
so no two of the three conditions are equivalent.
\end{enumerate}
The three conditions of Equation~\ref{eq:layers} are therefore distinct, and each is strictly
stronger than the one to its left.
\end{proof}

\subsection{The invariant family (Corollary~\ref{cor:family})}
\label{app:family}

\begin{proof}
By Theorem~\ref{thm:commutant}, a Hermitian $G$ is held by $U=\Cay(\Lop\otimes\Iden{m})$ exactly
when $[\Lop\otimes\Iden{m},G]=0$, and at every $\Lop\in\mathcal{L}_N$ exactly when this
commutator vanishes for every such $\Lop$. For sufficiency, take
$G=\Iden{N}\otimes C_1+\onevec\onevec^{\!\top}\otimes C_2$, so Lemma~\ref{lem:commall} yields
$[\Lop,\Iden{N}]=[\Lop,\onevec\onevec^{\!\top}]=0$ for every $\Lop\in\mathcal{L}_N$ and then
$[\Lop\otimes\Iden{m},G]=[\Lop,\Iden{N}]\otimes C_1+[\Lop,\onevec\onevec^{\!\top}]\otimes C_2=0$.
For necessity, write $G$ in block form as $G=\sum_{a,b}E_{ab}\otimes G_{ab}$ with $E_{ab}$ the $N\times N$ matrix
units and $G_{ab}\in\mathbb{C}^{m\times m}$. The commutator
$[\Lop\otimes\Iden{m},G]=\sum_{a,b}[\Lop,E_{ab}]\otimes G_{ab}$ vanishes for every
$\Lop\in\mathcal{L}_N$ only if, for every fixed $C$ in a basis of $\mathbb{C}^{m\times m}$, the
$N\times N$ coefficient matrix $A_C$ obtained by contracting $G$ against $C$ commutes with every
$\Lop$. By Lemma~\ref{lem:commall} each such $A_C$ lies in
$\operatorname{span}\{\Iden{N},\onevec\onevec^{\!\top}\}$, so
$G=\Iden{N}\otimes C_1+\onevec\onevec^{\!\top}\otimes C_2$ for some $C_1,C_2$, which is
Equation~\ref{eq:family}. Hermiticity of $G$ forces $C_1,C_2$ Hermitian. For the dimension, the two matrices $\Iden{N}$ and $\onevec\onevec^{\!\top}$ are linearly independent for $N\ge2$,
and the Hermitian $m\times m$ matrices form a real space of dimension $m^2$, so the family has
real dimension $2m^2$.
\end{proof}

\noindent
The summed quantity $p(z)=(\onevec^{\!\top}\otimes\Iden{m})z$ is linear rather than quadratic and
therefore not a member of Equation~\ref{eq:family}. It is nonetheless held, through the kernel of
$\Ham$ as in the proof of Corollary~\ref{cor:layered}, and its square $\|p(z)\|^2
=z^\dagger(\onevec\onevec^{\!\top}\otimes\Iden{m})z$ is a member with $(C_1,C_2)
=(0,\Iden{m})$.

\subsection{Applicability (Corollary~\ref{cor:applic})}
\label{app:applic}

\begin{proof}
With a linear decoder $y=Dz$, any function of $y$ is a function of $z$, and $q$ is held for every
$\theta$ exactly when $q\circ D$ is held on the latent for every $\theta$. By
Corollary~\ref{cor:family}, the functions of the latent held for every $\theta$ that are quadratic
forms are exactly those of Equation~\ref{eq:family}, which yields the equivalence for quadratic
$q$. If $q$ is not a
quadratic form in $y$, then $q\circ D$ is not a quadratic form in $z$, since composition with a
linear map preserves polynomial degree, so Corollary~\ref{cor:family} makes no statement about
it.
\end{proof}

\subsection{Lower bound for a unitary step (Proposition~\ref{prop:bound})}
\label{app:bound}

\begin{proof}
For any vectors $u,v$, the reverse triangle inequality is $\|u-v\|\ge\bigl|\|u\|-\|v\|\bigr|$.
Take $u=Uz_t$ and $v=z_{t+1}$, and use $\|Uz_t\|=\|z_t\|$ for unitary $U$. Equality holds when
$u$ and $v$ are non-negative multiples of one another.
\end{proof}


\subsection{Conservation of the summed quantity (Proposition~\ref{prop:measure})}
\label{app:prop}

\begin{proof}
Write the transition as $U=(Q\operatorname{diag}(d)Q^{\!\top})\otimes\Iden{m}$ with $Q\in O(N)$
and $d\in(S^1)^N$, and set $M=Q\operatorname{diag}(d)Q^{\!\top}$. The summed quantity is
$p(z)=(\onevec^{\!\top}\otimes\Iden{m})z$, and
\begin{equation*}
p(Uz)=(\onevec^{\!\top}\otimes\Iden{m})(M\otimes\Iden{m})z
=\bigl((M^{\!\top}\onevec)^{\!\top}\otimes\Iden{m}\bigr)z .
\end{equation*}
Requiring $p(Uz)=p(z)$ for every $z\in\mathbb{C}^{Nm}$ therefore requires
$M^{\!\top}\onevec=\onevec$. Substituting $M^{\!\top}=Q\operatorname{diag}(\bar d)Q^{\!\top}$ and
multiplying on the left by $Q^{\!\top}$, this is $\operatorname{diag}(\bar d)\,Q^{\!\top}\onevec
=Q^{\!\top}\onevec$, that is $(\bar d_k-1)c_k=0$ for every $k$ with $c=Q^{\!\top}\uniform$,
which is Equation~\ref{eq:conservingset} since $|d_k|=1$ makes $\bar d_k-1=0$ exactly when
$d_k-1=0$. The condition holds when every $k$ with $c_k\neq0$ has $d_k=1$. For the assembly,
$\Lop\onevec=0$ makes $\uniform$ the $\kzero$th column of $Q$, so $c=e_{\kzero}$, and
$\lambda_{\kzero}=0$ makes $d_{\kzero}=1$, so the condition holds with every other $d_k$ free.
For a transition with $d_k=1$ for every $k$, the condition holds for every $c$, and
$M=\Iden{N}$.
\end{proof}

\begin{remark}[The converse of Theorem~\ref{thm:kcd}]
\label{rem:converse}
The forward direction of Theorem~\ref{thm:kcd} is unconditional. The converse is not. Any unitary
$U$ without $-1$ in its spectrum can be written as $\Cay(\Ham)$ for the Hermitian
$\Ham=2\mathrm{i}(U-\Iden{})(U+\Iden{})^{-1}$, and in the eigenbasis of that $\Ham$ the step is
diagonal with unit-modulus entries. The summed quantity is preserved only when $\Ham$ additionally
has vanishing row sums in the coordinates of $p$, which is a condition on the pair
$(\Ham,p)$ and not on $U$ alone.
\end{remark}

\subsection{Numerical properties}
\label{app:numerics}

The guarantees above are exact in real arithmetic, and in floating-point the two invariants
behave differently.

\begin{proposition}[Floating-point floor]
\label{prop:precision}
Let $\epsilon$ denote the unit round-off of the arithmetic used for the rollout, and let
$\delta=|\lambda_{\kzero}|$ be the residual the eigensolver returns in place of the exact zero
eigenvalue. After $T$ steps of the modal recurrence,
\begin{align}
\label{eq:app:floornorm}
\text{norm:}\qquad
&\bigl|\,\|z_T\|^2-\|z_0\|^2\,\bigr| \;\le\; 2TN\epsilon\,\|z_0\|^2 \;+\; O(T^2\epsilon^2),\\[2pt]
\label{eq:app:floorsum}
\text{sum:}\qquad
&\bigl\|\,p(z_T)-p(z_0)\,\bigr\| \;\le\; T\bigl(\delta + 2\epsilon\bigr)\,\|p(z_0)\| \;+\;
 O\bigl(T^2(\delta+\epsilon)^2\bigr).
\end{align}
The two bounds differ in their governing term. Equation~\ref{eq:app:floornorm} is governed by the
round-off $\epsilon$ alone, and Equation~\ref{eq:app:floorsum} is governed by the eigensolver
residual $\delta$, which exceeds $\epsilon$ whenever the eigendecomposition is the less accurate of
the two operations.
\end{proposition}

\begin{proof}[Proof sketch]
For the norm, each step multiplies $N$ modes by numbers of modulus one, and a floating-point
product with a unit-modulus number introduces a relative error of at most $\epsilon$ per real
multiplication. Summing over modes and steps and discarding second-order terms yields
Equation~\ref{eq:app:floornorm}. The bound does not grow with the accuracy of the
eigendecomposition, since the computed $\Cay(\Lop)$ is unitary to round-off regardless $P$ and
$\Lambda$ were returned.

For the sum, the exact recurrence leaves $p$ fixed because $d_{\kzero}=1$. The computed phase is
$\hat{d}_{\kzero}=(1-\mathrm{i}\delta/2)/(1+\mathrm{i}\delta/2)$, whose argument is
$-2\arctan(\delta/2)$, so one step rotates the mode carrying $p$ by an angle of magnitude at most
$\delta$ and in addition incurs the round-off of the two modal transforms, bounded by $2\epsilon$.
The errors compose over $T$ steps, which yields Equation~\ref{eq:app:floorsum}.
\end{proof}

\noindent
The two invariants are not affected in the same way, and the asymmetry is visible in the measured
drift. Norm preservation is a property of the map at each step, since $\Cay(\Ham)$ is unitary
whatever numerical error entered $\Lop$, so the norm error does not accumulate and remains at the
round-off level for the whole rollout. Preservation of the sum requires $\lambda_{\kzero}$ to come
out of the eigendecomposition as exactly zero. A numerical eigensolver returns a residual $\delta$
that is small but non-zero, the phase $d_{\kzero}$ then differs from one by $O(\delta)$, and
Equation~\ref{eq:app:floorsum} accumulates the difference linearly in $T$. On the visual $N$-body
system the energy residual is flat across the rollout, at $1.4\mathrm{e}{-6}$ percent at step $5$
and $2.8\mathrm{e}{-6}$ at step $99$, while the momentum residual grows from $6.4\mathrm{e}{-6}$ to
$2.0\mathrm{e}{-5}$ over the same interval. Both lie at the float32 floor, and neither is
comparable with a residual measured at a different horizon. We eigendecompose $\Lop$ in float64 and evaluate the rollout in float32, so the reported residuals
are float32 quantities. The reference trajectories for the cross-domain systems are integrated in
float64, since a float32 integration of the Lennard-Jones force overflows and returns a reference
that is itself unusable.

\subsection{Batched form}
\label{app:batched}

The assembly is applied independently to every element of a batch, so the eigendecomposition is
batched over the leading dimension. With batch size $B$ the step costs $O(BN^3+BN^2m)$, and the
$N\times N$ eigendecomposition is the only part that does not vectorise over $m$. For the sizes used
here, $N$ from $3$ to $64$ and $m$ from $8$ to $40$, the modal transforms dominate the wall clock
and the eigendecomposition contributes an overhead that grows with $N$ and is independent of $m$.
The diagonal of $\Lop$ is formed by summing the off-diagonal entries of its own row rather than by
a separate parameter, so $\Lop\onevec=0$ holds to the accuracy of that sum and not to the accuracy
of a learned value. The modal transforms use the same $P$ in both directions, computed once per
step, so the round trip $(P\otimes\Iden{m})(P^{\!\top}\otimes\Iden{m})$ is the identity up to
round-off rather than up to the agreement of two separate factorisations.

%% file: Appendix_algorithm.tex
\newpage
\subsection{Full algorithm of the Kronecker--Cayley transition}
\label{app:algorithm}

\begin{algorithm}[H]
\caption{\textbf{The Kronecker--Cayley transition, training and rollout.}}
\label{alg:kc-full}
\small
\begin{algorithmic}[1]
\Require observations $x_{0:T}$ with actions $a_{0:T-1}$, objects $N$, subspace width $m$,
  encoder $E_\phi$, decoder $D_\psi$, coupling network $\mathrm{MLP}_\theta$, readout
  $r\in\mathbb{R}^{N}$, horizon $T$, batch $B$, learning rate $\eta$
\Ensure parameters $(\theta,\phi,\psi)$ and a rollout whose conserved family
  $\mathcal{Q}$ of Corollary~\ref{cor:family} is the same at every value of $\theta$
\Statex
\Statex \textit{\textcolor{lblblue}{Fixed before training.}}
\State $r\gets\onevec$, so that $r^{\!\top}\!\otimes\Iden{m}$ reads the latent zero mode
  \Comment{tied readout, Table~\ref{tab:hparams}}
\Statex
\Statex \textit{Training.}
\For{each minibatch $(x_{0:T},a_{0:T-1})$ of size $B$}
  \State $\hat{x}_{1:T}\gets\textsc{Rollout}(x_0,a_{0:T-1})$
  \State $\mathcal{L}\gets\tfrac{1}{T}\sum_{t=1}^{T}\lVert \hat{x}_t-x_t\rVert^2$
    \Comment{reconstruction over the rollout}
  \State $(\theta,\phi,\psi)\gets\mathrm{Adam}\bigl((\theta,\phi,\psi),\,
    \nabla\mathcal{L},\,\eta\bigr)$
\EndFor
\Statex
\Statex \textit{Rollout.}
\Function{Rollout}{$x_0$, $a_{0:T-1}$}
  \State $z_0\gets E_\phi(x_0)\in\mathbb{C}^{B\times Nm}$, split as
    $(z_{0,1},\dots,z_{0,N})$ with $z_{0,i}\in\mathbb{C}^{B\times m}$
  \For{$t=0,\dots,T-1$}
    \State \textcolor{lblgreen}{$w_{ij}\gets\mathrm{MLP}_\theta\bigl(\lVert z_{t,i}-z_{t,j}
      \rVert^2,\;a_t\bigr)$ for $i<j$, \; $w_{ji}\gets w_{ij}$}\label{ln:couple}
      \Comment{Eq.~\ref{eq:coupling}, the only learned line}
    \State \textcolor{lblblue}{$\Lop_{ij}\gets-w_{ij}$ for $i\neq j$, \;
      $\Lop_{ii}\gets\sum_{k\neq i}w_{ik}$}\label{ln:lap}
      \Comment{Eq.~\ref{eq:generator}, $\Lop\onevec=0$ by the row sum}
    \State \textcolor{lblblue}{$(\Lambda,P)\gets\textsc{SymEig}(\Lop)$ in float64}\label{ln:eig}
      \Comment{$\Lop=P\Lambda P^{\!\top}$, rollout in float32}
    \State \textcolor{lblblue}{$\dk\gets(1-\mathrm{i}\lk/2)/(1+\mathrm{i}\lk/2)$,
      $k=1,\dots,N$}
      \Comment{Eq.~\ref{eq:thmkcd}, $|\dk|=1$ and $d_{\kzero}=1$}
    \State \textcolor{lblblue}{$\zmodal\gets(P^{\!\top}\!\otimes\Iden{m})\,z_t$}
      \Comment{into the modal basis}
    \State \textcolor{lblblue}{$\zmodal^{(k)}\gets\dk\,\zmodal^{(k)}$, $k=1,\dots,N$}
      \Comment{Thm.~\ref{thm:kcd}, one phase per mode}
    \State \textcolor{lblblue}{$z_{t+1}\gets(P\otimes\Iden{m})\,\zmodal$}\label{ln:back}
      \Comment{the $P$ of line~\ref{ln:eig}, identity to round-off}
    \State $\hat{x}_{t+1}\gets D_\psi(z_{t+1})$
  \EndFor
  \State \Return $\hat{x}_{1:T}$
\EndFunction
\end{algorithmic}
\end{algorithm}

%% file: Appendix_exp_10.16.tex
\newpage
\section{Additional Experiments}
\label{app:experiments}

\setcounter{topnumber}{4}\setcounter{bottomnumber}{3}\setcounter{totalnumber}{6}
\renewcommand{\topfraction}{0.92}\renewcommand{\bottomfraction}{0.8}
\renewcommand{\textfraction}{0.06}\renewcommand{\floatpagefraction}{0.8}
\setlength{\textfloatsep}{10pt plus 2pt minus 3pt}
\setlength{\floatsep}{8pt plus 2pt minus 2pt}
\setlength{\intextsep}{8pt plus 2pt minus 2pt}
\makeatletter 
\setlength{\@fptop}{0pt}\setlength{\@fpsep}{10pt plus 2pt}\setlength{\@fpbot}{0pt plus 1fil}
\makeatother

\subsection{Q1: Mass Is Held Where Every Diagonal Baseline Loses It}
\label{app:exp-q1}
\label{app:exp-tables}

\begin{figure}[!htbp]
\centering
\includegraphics[width=0.92\linewidth]{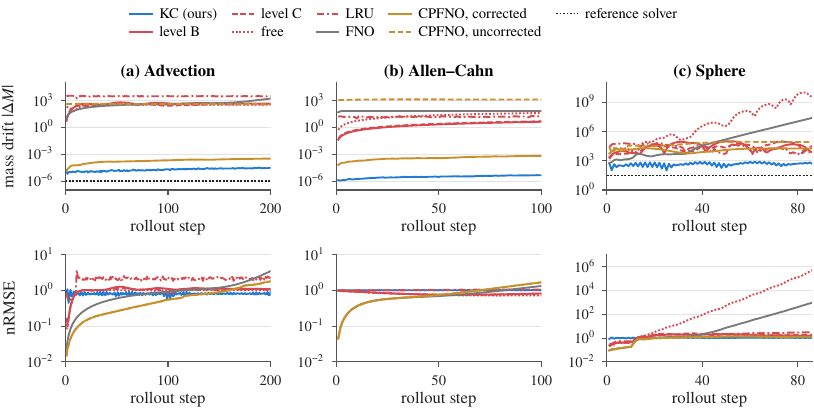}
\caption{\textbf{Drift and prediction error per rollout step.} Advection, Allen--Cahn and the
sphere, seeds $42$ to $44$, with the band spanning the three. The residual of KC rises smoothly
from the first step, the signature of accumulated round-off rather than of a learned bias, and
each corrector holds mass only while its uncorrected prediction stays bounded. FNO and CPFNO are
absent from the sphere panel, where the released trainer does not run, so that panel compares
the diagonal transitions alone.}
\label{fig:app-grid-perstep}
\end{figure}

\begingroup\looseness=-1
\noindent Table~\ref{tab:q1} summarises three seeds. Table~\ref{tab:app-q1-perseed} gives every
seed separately, so that the spread across seeds can be read against the gap between rows.
Table~\ref{tab:app-q1-zero} adds the drift of the latent zero mode, which is held at the same
order whether the decoder is tied or free, and locates the loss of physical mass at the
read-out. Table~\ref{tab:app-trivial} places every model against persistence and the constant
field with the spatial mean removed, which is the evidence for the accuracy cap discussed in
\S\ref{app:exp-diag}. Figure~\ref{fig:app-grid-perstep} then follows the drift and the
prediction error per rollout step, which shows that the residual of KC accumulates smoothly
rather than appearing at any one step.\par\endgroup

\begin{table}[!htbp]
\centering\scriptsize
\setlength{\tabcolsep}{5.5pt}
\renewcommand{\arraystretch}{0.95}
\setlength{\aboverulesep}{0.15ex}\setlength{\belowrulesep}{0.25ex}
\caption{\textbf{Mass drift and nRMSE at every seed.} Final-horizon values behind
Table~\ref{tab:q1}. A dash marks a drift the rollout does not define, CPFNO on the dam break at
seeds $43$ and $44$, where it leaves the axis at steps $48$ and $62$ and its drift at $T=100$ is
therefore read before divergence in Table~\ref{tab:app-q2-seeds} instead.}
\label{tab:app-q1-perseed}
\begin{tabular}{l@{\hspace{8pt}} rrr @{\hspace{11pt}} rrr}
\toprule
& \multicolumn{3}{c}{\textbf{Mass drift} $|\Delta M|$} & \multicolumn{3}{c}{\textbf{nRMSE}} \\
\textbf{Model} & seed 42 & seed 43 & seed 44 & seed 42 & seed 43 & seed 44 \\
\midrule
\tbandn{7}{dsband}{dsbtx}{Dam break}
level B & $27.1$ & $28.1$ & $28$ & $0.221$ & $0.23$ & $0.227$ \\
level C & $30.2$ & $61.9$ & $39.4$ & $0.223$ & $0.228$ & $0.229$ \\
free & $956$ & $93.8$ & $261$ & $0.166$ & $0.149$ & $0.155$ \\
LRU & $3.4\mathrm{e}{3}$ & $2.1\mathrm{e}{3}$ & $1.6\mathrm{e}{3}$ & $1.71$ & $0.519$ & $0.743$ \\
FNO & $3.6\mathrm{e}{6}$ & $2.3\mathrm{e}{2}$ & $5.4\mathrm{e}{11}$ & $1.8\mathrm{e}{3}$ & $0.105$ & $4.1\mathrm{e}{7}$ \\
CPFNO, uncorrected & $4.4\mathrm{e}{18}$ & --- & --- & $\infty$ & $\infty$ & $\infty$ \\
CPFNO, corrected & $1.3\mathrm{e}{6}$ & --- & --- & $\infty$ & $\infty$ & $\infty$ \\
\rowcolor{oursrow}[\tabcolsep][\tabcolsep]
KC, free decoder & $32$ & $403$ & $117$ & $0.226$ & $0.215$ & $0.221$ \\
\rowcolor{oursrow}[\tabcolsep][\tabcolsep]
\textbf{KC} & $9.3\mathrm{e}{-3}$ & $1.2\mathrm{e}{-3}$ & $0.0498$ & $0.114$ & $0.208$ & $0.114$ \\
\tbandn{7}{dsband}{dsbtx}{Advection}
level B & $457$ & $492$ & $483$ & $1.09$ & $1.11$ & $1.11$ \\
level C & $376$ & $357$ & $510$ & $0.998$ & $0.926$ & $1.14$ \\
free & $343$ & $456$ & $477$ & $1$ & $1.01$ & $1$ \\
LRU & $3.1\mathrm{e}{3}$ & $1.4\mathrm{e}{3}$ & $1.4\mathrm{e}{3}$ & $1.91$ & $1.49$ & $1.82$ \\
FNO & $5.8\mathrm{e}{4}$ & $898$ & $1.5\mathrm{e}{4}$ & $135$ & $2.52$ & $52.1$ \\
CPFNO, uncorrected & $398$ & $395$ & $395$ & $1.3$ & $1.89$ & $2.36$ \\
CPFNO, corrected & $3.3\mathrm{e}{-4}$ & $2.6\mathrm{e}{-4}$ & $2.6\mathrm{e}{-4}$ & $1.3$ & $1.89$ & $2.36$ \\
\rowcolor{oursrow}[\tabcolsep][\tabcolsep]
KC, free decoder & $6.2$ & $5.75$ & $10.7$ & $0.832$ & $0.507$ & $0.843$ \\
\rowcolor{oursrow}[\tabcolsep][\tabcolsep]
\textbf{KC} & $3.0\mathrm{e}{-5}$ & $0.0222$ & $0.0256$ & $0.924$ & $1.9$ & $2.55$ \\
\tbandn{7}{dsband}{dsbtx}{Allen--Cahn}
level B & $4.14$ & $6.49$ & $3.86$ & $0.818$ & $0.819$ & $0.803$ \\
level C & $4.65$ & $2.9$ & $2.6$ & $0.809$ & $0.816$ & $0.807$ \\
free & $39.5$ & $66.3$ & $59.8$ & $0.724$ & $0.69$ & $0.708$ \\
LRU & $17.9$ & $18.7$ & $19.3$ & $1.06$ & $1.06$ & $1.1$ \\
FNO & $66.5$ & $55$ & $62.2$ & $1.32$ & $2.04$ & $3.24$ \\
CPFNO, uncorrected & $1.2\mathrm{e}{3}$ & $5$ & $806$ & $1.66$ & $1.83$ & $2$ \\
CPFNO, corrected & $6.4\mathrm{e}{-4}$ & $6.3\mathrm{e}{-5}$ & $4.6\mathrm{e}{-4}$ & $1.66$ & $1.83$ & $2$ \\
\rowcolor{oursrow}[\tabcolsep][\tabcolsep]
KC, free decoder & $25.7$ & $52.3$ & $27.8$ & $1.94$ & $2.35$ & $1.92$ \\
\rowcolor{oursrow}[\tabcolsep][\tabcolsep]
\textbf{KC} & $4.4\mathrm{e}{-6}$ & $1.8\mathrm{e}{-5}$ & $1.0\mathrm{e}{-4}$ & $1.02$ & $0.84$ & $1.02$ \\
\tbandn{7}{dsband}{dsbtx}{Sphere}
level B & $5.9\mathrm{e}{4}$ & $3.7\mathrm{e}{4}$ & $9.0\mathrm{e}{3}$ & $1.75$ & $1.65$ & $1.65$ \\
level C & $1.3\mathrm{e}{4}$ & $5.1\mathrm{e}{4}$ & $2.1\mathrm{e}{4}$ & $1.6$ & $1.57$ & $1.68$ \\
free & $6.5\mathrm{e}{10}$ & $2.5\mathrm{e}{10}$ & $4.1\mathrm{e}{12}$ & $2.1\mathrm{e}{6}$ & $2.1\mathrm{e}{6}$ & $1.2\mathrm{e}{8}$ \\
LRU & $3.3\mathrm{e}{4}$ & $8.6\mathrm{e}{4}$ & $1.1\mathrm{e}{5}$ & $2.51$ & $3.5$ & $3.02$ \\
\rowcolor{oursrow}[\tabcolsep][\tabcolsep]
KC, free decoder & $561$ & $1.2\mathrm{e}{3}$ & $703$ & $1.03$ & $1.11$ & $1.23$ \\
\rowcolor{oursrow}[\tabcolsep][\tabcolsep]
\textbf{KC} & $337$ & $442$ & $759$ & $1.05$ & $1.03$ & $1.08$ \\
\bottomrule
\end{tabular}
\end{table}

\begin{table}[!htbp]
\centering\scriptsize
\setlength{\tabcolsep}{6pt}
\renewcommand{\arraystretch}{0.95}
\setlength{\aboverulesep}{0.15ex}\setlength{\belowrulesep}{0.25ex}
\caption{\textbf{Latent zero-mode drift at every seed.} The invariant the operator holds, before
the decoder. It stays between $10^{-6}$ and $10^{-2}$ under both read-outs, whereas the physical
mass separates them by three to six orders of magnitude on the three flat benchmarks.}
\label{tab:app-q1-zero}
\begin{tabular}{l@{\hspace{8pt}} rrr @{\hspace{11pt}} rrr}
\toprule
& \multicolumn{3}{c}{\textbf{KC}} & \multicolumn{3}{c}{\textbf{KC, free decoder}} \\
\textbf{Benchmark} & seed 42 & seed 43 & seed 44 & seed 42 & seed 43 & seed 44 \\
\midrule
Dam break & $7.8\mathrm{e}{-3}$ & $8.0\mathrm{e}{-5}$ & $7.1\mathrm{e}{-3}$ & $6.7\mathrm{e}{-3}$ & $2.1\mathrm{e}{-5}$ & $1.7\mathrm{e}{-3}$ \\
Advection & $4.5\mathrm{e}{-5}$ & $5.3\mathrm{e}{-3}$ & $6.7\mathrm{e}{-3}$ & $1.4\mathrm{e}{-4}$ & $2.3\mathrm{e}{-4}$ & $7.1\mathrm{e}{-4}$ \\
Allen--Cahn & $1.7\mathrm{e}{-5}$ & $3.2\mathrm{e}{-6}$ & $4.5\mathrm{e}{-5}$ & $5.1\mathrm{e}{-5}$ & $2.8\mathrm{e}{-6}$ & $7.4\mathrm{e}{-5}$ \\
Sphere & $8.3\mathrm{e}{-4}$ & $4.8\mathrm{e}{-4}$ & $1.5\mathrm{e}{-3}$ & $7.4\mathrm{e}{-4}$ & $6.7\mathrm{e}{-4}$ & $9.7\mathrm{e}{-4}$ \\
\bottomrule
\end{tabular}
\end{table}

\begin{table}[!htbp]
\centering\footnotesize
\setlength{\tabcolsep}{5.5pt}
\renewcommand{\arraystretch}{0.95}
\setlength{\aboverulesep}{0.15ex}\setlength{\belowrulesep}{0.25ex}
\caption{\textbf{Trivial baselines and the evolving-field error, seed $42$.} Persistence
outputs the first decoded frame at every step and the constant field is its spatial mean.
nRMSE$_{\text{ev}}$ removes the spatial mean, so a constant field scores $1$ by construction.}
\label{tab:app-trivial}
\begin{tabular}{l@{\hspace{8pt}} r@{\hspace{5pt}}r @{\hspace{11pt}} r@{\hspace{5pt}}r @{\hspace{11pt}} r@{\hspace{5pt}}r}
\toprule
& \multicolumn{2}{c}{\textbf{Dam-break}} & \multicolumn{2}{c}{\textbf{Advection}} & \multicolumn{2}{c}{\textbf{Allen--Cahn}} \\
\textbf{Model} & nRMSE & nRMSE$_{\text{ev}}$ & nRMSE & nRMSE$_{\text{ev}}$ & nRMSE & nRMSE$_{\text{ev}}$ \\
\midrule
\tbandn{7}{dsband}{dsbtx}{Trivial baselines}
persistence & $0.241$ & $2.32$ & $0.006$ & $0.009$ & $4.81$ & $4.82$ \\
constant field & $0.103$ & $1.00$ & $0.702$ & $1.00$ & $0.999$ & $1.00$ \\
\tbandn{7}{dsband}{dsbtx}{Diagonal recurrences}
level B & $0.221$ & $2.13$ & $1.09$ & $1.44$ & $\second{0.818}$ & $\second{0.818}$ \\
free & $\second{0.166}$ & $\second{1.51}$ & $\second{1.00}$ & $\best{1.00}$ & $\best{0.724}$ & $\best{0.724}$ \\
LRU & $1.71$ & $17.6$ & $1.91$ & $4.01$ & $1.06$ & $1.06$ \\
\tbandn{7}{dsband}{dsbtx}{Neural operators}
FNO & $1784$ & $1.8\mathrm{e}4$ & $3.44$ & $4.18$ & $1.32$ & $1.32$ \\
CPFNO & $\infty$ & $\infty$ & $1.78$ & $3.63$ & $1.66$ & $1.66$ \\
\midrule
\rowcolor{oursrow}[\tabcolsep][\tabcolsep]
\textbf{KC} & $\best{0.114}$ & $\best{1.10}$ & $\best{0.924}$ & $\second{1.23}$ & $1.02$ & $1.02$ \\
\bottomrule
\end{tabular}
\end{table}

\begin{figure}[!htbp]
\centering
\includegraphics[width=\linewidth]{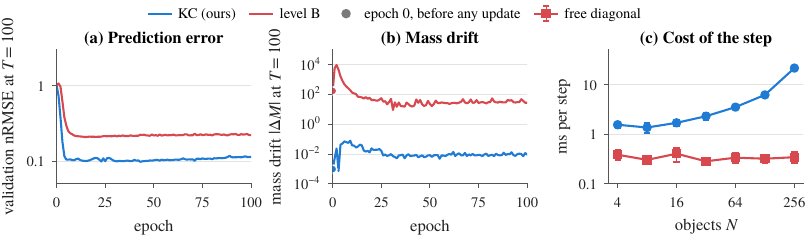}
\caption{\textbf{Training and cost.} Dam break, seed 42. \textbf{(a)} Validation nRMSE and
\textbf{(b)} mass drift at $T=100$ after every epoch, where the dot marks the untrained model.
KC starts at $10^{-3}$ before the first update and stays below $10^{-1}$, whereas level B
reaches $10^{4}$ at epoch 1 and settles at $27$. \textbf{(c)} Wall-clock per rollout step
against $N$: $3.5$\,ms at $N=64$ and $21$\,ms at $N=256$, against $0.3$\,ms for a free
diagonal.}
\label{fig:app-training-cost}
\end{figure}

\noindent\textbf{The spherical benchmark.}\;
\label{app:exp-sphere}
The PDEArena release distributes a $25$-trajectory subset whose pressure field is identically
zero at frame $0$, so a model taking that frame as input receives the same field on every
trajectory. Every model we trained under that protocol, diagonal or Fourier, collapsed to a
single latent, and its mass drift of exactly zero came from a constant output. We therefore
re-pose the benchmark as recorded in Table~\ref{tab:datasets}, and verify that no model collapses
by checking that the maximum pairwise latent distance across five test initial frames exceeds
$10^{-4}$ of the mean latent norm for every baseline. The physical mass on a sphere is the
$\cos(\text{lat})$-weighted integral, so a readout that assigns the unweighted patch sum to the
zero mode holds a different quantity: at seed $42$ it holds that mode to $4.7\times10^{-4}$ while
the physical mass drifts by $2.4\times10^{3}$, which is Corollary~\ref{cor:applic} in its negative
direction. Scaling each patch by the inverse of its mean $\cos(\text{lat})$ instead makes the zero
mode the area-weighted integral at patch resolution, held to $1.2\times10^{-2}$ against a
reference of $7.9\times10^{3}$. Under the per-pixel metric of Table~\ref{tab:q1}, which weights
every grid point by its own $\cos(\text{lat})$, the residual is $483$ over three seeds against a
reference of $30.8$, the variation of $\cos(\text{lat})$ across a patch spanning $22$ degrees of
latitude amplified near the poles by the $1/\cos$ scaling. CPFNO's corrector targets the
unweighted sum, so its corrected output drifts by $1.8\times10^{4}$ under that metric.

\FloatBarrier
\subsection{Q2: Correction at the Output Does Not Replace the Condition}
\label{app:exp-q2}

\begin{figure}[!htbp]
\centering
\includegraphics[width=\linewidth]{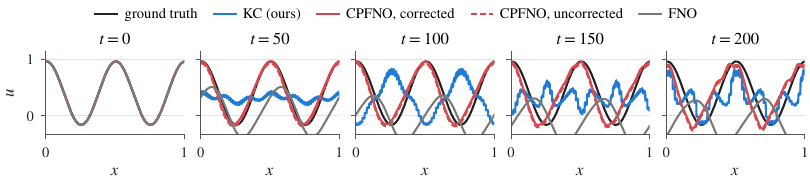}
\caption{\textbf{Advection, field at five steps, seed 42.} Ground truth, KC, CPFNO before and after its corrector, and FNO. KC holds the integral of $u$ to $3\times10^{-5}$ while its phase lags the reference, the cap of the per-patch encoder, whereas FNO loses amplitude and mass.}
\label{fig:app-advection-fields}
\end{figure}

\begin{figure}[!htbp]
\centering
\includegraphics[width=\linewidth]{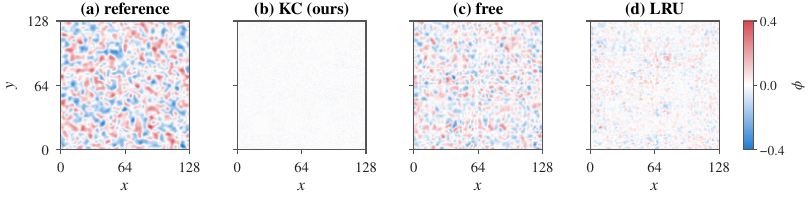}
\caption{\textbf{Allen--Cahn, field at the final horizon, seed 42.} Reference, KC, free diagonal and LRU on one colour scale. KC holds mass to $4\times10^{-6}$ with a near-constant field, whereas the free diagonal and LRU keep more texture while losing mass by $39.5$ and $17.9$.}
\label{fig:app-allencahn-fields}
\end{figure}

\noindent Figure~\ref{fig:chapman} shows the median over five dam-break seeds. The per-seed
values behind it are given in Table~\ref{tab:app-q2-seeds} for KC, CPFNO and FNO, and in
Table~\ref{tab:app-additive} for FNO with the fixed uniform corrector, where the last column
also re-targets CPFNO's corrector on the sphere to the area-weighted integral so that both
correctors and KC enforce the same quantity. Table~\ref{tab:app-q2-lines} then carries the two
neural operators to the remaining three benchmarks, and Figures~\ref{fig:app-advection-fields}
and \ref{fig:app-allencahn-fields} show the decoded fields on two of them. In every case the
corrector holds the quantity only while the uncorrected prediction stays bounded, which is the
conclusion of \S\ref{sec:exp-q2} on every benchmark reported here.

\begin{table}[!htbp]
\centering\footnotesize
\caption{\textbf{Dam-break shallow water across five seeds, $T=100$.} Mass drift, nRMSE, and
the first step at which nRMSE exceeds $100$. For CPFNO the drift is reported at $t\le50$, before
the median seed diverges.}
\label{tab:app-q2-seeds}
\setlength{\tabcolsep}{6pt}
\renewcommand{\arraystretch}{1.05}
\begin{tabular}{l ccccc}
\toprule
\rowcolor{hdr}[\tabcolsep][\tabcolsep]
\textbf{Metric} & \textbf{s42} & \textbf{s43} & \textbf{s44} & \textbf{s45} & \textbf{s46} \\
\midrule
\tgroup{6}{CPFNO}
corrected drift, $t\le50$ & $2\mathrm{e}{-3}$ & $2\mathrm{e}{-3}$ & $2\mathrm{e}{-3}$ & $2\mathrm{e}{-3}$ & $5.5\mathrm{e}{-3}$ \\
uncorrected drift, $t\le50$ & $1.6\mathrm{e}{4}$ & $1.6\mathrm{e}{4}$ & $1.6\mathrm{e}{4}$ & $1\mathrm{e}{4}$ & $1.1\mathrm{e}{4}$ \\
nRMSE & $\infty$ & $\infty$ & $\infty$ & $\infty$ & $\best{0.103}$ \\
diverges at & $83$ & $48$ & $62$ & $65$ & never \\
\tgroup{6}{FNO}
mass drift & $3.6\mathrm{e}{6}$ & $2.3\mathrm{e}{2}$ & $5.4\mathrm{e}{11}$ & $4.6\mathrm{e}{7}$ & $1.6\mathrm{e}{17}$ \\
nRMSE & $1783$ & $\best{0.105}$ & $4.1\mathrm{e}{7}$ & $5.7\mathrm{e}{3}$ & $3.1\mathrm{e}{13}$ \\
diverges at & $69$ & never & $94$ & $88$ & $83$ \\
\midrule
\rowcolor{oursrow}[\tabcolsep][\tabcolsep]
\textbf{KC}, mass drift & $\best{9.3\mathrm{e}{-3}}$ & $\best{1.2\mathrm{e}{-3}}$ & $\best{5.0\mathrm{e}{-2}}$ & $\best{2.0\mathrm{e}{-2}}$ & $\best{2.6\mathrm{e}{-2}}$ \\
\rowcolor{oursrow}[\tabcolsep][\tabcolsep]
\textbf{KC}, nRMSE & $\best{0.114}$ & $0.208$ & $\best{0.114}$ & $\best{0.115}$ & $0.115$ \\
\rowcolor{oursrow}[\tabcolsep][\tabcolsep]
\textbf{KC}, diverges at & never & never & never & never & never \\
\bottomrule
\end{tabular}
\end{table}

\begin{table}[!htbp]
\centering\footnotesize
\caption{\textbf{FNO with a fixed uniform corrector.} The corrector adds the missing mass
uniformly over the grid at inference, without training. Dam-break across five seeds
at $T=100$, the other three benchmarks at seed 42. The last column is CPFNO on the sphere with
a corrector that targets the $\cos(\text{lat})$-weighted integral.}
\label{tab:app-additive}
\setlength{\tabcolsep}{6pt}
\renewcommand{\arraystretch}{1.05}
\begin{tabular}{l ccccc}
\toprule
\rowcolor{hdr}[\tabcolsep][\tabcolsep]
\textbf{Metric} & \textbf{s42} & \textbf{s43} & \textbf{s44} & \textbf{s45} & \textbf{s46} \\
\midrule
\tgroup{6}{Dam-break, five seeds}
uncorrected drift & $6.7\mathrm{e}{5}$ & $4.0\mathrm{e}{14}$ & $3.1\mathrm{e}{6}$ & $2.7\mathrm{e}{24}$ & $1.3\mathrm{e}{18}$ \\
corrected drift & $6.0\mathrm{e}{-2}$ & $1.6\mathrm{e}{7}$ & $7.6\mathrm{e}{-2}$ & $1.2\mathrm{e}{17}$ & $2.2\mathrm{e}{10}$ \\
nRMSE & $403$ & $3.9\mathrm{e}{10}$ & $357$ & $\infty$ & $1.1\mathrm{e}{14}$ \\
diverges at & $95$ & $58$ & $97$ & $60$ & $73$ \\
\midrule
\rowcolor{hdr}[\tabcolsep][\tabcolsep]
& \textbf{Advection} & \textbf{Allen--Cahn} & \textbf{Sphere} & \textbf{Sphere, CPFNO} & \\
\midrule
\tgroup{6}{Other benchmarks, seed 42}
uncorrected drift & $451$ & $7.5\mathrm{e}{3}$ & $4.1\mathrm{e}{4}$ & $1.8\mathrm{e}{4}$ & \\
corrected drift & $2.3\mathrm{e}{-4}$ & $2.2\mathrm{e}{-3}$ & $4.0\mathrm{e}{-2}$ & $1.3\mathrm{e}{-2}$ & \\
nRMSE & $2.10$ & $1.81$ & $1.72$ & $1.55$ & \\
\bottomrule
\end{tabular}
\end{table}

\begin{table}[!htbp]
\centering\footnotesize
\caption{\textbf{CPFNO and FNO on the remaining three benchmarks, seed 42.} Advection values
are deterministic.}
\label{tab:app-q2-lines}
\setlength{\tabcolsep}{6pt}
\renewcommand{\arraystretch}{1.05}
\begin{tabular}{l ccc}
\toprule
\rowcolor{hdr}[\tabcolsep][\tabcolsep]
\textbf{Metric} & \textbf{Advection} & \textbf{Allen--Cahn} & \textbf{Sphere} \\
\midrule
\tgroup{4}{CPFNO}
corrected drift & $3.2\mathrm{e}{-4}$ & held & $1.8\mathrm{e}{4}$ \\
uncorrected drift & $391$ & $1290$ & $8.3\mathrm{e}{4}$ \\
nRMSE & $1.78$ & $1.66$ & $1.29$ \\
\tgroup{4}{FNO}
mass drift & $1.6\mathrm{e}{3}$ & $66.5$ & $2.4\mathrm{e}{7}$ \\
nRMSE & $3.44$ & $1.32$ & diverges \\
\bottomrule
\end{tabular}
\end{table}

\FloatBarrier
\subsection{Q3: Each Condition Holds Exactly One Quantity}
\label{app:exp-q3}

\begin{figure}[!htbp]
\centering
\includegraphics[width=\linewidth]{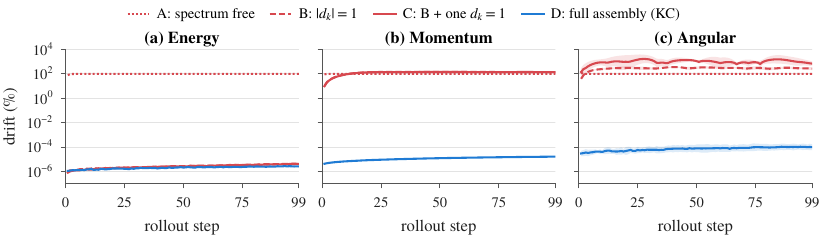}
\caption{\textbf{Drift per rollout step under the four levels.} Visual $N$-body, seeds
42/43/44, geometric mean with one standard deviation in log space shaded. Energy is held from
level B onward, momentum and the angular pairing only by the full assembly.}
\label{fig:app-layering-perstep}
\end{figure}

\begin{figure}[!htbp]
\centering
\includegraphics[width=0.90\linewidth]{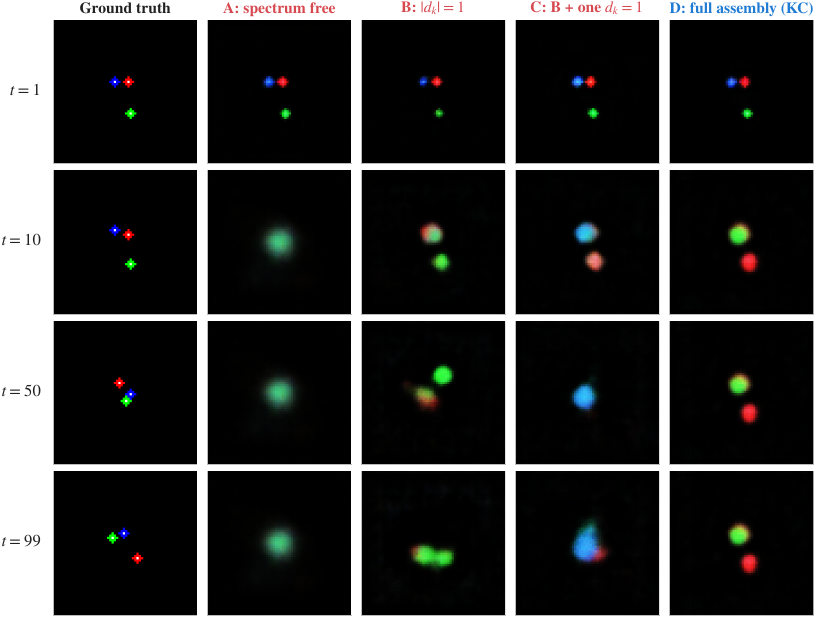}
\caption{\textbf{Decoded frames of the visual $N$-body rollout under the four levels.} Test
trajectory 0, seed 42, at four steps. None of the four follows the reference positions under the
shared encoder (\S\ref{app:exp-diag}). Level~A collapses to a single blob by $t=10$, levels~B
and~C merge objects and exchange their colours, and the full assembly keeps separate,
colour-stable objects over the full horizon.}
\label{fig:app-visual-frames}
\end{figure}

\noindent Figure~\ref{fig:layering} and Table~\ref{tab:q3} report the three controls as means
over three seeds. The per-seed values of the four levels are given in
Table~\ref{tab:app-q3-levels}, followed per step in Figure~\ref{fig:app-layering-perstep} and
decoded into frames in Figure~\ref{fig:app-visual-frames}. The matched perturbation is tabulated
at every $\varepsilon$ in Table~\ref{tab:perturb}, and Table~\ref{tab:app-q3-readout}
isolates the read-out direction by holding the operator and its spectrum fixed.

\begin{table}[!htbp]
\centering\footnotesize
\caption{\textbf{Four levels of the construction, per seed.} Visual $N$-body, drift at step
$99$ for seeds $42$/$43$/$44$. Levels A to D are the spectrum free, unit modulus on every
eigenvalue, B with one mode per object fixed at unity, and the full assembly.}
\label{tab:app-q3-levels}
\setlength{\tabcolsep}{2pt}
\renewcommand{\arraystretch}{1.05}
\scriptsize
\begin{tabular}{@{}l ccc l l l@{}}
\toprule
\rowcolor{hdr}[0pt][0pt]
& \multicolumn{3}{c}{\textbf{Condition present}} & \multicolumn{3}{c}{\textbf{Drift at step 99 (\%), seeds 42 / 43 / 44}} \\
\rowcolor{hdr}[0pt][0pt]
\textbf{Level} & $|\dk|{=}1$ & $\Lop\onevec{=}0$ & $\Lop{\otimes}\Iden{m}$ & energy & momentum & angular \\
\midrule
A & \mkno & \mkno & \mkno & $100$, $100$, $100$ & $100$, $100$, $100$ & $100$, $100$, $100$ \\
B & \mkyes & \mkno & \mkno & $4.0\mathrm{e}{-6}$, $4.1\mathrm{e}{-6}$, $5.1\mathrm{e}{-6}$ & $145$, $142$, $135$ & $411$, $225$, $222$ \\
C & \mkyes & \mkno & \mkno & $4.1\mathrm{e}{-6}$, $4.5\mathrm{e}{-6}$, $4.5\mathrm{e}{-6}$ & $138$, $138$, $140$ & $1060$, $786$, $412$ \\
\midrule
\rowcolor{oursrow}[0pt][0pt]
D (KC) & \mkyes & \mkyes & \mkyes & $\best{2.7\mathrm{e}{-6}}$, $\best{2.4\mathrm{e}{-6}}$, $\best{3.0\mathrm{e}{-6}}$ & $\best{1.7\mathrm{e}{-5}}$, $\best{1.6\mathrm{e}{-5}}$, $\best{1.9\mathrm{e}{-5}}$ & $\best{7.8\mathrm{e}{-5}}$, $\best{6.1\mathrm{e}{-5}}$, $\best{2.2\mathrm{e}{-4}}$ \\
\bottomrule
\end{tabular}
\end{table}

\begin{table}[!htbp]
\centering\footnotesize
\caption{\textbf{Matched perturbation, visual $N$-body, step 25, seeds $42$ to $44$.} One
symmetric draw of scale $\varepsilon$, applied once with the diagonal corrected so that the row
sums stay zero and once uncorrected, with the relative Frobenius change beside each.}
\label{tab:perturb}
\setlength{\tabcolsep}{6pt}
\renewcommand{\arraystretch}{1.05}
\begin{tabular}{l ccc ccc}
\toprule
\rowcolor{hdr}[\tabcolsep][\tabcolsep]
& \multicolumn{3}{c}{\textbf{row sums kept}} & \multicolumn{3}{c}{\textbf{row sums broken}} \\
\rowcolor{hdr}[\tabcolsep][\tabcolsep]
$\varepsilon$ & $\|\Delta\Lop\|_F/\|\Lop\|_F$ & momentum & energy & $\|\Delta\Lop\|_F/\|\Lop\|_F$ & momentum & energy \\
\midrule
$0$    & $0$ & $2.3\mathrm{e}{-5}$ & $1.7\mathrm{e}{-6}$ & $0$ & $2.3\mathrm{e}{-5}$ & $1.7\mathrm{e}{-6}$ \\
$0.01$ & $4.4\mathrm{e}{-3}$ & $3.3\mathrm{e}{-5}$ & $1.8\mathrm{e}{-6}$ & $4.6\mathrm{e}{-3}$ & $4.06$ & $1.9\mathrm{e}{-6}$ \\
$0.05$ & $2.2\mathrm{e}{-2}$ & $3.3\mathrm{e}{-5}$ & $1.7\mathrm{e}{-6}$ & $2.3\mathrm{e}{-2}$ & $20.2$ & $1.7\mathrm{e}{-6}$ \\
$0.1$  & $4.4\mathrm{e}{-2}$ & $3.3\mathrm{e}{-5}$ & $1.8\mathrm{e}{-6}$ & $4.6\mathrm{e}{-2}$ & $39.5$ & $1.9\mathrm{e}{-6}$ \\
$0.2$  & $8.7\mathrm{e}{-2}$ & $3.2\mathrm{e}{-5}$ & $1.8\mathrm{e}{-6}$ & $9.1\mathrm{e}{-2}$ & $72.7$ & $1.9\mathrm{e}{-6}$ \\
$0.5$  & $2.2\mathrm{e}{-1}$ & $3.5\mathrm{e}{-5}$ & $1.8\mathrm{e}{-6}$ & $2.3\mathrm{e}{-1}$ & $109.7$ & $2.0\mathrm{e}{-6}$ \\
\bottomrule
\end{tabular}
\end{table}

\begin{table}[!htbp]
\centering\footnotesize
\caption{\textbf{Same operator, different read-out, 2000 steps, seeds 42/43/44.} Maximum drift
in percent when the summed quantity is read along the null-mode eigenvector against the
eigenvector of largest $|\lambda|$, with the operator and its spectrum held fixed.}
\label{tab:app-q3-readout}
\setlength{\tabcolsep}{8pt}
\renewcommand{\arraystretch}{1.05}
\begin{tabular}{l cc}
\toprule
\rowcolor{hdr}[\tabcolsep][\tabcolsep]
\textbf{System} & \textbf{null mode} & \textbf{non-null mode} \\
\midrule
synthetic graph, $N=3$ & $\best{3.98\mathrm{e}{-3}\pm3.95\mathrm{e}{-3}}$ & $200.0\pm2.4\mathrm{e}{-5}$ \\
molecular dynamics, $N=16$ & $\best{3.32\mathrm{e}{-4}\pm2.85\mathrm{e}{-4}}$ & $247.4\pm43.8$ \\
\bottomrule
\end{tabular}
\end{table}

\FloatBarrier
\subsection{Q4: Training Reaches the Condition Only by Freezing the Spectrum}
\label{app:exp-q4}

\noindent Figure~\ref{fig:alignment} summarises sixteen trained seeds. The angle of each basis
to the uniform vector before and after training is given in Table~\ref{tab:app-q4-align} for
the near-identity initialisation and in Table~\ref{tab:app-q4-haar} for the Haar-random one.
Table~\ref{tab:app-q4-mechanism} then
reports the spectral activity of every seed and, for three representative seeds, the two
factors of Equation~\ref{eq:conservingset}, which shows that the two conserving seeds satisfy
the condition by driving every $d_k$ to $1$. Tables~\ref{app:exp-penalty} and
\ref{tab:app-q4-converge} give the penalty sweep and the sample-size sweep behind
Figures~\ref{fig:alignment}b and \ref{fig:alignment}c, where no penalty weight reaches the
condition and KC holds it at every training-set size.

\begin{table}[!htbp]
\centering\footnotesize
\caption{\textbf{Angle of a learned basis to the uniform vector, eight seeds.} $\varphi$ at
initialisation and after training under a conservation penalty at $\lambda=10$, with momentum
drift at step $25$. The basis is initialised near the identity, at which $\varphi$ takes its
maximum $\arccos(1/\sqrt3)=54.74^\circ$.}
\label{tab:app-q4-align}
\setlength{\tabcolsep}{5.5pt}
\renewcommand{\arraystretch}{1.05}
\begin{tabular}{l rrrrrrrr r}
\toprule
\rowcolor{hdr}[\tabcolsep][\tabcolsep]
\textbf{seed} & 42 & 43 & 44 & 45 & 46 & 47 & 48 & 49 & \textbf{assembled} \\
\midrule
$\varphi$, init  & $54.1^\circ$ & $54.2^\circ$ & $54.3^\circ$ & $54.5^\circ$ & $53.6^\circ$ & $54.3^\circ$ & $53.2^\circ$ & $54.3^\circ$ & --- \\
$\varphi$, final & $44.9^\circ$ & $48.1^\circ$ & $54.7^\circ$ & $54.7^\circ$ & $54.7^\circ$ & $54.7^\circ$ & $54.7^\circ$ & $54.7^\circ$ & \cellcolor{oursrow}$\best{0.01^\circ}$ \\
momentum drift   & $159$ & $932$ & $174$ & $43.9$ & $174$ & $44.0$ & $174$ & $44.4$ & \cellcolor{oursrow}$\best{10^{-5}}$ \\
\bottomrule
\end{tabular}
\end{table}

\begin{table}[!htbp]
\centering\footnotesize
\caption{\textbf{The same test under Haar-random initialisation, eight seeds.} $\varphi$ at
initialisation and after training at $\lambda=10$, with momentum drift at step $25$. Seeds
$47$ and $49$ conserve, by freezing the spectrum (Table~\ref{tab:app-q4-mechanism}).}
\label{tab:app-q4-haar}
\setlength{\tabcolsep}{8pt}
\renewcommand{\arraystretch}{1.05}
\begin{tabular}{l rrrrrrrr}
\toprule
\rowcolor{hdr}[\tabcolsep][\tabcolsep]
\textbf{seed} & 42 & 43 & 44 & 45 & 46 & 47 & 48 & 49 \\
\midrule
$\varphi$, init  & $26.3^\circ$ & $41.8^\circ$ & $42.2^\circ$ & $44.8^\circ$ & $39.1^\circ$ & $35.4^\circ$ & $36.4^\circ$ & $37.0^\circ$ \\
$\varphi$, final & $29.3^\circ$ & $42.0^\circ$ & $42.1^\circ$ & $44.9^\circ$ & $37.0^\circ$ & $32.5^\circ$ & $37.5^\circ$ & $35.6^\circ$ \\
momentum drift   & $137$ & $373$ & $401$ & $329$ & $127$ & $\best{0.090}$ & $88.3$ & $\second{0.196}$ \\
\bottomrule
\end{tabular}
\end{table}

\begin{table}[!htbp]
\centering\footnotesize
\caption{\textbf{Spectral activity and the conservation condition.} Above, mean $|\lambda_k|$
over the modes that do not carry the uniform vector, for all sixteen trained seeds. Below, the
two factors of Equation~\ref{eq:conservingset} for three representative seeds. The conserving
seeds satisfy the condition by driving every $d_k$ to $1$.}
\label{tab:app-q4-mechanism}
\setlength{\tabcolsep}{5pt}
\renewcommand{\arraystretch}{1.05}
\begin{tabular}{ll rr}
\toprule
\rowcolor{hdr}[\tabcolsep][\tabcolsep]
\textbf{init} & \textbf{seed} & \textbf{mean $|\lambda_k|$} & \textbf{drift (\%)} \\
\midrule
\tgroup{4}{Free orthogonal basis, trained under the penalty}
Haar & 42, 43, 44 & $2.85$, $2.83$, $2.85$ & $137$, $373$, $401$ \\
Haar & 45, 46, 48 & $1.13$, $1.80$, $1.74$ & $329$, $127$, $88.3$ \\
Haar & 47, 49 & $2.8\mathrm{e}{-5}$, $1.3\mathrm{e}{-4}$ & $\second{0.090}$, $\second{0.196}$ \\
near-identity & 42, 43 & $2.18$, $1.83$ & $159$, $932$ \\
near-identity & 44, 46, 48 & $1.59$, $1.59$, $1.59$ & $174$, $174$, $174$ \\
near-identity & 45, 47, 49 & $0.77$, $0.77$, $0.77$ & $43.9$, $43.9$, $44.5$ \\
\midrule
\rowcolor{oursrow}[\tabcolsep][\tabcolsep]
assembled & 42 & $\best{2.87}$ & $\best{3.5\mathrm{e}{-5}}$ \\
\bottomrule
\end{tabular}

\medskip
\begin{tabular}{l ccc ccc ccc}
\toprule
\rowcolor{hdr}[\tabcolsep][\tabcolsep]
& \multicolumn{3}{c}{$d_k$} & \multicolumn{3}{c}{$c_k$} & \multicolumn{3}{c}{$|(d_k-1)c_k|$} \\
\midrule
Haar 47 & $1$ & $1$ & $1$ & $0.46$ & $0.27$ & $-0.84$ & $10^{-5}$ & $10^{-5}$ & $10^{-5}$ \\
Haar 42 & $e^{1.92\mathrm{i}}$ & $e^{1.92\mathrm{i}}$ & $e^{0.74\mathrm{i}}$ & $0.35$ & $-0.35$ & $0.87$ & $0.57$ & $0.57$ & $0.63$ \\
\midrule
\rowcolor{oursrow}[\tabcolsep][\tabcolsep]
assembled & $e^{1.92\mathrm{i}}$ & $e^{1.92\mathrm{i}}$ & $1$ & $0$ & $0$ & $1$ & $0$ & $0$ & $0$ \\
\bottomrule
\end{tabular}
\end{table}

\begin{table}[!htbp]
\centering\footnotesize
\setlength{\tabcolsep}{5pt}
\renewcommand{\arraystretch}{1.05}
\begin{minipage}[t]{0.44\linewidth}
\caption{\textbf{Penalty sweep.} Momentum drift at step $25$, per seed.}
\label{app:exp-penalty}
\end{minipage}\hfill
\begin{minipage}[t]{0.54\linewidth}
\caption{\textbf{Learned conservation at convergence.} Momentum drift at step $25$
against training-set size.}
\label{tab:app-q4-converge}
\end{minipage}

\smallskip
\begin{minipage}[t]{0.44\linewidth}
\centering
\begin{tabular}[t]{l ccc c}
\toprule
\rowcolor{hdr}[\tabcolsep][\tabcolsep]
$\lambda$ & \textbf{42} & \textbf{43} & \textbf{44} & \textbf{PSNR (dB)} \\
\midrule
$0$   & $\best{86.5}$  & $\best{99.2}$  & $\best{42.0}$  & $18.11$ \\
$0.1$ & $173.7$ & $\second{388.8}$ & $173.6$ & $\best{19.37}$ \\
$1$   & $177.4$ & $436.5$ & $173.6$ & $19.35$ \\
$10$  & $\second{159.3}$ & $932.0$ & $173.6$ & $\second{19.36}$ \\
\bottomrule
\end{tabular}
\end{minipage}\hfill
\begin{minipage}[t]{0.54\linewidth}
\centering
\begin{tabular}[t]{l ccc c}
\toprule
\rowcolor{hdr}[\tabcolsep][\tabcolsep]
$n_{\text{train}}$ & \textbf{KC} & \textbf{HNN} & \textbf{GNN} & \textbf{PSNR gap} \\
\midrule
$50$   & \cellcolor{oursrow}$\best{1.29\mathrm{e}{-5}}$ & $\second{16.9}$ & $247$ & $-0.31$ \\
$100$  & \cellcolor{oursrow}$\best{1.30\mathrm{e}{-5}}$ & $\second{12.1}$ & $247$ & $-0.26$ \\
$250$  & \cellcolor{oursrow}$\best{1.30\mathrm{e}{-5}}$ & $\second{6.46}$ & $246$ & $-0.22$ \\
$500$  & \cellcolor{oursrow}$\best{1.30\mathrm{e}{-5}}$ & $\second{2.63}$ & $246$ & $-0.12$ \\
$1000$ & \cellcolor{oursrow}$\best{1.38\mathrm{e}{-5}}$ & $\second{1.49}$ & $248$ & $-0.08$ \\
$5000$ & \cellcolor{oursrow}$\best{1.41\mathrm{e}{-5}}$ & $\second{1.52}$ & $248$ & $-0.10$ \\
\bottomrule
\end{tabular}
\end{minipage}
\end{table}

\FloatBarrier
\subsection{Q5: The Boundary Is Fixed Before Training}
\label{app:exp-q5}

\begin{figure}[!htbp]
\centering
\includegraphics[width=\linewidth]{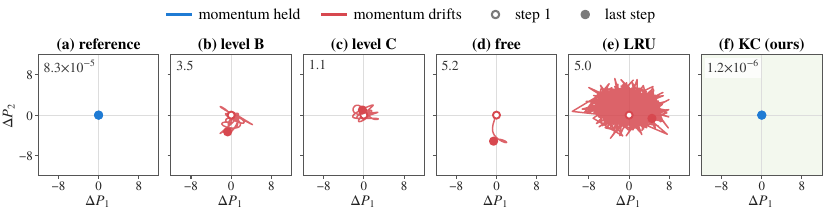}
\caption{\textbf{Total linear momentum of the charged $N$-body rollout, one test trajectory,
seed $42$.} Each path is $P(t)-P(1)$ over $1000$ steps in the plane carrying most of its own
variation, so distance from the origin is the drift. Each panel states $\|P(1000)-P(1)\|$
against $\|P(1)\|=1.20$. The momentum stays at a point under the reference integrator and the
full assembly, and leaves it under every transition that constrains the spectrum alone.}
\label{fig:app-nbody-momentum}
\end{figure}

\noindent Table~\ref{tab:q5-nbody} reports linear and angular momentum on the three $N$-body
systems and Table~\ref{tab:q5-cross} the four operator-level systems from other fields, the
set behind Figure~\ref{fig:crossdomain} of the main text. The decoded total momentum of one
charged test trajectory is drawn per transition in Figure~\ref{fig:app-nbody-momentum}. The reference
floor of each system is given in Table~\ref{tab:app-q5-floor}, which shows that the integrator
itself conserves energy only on the spring system, and Table~\ref{tab:app-q5-nbody} therefore
adds the energy column on that system alone. Table~\ref{app:exp-outside} then tests the quantity boundary inside the latent, by comparing the
zero mode of the family with five random Hermitian forms outside it under the same trained
operator, and Table~\ref{tab:app-q5-robertson} tests the dynamics boundary on a stiff system
that relaxes to equilibrium, where Proposition~\ref{prop:bound} predicts that a unitary step
fits less accurately than one that can contract. Finally, Table~\ref{app:exp-spring} reports the
spring benchmark of \citet{thangamuthu2022unravelling} on its own three metrics.

\begin{table}[!htbp]
\centering\footnotesize
\setlength{\tabcolsep}{9pt}
\renewcommand{\arraystretch}{0.95}
\setlength{\aboverulesep}{0.15ex}\setlength{\belowrulesep}{0.25ex}
\caption{\textbf{$N$-body conservation at $T=1000$, seeds $42$ to $44$.} Drift in absolute
units, geometric mean over the seeds that stayed finite. An asterisk marks a row where the free
diagonal left the axis on one seed.}
\label{tab:q5-nbody}
\begin{tabular}{l rrrr rr}
\toprule
\textbf{System} & \textbf{B} & \textbf{C} & \textbf{free} & \textbf{LRU} & \cellcolor{oursrow}\textbf{KC-fd} & \cellcolor{oursrow}\textbf{KC} \\
\midrule
\tbandn{7}{dsband}{dsbtx}{Linear momentum}
charged & $3.29$ & $\second{2.66}$ & $1.9\mathrm{e}4$ & $6.41$ & \cellcolor{oursrow}$3.14$ & \cellcolor{oursrow}$\best{2.2\mathrm{e}{-6}}$ \\
gravity & $4.06$ & $\second{3.92}$ & $4.7\mathrm{e}3^{*}$ & $4.70$ & \cellcolor{oursrow}$7.28$ & \cellcolor{oursrow}$\best{3.6\mathrm{e}{-6}}$ \\
spring & $\second{1.25}$ & $1.36$ & $5.64$ & $3.79$ & \cellcolor{oursrow}$1.49$ & \cellcolor{oursrow}$\best{1.6\mathrm{e}{-6}}$ \\
\tbandn{7}{dsband}{dsbtx}{Angular momentum}
charged & $10.4$ & $\second{9.10}$ & $20.0^{*}$ & $26.1$ & \cellcolor{oursrow}$13.7$ & \cellcolor{oursrow}$\best{1.22}$ \\
gravity & $17.4$ & $\second{15.7}$ & $5.7\mathrm{e}7^{*}$ & $22.8$ & \cellcolor{oursrow}$20.0$ & \cellcolor{oursrow}$\best{1.31}$ \\
spring & $\second{0.978}$ & $1.77$ & $6.06$ & $4.40$ & \cellcolor{oursrow}$1.60$ & \cellcolor{oursrow}$\best{0.327}$ \\
\tbandn{7}{dsband}{dsbtx}{Energy}
spring & $\second{0.323}$ & $0.369$ & $12.6$ & $1.53$ & \cellcolor{oursrow}$0.927$ & \cellcolor{oursrow}$\best{0.124}$ \\
\bottomrule
\end{tabular}
\end{table}

\begin{table}[!htbp]
\centering\footnotesize
\setlength{\tabcolsep}{8pt}
\renewcommand{\arraystretch}{0.95}
\setlength{\aboverulesep}{0.15ex}\setlength{\belowrulesep}{0.25ex}
\caption{\textbf{Transfer to four dynamical systems, operator level, $2000$ steps, seeds
$42$/$43$/$44$.} Time-integrated mean drift of the conserved quantity in percent, with the row
sums of the Laplacian intact and broken by a matched perturbation
(Figure~\ref{fig:crossdomain}).}
\label{tab:q5-cross}
\begin{tabular}{l l cc}
\toprule
\textbf{System} & \textbf{Conserved quantity} & \textbf{intact}$\downarrow$ & \textbf{broken} \\
\midrule
molecular dynamics, $N{=}16$ & total momentum & $\best{1.4\mathrm{e}{-4}}$ & $100.5\pm0.5$ \\
reaction network, $N{=}10$ & total mass & $\best{6.3\mathrm{e}{-3}}$ & $126.9\pm0.3$ \\
pharmacokinetics, $N{=}4$ & total drug amount & $\best{1.0\mathrm{e}{-3}}$ & $127.8\pm0.8$ \\
power-grid swing, $N{=}12$ & angular momentum & $\best{5.3\mathrm{e}{-3}}$ & $127.0\pm0.3$ \\
\bottomrule
\end{tabular}
\end{table}

\begin{table}[!htbp]
\centering\footnotesize
\caption{\textbf{Reference conservation on the three $N$-body systems, 20 runs.} Absolute drift
over the full trajectory, with the initial magnitude in parentheses.}
\label{tab:app-q5-floor}
\setlength{\tabcolsep}{6pt}
\renewcommand{\arraystretch}{1.05}
\begin{tabular}{l ccc}
\toprule
\rowcolor{hdr}[\tabcolsep][\tabcolsep]
\textbf{System} & \textbf{Energy} & \textbf{Linear momentum} & \textbf{Angular momentum} \\
\midrule
Charged, $N=5$, 3D & $1.26$ ($0.69$) & $1.7\mathrm{e}{-4}$ ($1.05$) & $0.018$ ($1.47$) \\
Gravity, $N=10$, 3D & $7.2$ ($13.0$) & $4.0\mathrm{e}{-14}$ ($\approx0$) & $7.9\mathrm{e}{-14}$ ($6.58$) \\
Spring, $N=5$, 2D & $9.8\mathrm{e}{-5}$ ($0.63$) & $3.4\mathrm{e}{-15}$ ($1.09$) & $2.4\mathrm{e}{-14}$ ($0.43$) \\
\bottomrule
\end{tabular}
\end{table}

\begin{table}[!htbp]
\centering\scriptsize
\setlength{\tabcolsep}{4pt}
\renewcommand{\arraystretch}{0.95}
\setlength{\aboverulesep}{0.15ex}\setlength{\belowrulesep}{0.25ex}
\caption{\textbf{$N$-body drift at $T=1000$, every seed.} Values behind
Table~\ref{tab:q5-nbody}. A dash in the first two blocks marks a seed on which the free diagonal
left the axis, and energy is reported on the spring system alone, the one whose reference
conserves it.}
\label{tab:app-q5-nbody}
\begin{tabular}{l@{\hspace{6pt}} rrr @{\hspace{8pt}} rrr @{\hspace{8pt}} rrr}
\toprule
& \multicolumn{3}{c}{\textbf{charged}} & \multicolumn{3}{c}{\textbf{gravity}} & \multicolumn{3}{c}{\textbf{spring}} \\
\textbf{Model} & s42 & s43 & s44 & s42 & s43 & s44 & s42 & s43 & s44 \\
\midrule
\tbandn{10}{dsband}{dsbtx}{Linear momentum}
level B & $3.32$ & $3.13$ & $3.42$ & $4.11$ & $4.26$ & $3.81$ & $1.21$ & $1.38$ & $1.16$ \\
level C & $3.13$ & $2.31$ & $2.61$ & $3.83$ & $4.43$ & $3.54$ & $1.03$ & $1.66$ & $1.46$ \\
free & $3.56$ & $4.00$ & $5.1\mathrm{e}{11}$ & --- & $3.30$ & $6.7\mathrm{e}{6}$ & $3.47$ & $7.03$ & $7.38$ \\
LRU & $7.61$ & $6.24$ & $5.55$ & $4.15$ & $4.78$ & $5.22$ & $3.93$ & $4.44$ & $3.13$ \\
\rowcolor{oursrow}[\tabcolsep][\tabcolsep]
KC, free decoder & $2.27$ & $4.31$ & $3.15$ & $6.25$ & $7.74$ & $7.96$ & $0.514$ & $2.69$ & $2.38$ \\
\rowcolor{oursrow}[\tabcolsep][\tabcolsep]
\textbf{KC} & $1.9\mathrm{e}{-6}$ & $2.1\mathrm{e}{-6}$ & $2.6\mathrm{e}{-6}$ & $3.0\mathrm{e}{-6}$ & $4.3\mathrm{e}{-6}$ & $3.5\mathrm{e}{-6}$ & $8.6\mathrm{e}{-7}$ & $2.5\mathrm{e}{-6}$ & $1.8\mathrm{e}{-6}$ \\
\tbandn{10}{dsband}{dsbtx}{Angular momentum}
level B & $11.5$ & $8.84$ & $10.9$ & $14.8$ & $19.9$ & $18.0$ & $0.867$ & $1.26$ & $0.859$ \\
level C & $10.4$ & $9.28$ & $7.82$ & $18.3$ & $11.6$ & $18.4$ & $1.43$ & $1.36$ & $2.89$ \\
free & $16.7$ & $24.0$ & --- & --- & $32.8$ & $9.7\mathrm{e}{13}$ & $3.44$ & $2.85$ & $22.7$ \\
LRU & $24.7$ & $28.0$ & $25.7$ & $15.6$ & $16.0$ & $47.1$ & $2.72$ & $6.20$ & $5.05$ \\
\rowcolor{oursrow}[\tabcolsep][\tabcolsep]
KC, free decoder & $18.6$ & $14.0$ & $9.89$ & $16.3$ & $22.1$ & $22.3$ & $0.741$ & $2.60$ & $2.12$ \\
\rowcolor{oursrow}[\tabcolsep][\tabcolsep]
\textbf{KC} & $1.77$ & $1.37$ & $0.753$ & $3.53$ & $0.757$ & $0.846$ & $0.429$ & $0.309$ & $0.263$ \\
\tbandn{10}{dsband}{dsbtx}{Energy, spring system only}
level B & \multicolumn{6}{c}{---} & $0.259$ & $0.431$ & $0.300$ \\
level C & \multicolumn{6}{c}{---} & $0.305$ & $0.404$ & $0.409$ \\
free & \multicolumn{6}{c}{---} & $5.13$ & $17.9$ & $21.8$ \\
LRU & \multicolumn{6}{c}{---} & $1.57$ & $1.51$ & $1.52$ \\
\rowcolor{oursrow}[\tabcolsep][\tabcolsep]
KC, free decoder & \multicolumn{6}{c}{---} & $0.143$ & $2.78$ & $2.00$ \\
\rowcolor{oursrow}[\tabcolsep][\tabcolsep]
\textbf{KC} & \multicolumn{6}{c}{---} & $0.0928$ & $0.222$ & $0.0927$ \\
\bottomrule
\end{tabular}
\end{table}

\begin{table}[!htbp]
\centering\scriptsize
\renewcommand{\arraystretch}{1.05}
\begin{minipage}[t]{0.44\linewidth}
\caption{\textbf{Family and outside-family drift at $T=1000$, KC, seed 42.} Absolute drift of
the latent zero mode, which lies in the family of Corollary~\ref{cor:family}, beside five random
Hermitian forms outside it, under the same trained operator.}
\label{app:exp-outside}
\end{minipage}\hfill
\begin{minipage}[t]{0.53\linewidth}
\caption{\textbf{The dynamics boundary on the Robertson reaction network, $T=50$.} A stiff
system that relaxes to equilibrium, on which Proposition~\ref{prop:bound} bounds the per-step
error of any unitary step. Means over seeds $42$/$43$/$44$. The latent zero mode is held, and
the baselines that can contract fit the state more accurately.}
\label{tab:app-q5-robertson}
\end{minipage}

\smallskip
\begin{minipage}[t]{0.44\linewidth}
\centering
\setlength{\tabcolsep}{3pt}
\begin{tabular}[t]{l ccc}
\toprule
\rowcolor{hdr}[\tabcolsep][\tabcolsep]
\textbf{Quadratic form} & \textbf{charged} & \textbf{gravity} & \textbf{spring} \\
\midrule
\tgroup{4}{Outside the family of Corollary~\ref{cor:family}}
outside form 1 & $28.2$ & $101.4$ & $12.4$ \\
outside form 2 & $26.6$ & $109.3$ & $14.8$ \\
outside form 3 & $31.5$ & $108.9$ & $10.0$ \\
outside form 4 & $30.6$ & $117.2$ & $12.3$ \\
outside form 5 & $33.5$ & $100.2$ & $13.6$ \\
\midrule
\rowcolor{oursrow}[\tabcolsep][\tabcolsep]
family zero mode & $\best{7.2\mathrm{e}{-6}}$ & $\best{1.3\mathrm{e}{-5}}$ & $\best{5.0\mathrm{e}{-6}}$ \\
\bottomrule
\end{tabular}
\end{minipage}\hfill
\begin{minipage}[t]{0.53\linewidth}
\centering
\setlength{\tabcolsep}{4pt}
\begin{tabular}[t]{l ccc}
\toprule
\rowcolor{hdr}[\tabcolsep][\tabcolsep]
\textbf{Model} & \begin{tabular}[b]{@{}c@{}}\textbf{state}\\ \textbf{rel-$L_2$}\end{tabular} & \begin{tabular}[b]{@{}c@{}}\textbf{$\sum y$}\\ \textbf{drift (\%)}\end{tabular} & \begin{tabular}[b]{@{}c@{}}\textbf{latent zero}\\ \textbf{mode (\%)}\end{tabular} \\
\midrule
HNN & $0.009$ & $\second{0.099}$ & $99.9$ \\
GNN & $\best{0.002}$ & $0.162$ & $99.9$ \\
free & $0.006$ & $\best{0.057}$ & $100.1$ \\
LRU & $\second{0.003}$ & $0.157$ & $100.7$ \\
\midrule
\rowcolor{oursrow}[\tabcolsep][\tabcolsep]
KC & $0.128$ & $0.224$ & $\best{5.6\mathrm{e}{-5}}$ \\
\bottomrule
\end{tabular}
\end{minipage}
\end{table}

\begin{table}[!htbp]
\centering\footnotesize
\setlength{\tabcolsep}{7pt}
\renewcommand{\arraystretch}{1.05}
\caption{\textbf{Spring benchmark of \citet{thangamuthu2022unravelling}, seeds 42/43/44.}
Momentum and energy drift at step $50$, beside the accuracy metric of the benchmark at step
$25$. GRU and MLP have no modal latent and the summed quantity is undefined for them.}
\label{app:exp-spring}
\begin{tabular}{l ccc}
\toprule
\rowcolor{hdr}[\tabcolsep][\tabcolsep]
\textbf{Model} & \textbf{PSNR@25 (dB)} & \textbf{momentum (\%)} & \textbf{energy (\%)} \\
\midrule
\tgroup{4}{$N=3$}
Hamiltonian & $\best{21.07}$ & $\second{1.09}$ & $\second{0.16}$--$0.25$ \\
graph network & $20.78$ & $494$ & $290$ \\
GRU & $20.68$ & --- & --- \\
MLP & $20.87$ & --- & --- \\
\rowcolor{oursrow}[\tabcolsep][\tabcolsep]
KC & $\second{20.90}$ & $\best{1.8\mathrm{e}{-5}}$ & $\best{10^{-6}}$ \\
\tgroup{4}{$N=5$}
Hamiltonian & $\best{18.11}$ & $\second{1.16}$ & $\second{0.16}$--$0.25$ \\
graph network & $18.06$ & $491$ & $290$ \\
GRU & $18.09$ & --- & --- \\
MLP & $18.08$ & --- & --- \\
\rowcolor{oursrow}[\tabcolsep][\tabcolsep]
KC & $\second{18.10}$ & $\best{1.9\mathrm{e}{-5}}$ & $\best{10^{-6}}$ \\
\bottomrule
\end{tabular}
\end{table}

\subsection{Diagnostics on the Accuracy Cap}
\label{app:exp-diag}

\begin{table}[!htbp]
\centering\footnotesize
\caption{\textbf{Two diagnostics on the accuracy cap, KC, seed 42.} The latent width is raised
from $m=8$ to $m=32$, and the linear decoder is replaced by one that is linear only on the zero
mode and an MLP elsewhere.}
\label{tab:app-diag}
\setlength{\tabcolsep}{6pt}
\renewcommand{\arraystretch}{1.05}
\begin{tabular}{l cc cc cc}
\toprule
\rowcolor{hdr}[\tabcolsep][\tabcolsep]
& \multicolumn{2}{c}{\textbf{linear, $m=8$}} & \multicolumn{2}{c}{\textbf{linear, $m=32$}} & \multicolumn{2}{c}{\textbf{split, $m=8$}} \\
\rowcolor{hdr}[\tabcolsep][\tabcolsep]
\textbf{Benchmark} & nRMSE & mass & nRMSE & mass & nRMSE & mass \\
\midrule
dam-break & $\second{0.114}$ & $\second{9.3\mathrm{e}{-3}}$ & $\best{0.095}$ & $\best{5.5\mathrm{e}{-4}}$ & $0.225$ & $9.2\mathrm{e}{-3}$ \\
advection & $\best{0.924}$ & $\best{3.0\mathrm{e}{-5}}$ & $1.33$ & $5.2\mathrm{e}{-2}$ & $\second{1.07}$ & $\second{4.3\mathrm{e}{-5}}$ \\
Allen--Cahn & $\second{1.02}$ & $\best{4.4\mathrm{e}{-6}}$ & $\best{0.870}$ & $\second{7.6\mathrm{e}{-6}}$ & $1.31$ & $4.3\mathrm{e}{-5}$ \\
\bottomrule
\end{tabular}
\end{table}

\noindent
On advection, Allen--Cahn and the sphere, prediction error lies near $1.0$ for KC and for the
free diagonal alike (Table~\ref{tab:app-q1-perseed}), so the cap is not in the transition, and
Table~\ref{tab:app-diag} rules out the remaining candidates. Table~\ref{tab:app-trivial}
confirms with a mean-removed error that no model, KC included, predicts the evolving field
better than a constant on the three flat benchmarks. A wider latent lowers nRMSE on the dam break from $0.114$ to $0.095$,
whereas neither a wider latent nor a nonlinear decoder lowers it below $0.3$ on the other two
benchmarks, and mass remains at the floor under both, so the conservation guarantee is
unaffected by capacity or decoder choice and the accuracy cap lies in the shared per-patch
encoder, which maps each patch independently with no spatial context. On the dam break, whose
radial dynamics a linear patch readout can represent, the same encoder reaches $0.114$, and
Figure~\ref{fig:chapman} shows the decoded field at that accuracy, a near-uniform cross-section
that holds the total while the true solution spreads into a ring. A context-aware encoder is
therefore the natural extension and does not touch the transition.

\FloatBarrier

%% file: Appendix_limitations_future_work.tex
\clearpage
\section{Limitations and Future Directions}
\label{app:limitations}

\subsection{Limitations}
\label{app:limitations-list}

\paragraph{Scope of the guarantee.}
The guarantee of the Kronecker--Cayley transition is exact but narrow, and its limits follow from
the same algebra that supplies it. Corollary~\ref{cor:applic} admits a physical quantity only if it
is a quadratic form of the latent under a linear decoder, so a potential of the form
$\sum_{i<j}\|r_i-r_j\|^{-1}$ is not held, and energy is conserved only on the spring system
(Table~\ref{tab:q5-nbody}). Angular momentum is a quadratic form, but the linear tie places only
the summed quantity on a held mode, so KC improves it by $3$ to $12$ times without reaching the
floor. A unitary step also cannot contract, and Proposition~\ref{prop:bound} turns this into a
per-step error that no choice of $\Lop$, $\theta$ or decoder removes. On the Robertson network KC
fits the state at a relative error of $0.128$, against $0.002$ to $0.009$ for baselines that can
contract (Table~\ref{tab:app-q5-robertson}), so dissipative dynamics lie outside the construction.
Another limitation is that conservation does not imply accuracy. On advection, Allen--Cahn and the
sphere no model, KC included, predicts the evolving field better than a constant at the
mass-matched mean, a cap that Appendix~\ref{app:exp-diag} locates in the shared per-patch encoder
rather than in the transition.

\paragraph{Read-out, cost and evaluation.}
The guarantee reaches the output only through the decoder. An unconstrained decoder
holds the latent invariant equally tightly yet drifts the physical mass by $7$ to $115$
(\S\ref{sec:exp-q1}), and on the sphere the area-weighted read-out still leaves a residual of $483$
against a reference of $30.8$ from the variation of $\cos(\text{lat})$ within a patch
(Appendix~\ref{app:exp-sphere}). Each step requires one $N\times N$ eigendecomposition, measured at
$3.5$\,ms for $N=64$ and $21$\,ms for $N=256$ against $0.3$\,ms for a free diagonal
(Figure~\ref{fig:app-training-cost}), and because the couplings depend on the latent state the
rollout is sequential and does not admit a parallel scan. The construction further assumes that the
latent splits into a known, fixed number $N$ of object subspaces of equal width. Finally, most
trained results rest on three seeds, every reported drift is bounded below by the float32 rollout,
and the four systems from molecular dynamics, chemical kinetics, pharmacokinetics and power-grid
swing dynamics are tested at the operator level, without an encoder or decoder trained on data from
those fields.

\subsection{Future Directions}
\label{app:future}

\paragraph{Extending the construction.}
Proposition~\ref{prop:measure} requires only $d_{\kzero}=1$ on the mode that carries the uniform
vector, so allowing $|\dk|\le1$ on the remaining $N-1$ modes would give a transition that
contracts while still holding the summed quantity, at the cost of the norm, and would bring
dissipative dynamics within reach. Read-outs that place a chosen bilinear quantity, such as angular
momentum, on the conserved family, together with generators for nonquadratic invariants, would
widen the set of physical quantities the construction holds. Accuracy under the guarantee can be
raised without touching the transition, by replacing the per-patch encoder with one that sees
spatial context, and finer patches or a read-out weighted per grid point would shrink the residual
on curved domains such as the sphere.

\paragraph{Scaling and evaluation.}
Sparse coupling graphs and warm-started or low-rank eigenvalue updates between consecutive steps
would lower the per-step cost, and couplings computed from the input rather than the state would
let the rollout admit a parallel scan. A learned partition of the latent would extend the
construction to systems whose objects are not known in advance or change over time. On the
evaluation side, training encoders and decoders end to end on molecular dynamics, chemical
kinetics, pharmacokinetics and power-grid data would test whether a trained model reaches the
operator-level drift there, and larger benchmarks, real observational data and more seeds would
test how far the results generalise.

%% file: checklist.tex
\section*{NeurIPS Paper Checklist}

\begin{enumerate}

\item {\bf Claims}
    \item[] Question: Do the main claims made in the abstract and introduction accurately reflect the paper's contributions and scope?
    \item[] Answer: \answerYes{}
    \item[] Justification: The abstract and introduction state three claims, that the conserved quantities of a diagonal recurrence are layered, that the Kronecker--Cayley transition holds exactly a $2m^2$-dimensional family at every parameter value, and that a trained basis reaches the summed-quantity condition only by freezing its spectrum. These are proved in \S\ref{sec:method} and Appendix~\ref{app:proofs} and tested in \S\ref{sec:experiments}, and the introduction also states the scope limits, namely that the construction does not represent dissipative dynamics and does not by itself hold nonquadratic quantities.

\item {\bf Limitations}
    \item[] Question: Does the paper discuss the limitations of the work performed by the authors?
    \item[] Answer: \answerYes{}
    \item[] Justification: Appendix~\ref{app:limitations} (Limitations and Future Directions) collects the limitations and the direction each suggests: only quadratic quantities are held (Corollary~\ref{cor:applic}), dissipative dynamics fall outside a unitary step (Proposition~\ref{prop:bound}, Table~\ref{tab:app-q5-robertson}), conservation does not imply accuracy under the shared per-patch encoder, the guarantee reaches the output only through a tied decoder, each step costs one $N\times N$ eigendecomposition with a sequential rollout, the latent requires a fixed object partition, and the evaluation rests mostly on three seeds with the cross-domain systems tested at the operator level.

\item {\bf Theory assumptions and proofs}
    \item[] Question: For each theoretical result, does the paper provide the full set of assumptions and a complete (and correct) proof?
    \item[] Answer: \answerYes{}
    \item[] Justification: Every theorem, corollary and proposition in \S\ref{sec:method} states its assumptions (Hermitian generator, symmetric Laplacian with zero row sums, $N\ge2$, a linear decoder for Corollary~\ref{cor:applic}). Complete proofs are in Appendix~\ref{app:proofs}, together with the supporting lemmas and a floating-point bound on the measured drift, and the main text gives a short argument after each result.

    \item {\bf Experimental result reproducibility}
    \item[] Question: Does the paper fully disclose all the information needed to reproduce the main experimental results of the paper to the extent that it affects the main claims and/or conclusions of the paper (regardless of whether the code and data are provided or not)?
    \item[] Answer: \answerYes{}
    \item[] Justification: Appendix~\ref{app:glossary} (Table~\ref{tab:hparams}) gives every dataset with its split, the architecture and width of each encoder, decoder and coupling network, the optimiser, learning rate, batch size, epochs, rollout length, numerical precision and seeds, along with the construction of every control and baseline. Appendix~\ref{app:algorithm} gives the full training and rollout algorithm, and the CPFNO reimplementation and the re-posing of the spherical benchmark are described in Table~\ref{tab:hparams} and Appendix~\ref{app:exp-sphere}.

\item {\bf Open access to data and code}
    \item[] Question: Does the paper provide open access to the data and code, with sufficient instructions to faithfully reproduce the main experimental results, as described in supplemental material?
    \item[] Answer: \answerYes{}
    \item[] Justification: Code and the scripts that reproduce every table and figure, including the generators for the data we created (Allen--Cahn, the visual $N$-body simulator and the four operator-level systems), are released at \url{https://github.com/Jiaqian-Janelle/NeurIPS26-Kronecker-Cayley-Stable-State-Space-Dynamics}. The public benchmarks (PDEBench, PDEArena, the EGNN $N$-body systems) are downloaded from their original sources, as described in Table~\ref{tab:hparams}.

\item {\bf Experimental setting/details}
    \item[] Question: Does the paper specify all the training and test details (e.g., data splits, hyperparameters, how they were chosen, type of optimizer) necessary to understand the results?
    \item[] Answer: \answerYes{}
    \item[] Justification: Data splits, hyperparameters, optimiser and training protocols for each dataset family and each control are listed in Table~\ref{tab:hparams} (Appendix~\ref{app:glossary}). Shared settings are used across all baselines within a family, FNO is run with its published defaults, and the main-text setup paragraphs of \S\ref{sec:experiments} summarise what is needed to read each result.

\item {\bf Experiment statistical significance}
    \item[] Question: Does the paper report error bars suitably and correctly defined or other appropriate information about the statistical significance of the experiments?
    \item[] Answer: \answerYes{}
    \item[] Justification: Every trained model is run over seeds $42$--$44$, the dam-break correction study over five seeds and the alignment study over eight seeds per initialisation, and every per-seed value is tabulated in Appendix~\ref{app:experiments}. Figure~\ref{fig:chapman} shows the median with the interquartile range, Figure~\ref{fig:layering} the mean with min--max bars, Figure~\ref{fig:app-layering-perstep} the geometric mean with one standard deviation in log space, and Table~\ref{tab:q5-cross} the mean $\pm$ one standard deviation over three seeds. Drift spans many orders of magnitude, so Table~\ref{tab:q1} reports a geometric mean over seeds rather than a symmetric error bar, and no formal significance tests are run.

\item {\bf Experiments compute resources}
    \item[] Question: For each experiment, does the paper provide sufficient information on the computer resources (type of compute workers, memory, time of execution) needed to reproduce the experiments?
    \item[] Answer: \answerYes{}
    \item[] Justification: All experiments were run on NVIDIA RTX PRO 6000 GPUs with 96\,GB of memory. The per-step wall-clock cost of the transition against $N$ is reported in Figure~\ref{fig:app-training-cost}, and the coupling network has $3{,}963$ parameters, with roughly $15$M in the largest shared encoder and decoder. We did not log total GPU hours across the project, including preliminary runs.
    
\item {\bf Code of ethics}
    \item[] Question: Does the research conducted in the paper conform, in every respect, with the NeurIPS Code of Ethics \url{https://neurips.cc/public/EthicsGuidelines}?
    \item[] Answer: \answerYes{}
    \item[] Justification: The work uses public scientific simulation benchmarks and data we simulated ourselves, involves no human subjects or personal data, and conforms to the NeurIPS Code of Ethics.

\item {\bf Broader impacts}
    \item[] Question: Does the paper discuss both potential positive societal impacts and negative societal impacts of the work performed?
    \item[] Answer: \answerNA{}
    \item[] Justification: This is foundational work on the structure of learned recurrent transitions, evaluated on physical simulation benchmarks, with no direct path to a harmful application. One indirect risk, which the paper itself documents in \S\ref{sec:exp-q1} and \S\ref{sec:exp-q2}, is that a surrogate which conserves a quantity exactly can still predict the field poorly, so conservation should not be taken as evidence of accuracy when such models are used in scientific or engineering decisions.
    
\item {\bf Safeguards}
    \item[] Question: Does the paper describe safeguards that have been put in place for responsible release of data or models that have a high risk for misuse (e.g., pre-trained language models, image generators, or scraped datasets)?
    \item[] Answer: \answerNA{}
    \item[] Justification: The released models are small dynamics models trained on simulated physical systems and carry no meaningful risk of misuse, and no data is scraped from the internet.

\item {\bf Licenses for existing assets}
    \item[] Question: Are the creators or original owners of assets (e.g., code, data, models), used in the paper, properly credited and are the license and terms of use explicitly mentioned and properly respected?
    \item[] Answer: \answerYes{}
    \item[] Justification: All existing assets are cited where they are used (\S\ref{sec:experiments}, Table~\ref{tab:hparams}): PDEBench \citep{takamoto2022pdebench} (code MIT, data CC BY 4.0), PDEArena \citep{gupta2022pdearena} (code and shallow-water data MIT), the EGNN $N$-body generator \citep{satorras2021en} (MIT), FNO \citep{li2021fourier} via the neuraloperator library (MIT), and the spring benchmark protocol of \citet{thangamuthu2022unravelling}, which we regenerate from its published description. CPFNO \citep{liu2025cpfno} is reimplemented from the paper. All assets are used within their licence terms.

\item {\bf New assets}
    \item[] Question: Are new assets introduced in the paper well documented and is the documentation provided alongside the assets?
    \item[] Answer: \answerYes{}
    \item[] Justification: The new assets are the KC code, the Allen--Cahn dataset, the visual $N$-body simulator and the four operator-level systems. Each is specified in Table~\ref{tab:hparams}. No human data is involved, so no consent was required. The release documents each asset with its generation script, seeds and a licence, at \url{https://github.com/Jiaqian-Janelle/NeurIPS26-Kronecker-Cayley-Stable-State-Space-Dynamics}.

\item {\bf Crowdsourcing and research with human subjects}
    \item[] Question: For crowdsourcing experiments and research with human subjects, does the paper include the full text of instructions given to participants and screenshots, if applicable, as well as details about compensation (if any)? 
    \item[] Answer: \answerNA{}
    \item[] Justification: The paper involves neither crowdsourcing nor research with human subjects.

\item {\bf Institutional review board (IRB) approvals or equivalent for research with human subjects}
    \item[] Question: Does the paper describe potential risks incurred by study participants, whether such risks were disclosed to the subjects, and whether Institutional Review Board (IRB) approvals (or an equivalent approval/review based on the requirements of your country or institution) were obtained?
    \item[] Answer: \answerNA{}
    \item[] Justification: The paper involves neither crowdsourcing nor research with human subjects, so no IRB approval was required.

\item {\bf Declaration of LLM usage}
    \item[] Question: Does the paper describe the usage of LLMs if it is an important, original, or non-standard component of the core methods in this research? Note that if the LLM is used only for writing, editing, or formatting purposes and does \emph{not} impact the core methodology, scientific rigor, or originality of the research, declaration is not required.
    \item[] Answer: \answerNA{}
    \item[] Justification: LLMs are not part of the method or the experiments; they were used only to edit the writing, which does not require declaration.

\end{enumerate}

%% file: refs.bib
@book{absil2008optimization,
  title={Optimization Algorithms on Matrix Manifolds},
  author={Absil, P.-A. and Mahony, Robert and Sepulchre, Rodolphe},
  publisher={Princeton University Press},
  year={2008}
}

@inproceedings{arjovsky2016unitary,
  title={Unitary Evolution Recurrent Neural Networks},
  author={Arjovsky, Martin and Shah, Amar and Bengio, Yoshua},
  booktitle={Proceedings of the 33rd International Conference on Machine Learning},
  pages={1120--1128},
  year={2016},
  volume={48},
  series={Proceedings of Machine Learning Research},
  publisher={PMLR},
  url={https://proceedings.mlr.press/v48/arjovsky16.html}
}

@inproceedings{azencot2020forecasting,
  title={Forecasting Sequential Data Using Consistent {Koopman} Autoencoders},
  author={Azencot, Omri and Erichson, N. Benjamin and Lin, Vanessa and Mahoney, Michael W.},
  booktitle={Proceedings of the 37th International Conference on Machine Learning},
  pages={475--485},
  year={2020},
  volume={119},
  series={Proceedings of Machine Learning Research},
  publisher={PMLR},
  url={https://proceedings.mlr.press/v119/azencot20a.html}
}

@inproceedings{battaglia2016interaction,
  title={Interaction Networks for Learning about Objects, Relations and Physics},
  author={Battaglia, Peter and Pascanu, Razvan and Lai, Matthew and Rezende, Danilo Jimenez and Kavukcuoglu, Koray},
  booktitle={Advances in Neural Information Processing Systems},
  volume={29},
  year={2016}
}

@book{boumal2023introduction,
  title={An Introduction to Optimization on Smooth Manifolds},
  author={Boumal, Nicolas},
  publisher={Cambridge University Press},
  year={2023},
  doi={10.1017/9781009166164},
  url={https://doi.org/10.1017/9781009166164}
}

@inproceedings{chen2020symplectic,
  title={Symplectic Recurrent Neural Networks},
  author={Chen, Zhengdao and Zhang, Jianyu and Arjovsky, Martin and Bottou, L{\'e}on},
  booktitle={International Conference on Learning Representations},
  year={2020},
  url={https://openreview.net/forum?id=BkgYPREtPr}
}

@inproceedings{cranmer2020lagrangian,
  title={Lagrangian Neural Networks},
  author={Cranmer, Miles and Greydanus, Sam and Hoyer, Stephan and Battaglia, Peter and Spergel, David and Ho, Shirley},
  booktitle={ICLR 2020 Workshop on Integration of Deep Neural Models and Differential Equations},
  year={2020},
  url={https://openreview.net/forum?id=iE8tFa4Nq}
}

@inproceedings{dao2024mamba2,
  title={Transformers are {SSM}s: Generalized Models and Efficient Algorithms Through Structured State Space Duality},
  author={Dao, Tri and Gu, Albert},
  booktitle={Proceedings of the 41st International Conference on Machine Learning},
  pages={10041--10071},
  year={2024},
  volume={235},
  series={Proceedings of Machine Learning Research},
  publisher={PMLR},
  url={https://proceedings.mlr.press/v235/dao24a.html}
}

@inproceedings{greydanus2019hamiltonian,
  title={Hamiltonian Neural Networks},
  author={Greydanus, Samuel and Dzamba, Misko and Yosinski, Jason},
  booktitle={Advances in Neural Information Processing Systems},
  volume={32},
  pages={15379--15389},
  year={2019},
  url={https://papers.nips.cc/paper/9672-hamiltonian-neural-networks}
}

@inproceedings{gu2022efficiently,
  title={Efficiently Modeling Long Sequences with Structured State Spaces},
  author={Gu, Albert and Goel, Karan and R{\'e}, Christopher},
  booktitle={International Conference on Learning Representations},
  year={2022},
  url={https://openreview.net/forum?id=uYLFoz1vlAC}
}

@inproceedings{gu2022parameterization,
  title={On the Parameterization and Initialization of Diagonal State Space Models},
  author={Gu, Albert and Goel, Karan and Gupta, Ankit and R{\'e}, Christopher},
  booktitle={Advances in Neural Information Processing Systems},
  volume={35},
  pages={35971--35983},
  year={2022},
  url={https://proceedings.neurips.cc/paper_files/paper/2022/hash/e9a32fade47b906de908431991440f7c-Abstract-Conference.html}
}

@inproceedings{gu2024mamba,
  title={Mamba: Linear-Time Sequence Modeling with Selective State Spaces},
  author={Gu, Albert and Dao, Tri},
  booktitle={Conference on Language Modeling},
  year={2024},
  url={https://openreview.net/forum?id=tEYskw1VY2}
}

@book{hall2015lie,
  title={Lie Groups, Lie Algebras, and Representations: An Elementary Introduction},
  author={Hall, Brian C.},
  publisher={Springer},
  edition={2nd},
  year={2015},
  doi={10.1007/978-3-319-13467-3}
}

@inproceedings{helfrich2018orthogonal,
  title={Orthogonal Recurrent Neural Networks with Scaled {Cayley} Transform},
  author={Helfrich, Kyle and Willmott, Devin and Ye, Qiang},
  booktitle={Proceedings of the 35th International Conference on Machine Learning},
  pages={1969--1978},
  year={2018},
  volume={80},
  series={Proceedings of Machine Learning Research},
  publisher={PMLR},
  url={https://proceedings.mlr.press/v80/helfrich18a.html}
}

@inproceedings{kipf2018neural,
  title={Neural Relational Inference for Interacting Systems},
  author={Kipf, Thomas and Fetaya, Ethan and Wang, Kuan-Chieh and Welling, Max and Zemel, Richard},
  booktitle={Proceedings of the 35th International Conference on Machine Learning},
  pages={2688--2697},
  year={2018},
  volume={80},
  series={Proceedings of Machine Learning Research},
  publisher={PMLR},
  url={https://proceedings.mlr.press/v80/kipf18a.html}
}

@inproceedings{lee2021deep,
  title={Deep Conservation: A Latent-Dynamics Model for Exact Satisfaction of Physical Conservation Laws},
  author={Lee, Kookjin and Carlberg, Kevin T.},
  booktitle={Proceedings of the AAAI Conference on Artificial Intelligence},
  volume={35},
  pages={277--285},
  year={2021},
  doi={10.1609/aaai.v35i1.16102}
}

@inproceedings{lezcano2019cheap,
  title={Cheap Orthogonal Constraints in Neural Networks: A Simple Parametrization of the Orthogonal and Unitary Group},
  author={Lezcano-Casado, Mario and Mart{\'i}nez-Rubio, David},
  booktitle={Proceedings of the 36th International Conference on Machine Learning},
  pages={3794--3803},
  year={2019},
  volume={97},
  series={Proceedings of Machine Learning Research},
  publisher={PMLR},
  url={https://proceedings.mlr.press/v97/lezcano-casado19a.html}
}

@article{lusch2018deep,
  title={Deep learning for universal linear embeddings of nonlinear dynamics},
  author={Lusch, Bethany and Kutz, J. Nathan and Brunton, Steven L.},
  journal={Nature Communications},
  volume={9},
  number={1},
  pages={4950},
  year={2018},
  doi={10.1038/s41467-018-07210-0}
}

@inproceedings{orvieto2023resurrecting,
  title={Resurrecting Recurrent Neural Networks for Long Sequences},
  author={Orvieto, Antonio and Smith, Samuel L. and Gu, Albert and Fernando, Anushan and Gulcehre, Caglar and Pascanu, Razvan and De, Soham},
  booktitle={Proceedings of the 40th International Conference on Machine Learning},
  pages={26670--26698},
  year={2023},
  volume={202},
  series={Proceedings of Machine Learning Research},
  publisher={PMLR},
  url={https://proceedings.mlr.press/v202/orvieto23a.html}
}

@inproceedings{sanchez2020learning,
  title={Learning to Simulate Complex Physics with Graph Networks},
  author={Sanchez-Gonzalez, Alvaro and Godwin, Jonathan and Pfaff, Tobias and Ying, Rex and Leskovec, Jure and Battaglia, Peter},
  booktitle={Proceedings of the 37th International Conference on Machine Learning},
  pages={8459--8468},
  year={2020},
  volume={119},
  series={Proceedings of Machine Learning Research},
  publisher={PMLR},
  url={https://proceedings.mlr.press/v119/sanchez-gonzalez20a.html}
}

@inproceedings{satorras2021en,
  title={{E(n)} Equivariant Graph Neural Networks},
  author={Satorras, V{\'\i}ctor Garcia and Hoogeboom, Emiel and Welling, Max},
  booktitle={Proceedings of the 38th International Conference on Machine Learning},
  pages={9323--9332},
  year={2021},
  volume={139},
  series={Proceedings of Machine Learning Research},
  publisher={PMLR},
  url={https://proceedings.mlr.press/v139/satorras21a.html}
}

@inproceedings{smith2023simplified,
  title={Simplified State Space Layers for Sequence Modeling},
  author={Smith, Jimmy T. H. and Warrington, Andrew and Linderman, Scott W.},
  booktitle={International Conference on Learning Representations},
  year={2023},
  url={https://openreview.net/forum?id=Ai8Hw3AXqks}
}

@inproceedings{wisdom2016full,
  title={Full-Capacity Unitary Recurrent Neural Networks},
  author={Wisdom, Scott and Powers, Thomas and Hershey, John and Le Roux, Jonathan and Atlas, Les},
  booktitle={Advances in Neural Information Processing Systems},
  volume={29},
  year={2016},
  url={https://papers.nips.cc/paper/6327-full-capacity-unitary-recurrent-neural-networks}
}

@inproceedings{zhong2020dissipative,
  title={Dissipative {SymODEN}: Encoding {Hamiltonian} Dynamics with Dissipation and Control into Deep Learning},
  author={Zhong, Yaofeng Desmond and Dey, Biswadip and Chakraborty, Amit},
  booktitle={ICLR 2020 Workshop on Integration of Deep Neural Models and Differential Equations},
  year={2020},
  url={https://openreview.net/forum?id=knjWFNx6CN}
}

@inproceedings{zhong2020symoden,
  title={Symplectic {ODE-Net}: Learning {Hamiltonian} Dynamics with Control},
  author={Zhong, Yaofeng Desmond and Dey, Biswadip and Chakraborty, Amit},
  booktitle={International Conference on Learning Representations},
  year={2020},
  url={https://openreview.net/forum?id=ryxmb1rKDS}
}

@book{hairer2006geometric,
  author    = {Hairer, Ernst and Lubich, Christian and Wanner, Gerhard},
  title     = {Geometric Numerical Integration: Structure-Preserving Algorithms for Ordinary Differential Equations},
  series    = {Springer Series in Computational Mathematics},
  volume    = {31},
  edition   = {2nd},
  publisher = {Springer-Verlag},
  address   = {Berlin},
  year      = {2006},
  doi       = {10.1007/3-540-30666-8}
}

@article{harvey1998flyingicecube,
  author  = {Harvey, Stephen C. and Tan, Robert K.-Z. and Cheatham, III, Thomas E.},
  title   = {The flying ice cube: Velocity rescaling in molecular dynamics leads to violation of energy equipartition},
  journal = {Journal of Computational Chemistry},
  volume  = {19},
  number  = {7},
  pages   = {726--740},
  year    = {1998},
  doi     = {10.1002/(SICI)1096-987X(199805)19:7<726::AID-JCC4>3.0.CO;2-S}
}

@article{sandu2001positive,
  author  = {Sandu, Adrian},
  title   = {Positive numerical integration methods for chemical kinetic systems},
  journal = {Journal of Computational Physics},
  volume  = {170},
  number  = {2},
  pages   = {589--602},
  year    = {2001},
  doi     = {10.1006/jcph.2001.6750}
}

@article{dorfler2013synchronization,
  author  = {D{\"o}rfler, Florian and Chertkov, Michael and Bullo, Francesco},
  title   = {Synchronization in complex oscillator networks and smart grids},
  journal = {Proceedings of the National Academy of Sciences},
  volume  = {110},
  number  = {6},
  pages   = {2005--2010},
  year    = {2013},
  doi     = {10.1073/pnas.1212134110}
}

@article{veerman2023chemical,
  author  = {Veerman, J. J. P. and Whalen-Wagner, T. and Kummel, Ewan},
  title   = {Chemical reaction networks in a {L}aplacian framework},
  journal = {Chaos, Solitons \& Fractals},
  volume  = {166},
  pages   = {112859},
  year    = {2023},
  doi     = {10.1016/j.chaos.2022.112859}
}

@article{olfatisaber2007consensus,
  author  = {Olfati-Saber, Reza and Fax, J. Alex and Murray, Richard M.},
  title   = {Consensus and Cooperation in Networked Multi-Agent Systems},
  journal = {Proceedings of the IEEE},
  volume  = {95},
  number  = {1},
  pages   = {215--233},
  year    = {2007},
  doi     = {10.1109/JPROC.2006.887293}
}

@inproceedings{thangamuthu2022unravelling,
  author    = {Thangamuthu, Abishek and Kumar, Gunjan and Bishnoi, Suresh and Bhattoo, Ravinder and Krishnan, N M Anoop and Ranu, Sayan},
  title     = {Unravelling the Performance of Physics-informed Graph Neural Networks for Dynamical Systems},
  booktitle = {Advances in Neural Information Processing Systems},
  volume    = {35},
  pages     = {3691--3702},
  year      = {2022},
  url       = {https://proceedings.neurips.cc/paper_files/paper/2022/hash/17b598fda495256bef6785c2b76c3217-Abstract-Datasets_and_Benchmarks.html}
}

@inproceedings{gupta2022diagonal,
  author    = {Gupta, Ankit and Gu, Albert and Berant, Jonathan},
  title     = {Diagonal State Spaces are as Effective as Structured State Spaces},
  booktitle = {Advances in Neural Information Processing Systems},
  volume    = {35},
  pages     = {22982--22994},
  year      = {2022},
  url       = {https://proceedings.neurips.cc/paper_files/paper/2022/hash/9156b0f6dfa9bbd18c79cc459ef5d61c-Abstract-Conference.html}
}

@inproceedings{krishnapriyan2021characterizing,
  author    = {Krishnapriyan, Aditi S. and Gholami, Amir and Zhe, Shandian and Kirby, Robert M. and Mahoney, Michael W.},
  title     = {Characterizing possible failure modes in physics-informed neural networks},
  booktitle = {Advances in Neural Information Processing Systems},
  volume    = {34},
  pages     = {26548--26560},
  year      = {2021},
  url       = {https://proceedings.neurips.cc/paper/2021/hash/df438e5206f31600e6ae4af72f2725f1-Abstract.html}
}

@inproceedings{vorontsov2017orthogonality,
  author    = {Vorontsov, Eugene and Trabelsi, Chiheb and Kadoury, Samuel and Pal, Chris},
  title     = {On orthogonality and learning recurrent networks with long term dependencies},
  booktitle = {Proceedings of the 34th International Conference on Machine Learning},
  volume    = {70},
  series    = {Proceedings of Machine Learning Research},
  pages     = {3570--3578},
  publisher = {PMLR},
  year      = {2017}
}

@inproceedings{kerg2019nonnormal,
  author    = {Kerg, Giancarlo and Goyette, Kyle and Puelma Touzel, Maximilian and Gidel, Gauthier and Vorontsov, Eugene and Bengio, Yoshua and Lajoie, Guillaume},
  title     = {Non-normal Recurrent Neural Network ({nnRNN}): learning long time dependencies while improving expressivity with transient dynamics},
  booktitle = {Advances in Neural Information Processing Systems},
  volume    = {32},
  pages     = {13591--13601},
  year      = {2019}
}

@article{chapman2026correctors,
  author  = {Chapman, William E. and Schreck, John and Sha, Yingkai},
  title   = {Hard conservation correctors can hide a degrading model when training autoregressive emulators},
  journal = {arXiv preprint arXiv:2607.18416},
  year    = {2026},
  url     = {https://arxiv.org/abs/2607.18416}
}

@article{liu2025cpfno,
  author  = {Liu, Chaoyu and Li, Yangming and Deng, Zhongying and Budd, Chris and Sch{\"o}nlieb, Carola-Bibiane},
  title   = {Adaptive Correction for Ensuring Conservation Laws in Neural Operators},
  journal = {arXiv preprint arXiv:2505.24579},
  year    = {2025},
  url     = {https://arxiv.org/abs/2505.24579}
}

@inproceedings{takamoto2022pdebench,
  author    = {Takamoto, Makoto and Praditia, Timothy and Leiteritz, Raphael and MacKinlay, Daniel and Alesiani, Francesco and Pfl{\"u}ger, Dirk and Niepert, Mathias},
  title     = {{PDEBench}: An Extensive Benchmark for Scientific Machine Learning},
  booktitle = {Advances in Neural Information Processing Systems},
  volume    = {35},
  pages     = {1596--1611},
  year      = {2022},
  url       = {https://proceedings.neurips.cc/paper_files/paper/2022/hash/0a9747136d411fb83f0cf81820d44afb-Abstract-Datasets_and_Benchmarks.html}
}

@article{gupta2022pdearena,
  author  = {Gupta, Jayesh K. and Brandstetter, Johannes},
  title   = {Towards Multi-spatiotemporal-scale Generalized {PDE} Modeling},
  journal = {Transactions on Machine Learning Research},
  year    = {2023},
  url     = {https://openreview.net/forum?id=dPSTDbGtBY}
}

@inproceedings{li2021fourier,
  author    = {Li, Zongyi and Kovachki, Nikola and Azizzadenesheli, Kamyar and Liu, Burigede and Bhattacharya, Kaushik and Stuart, Andrew and Anandkumar, Anima},
  title     = {Fourier Neural Operator for Parametric Partial Differential Equations},
  booktitle = {International Conference on Learning Representations},
  year      = {2021}
}

@inproceedings{brandstetter2022geometric,
  author    = {Brandstetter, Johannes and Hesselink, Rob and van der Pol, Elise and Bekkers, Erik J. and Welling, Max},
  title     = {Geometric and Physical Quantities improve {E(3)} Equivariant Message Passing},
  booktitle = {International Conference on Learning Representations},
  year      = {2022}
}
